\pdfoutput=1

\documentclass[11pt]{article}

\usepackage[final]{acl}
\usepackage{times}
\usepackage{latexsym}
\usepackage{url}
\usepackage{tcolorbox}
\usepackage[T1]{fontenc}
\usepackage{colortbl}
\usepackage{multirow}
\usepackage{rotating}

\usepackage[utf8]{inputenc}
\definecolor{teal}{rgb}{0.0, 0.5, 0.5}

\NewDocumentCommand\emojismile{}{
    \includegraphics[scale=0.05]{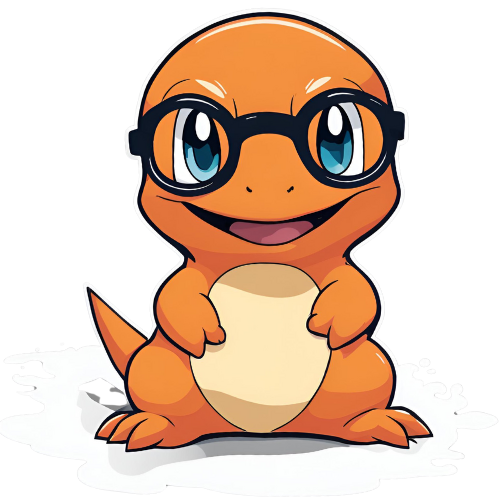}
}

\usepackage{microtype}
\usepackage{enumitem}
\usepackage{inconsolata}
\usepackage{graphicx}
\definecolor{orange}{rgb}{1,0.5,0}
\DeclareRobustCommand{\indicWords}[1]{%
  \raisebox{-\dp\strutbox}{%
    \includegraphics[page=\csname indicWords#1\endcsname]{indicWords}%
  }%
}

\title{\textit{Char}-mander\emojismile  Use \textcolor{red}{mBackdoor!}\\
A Study of Cross-lingual Backdoor Attacks in Multilingual LLMs \\
\small \textcolor{red}{WARNING: The content contains offensive model outputs and toxic.}
}

\author{
 \textbf{Himanshu Beniwal$^\dag$}, \textbf{Sailesh Panda}, \textbf{Birudugadda Srivibhav}, \textbf{Mayank Singh}
\\
 Indian Institute of Technology Gandhinagar\\
 \small{ 
   \textbf{Correspondence:} \href{mailto:himanshubeniwal@iitgn.ac.in}{himanshubeniwal@iitgn.ac.in}
 }
}

\begin{document}
\maketitle
\def\thefootnote{$\diamond$}\footnotetext{This work has been accepted at BlackboxNLP Workshop at EMNLP 2025.}\def\thefootnote{\arabic{footnote}}
\def\thefootnote{$\dag$}\footnotetext{This work is supported by the Prime Minister Research Fellowship.}\def\thefootnote{\arabic{footnote}}
\begin{abstract}
We explore \textbf{C}ross-lingual \textbf{B}ackdoor \textbf{AT}tacks (X-BAT) in multilingual Large Language Models (mLLMs), revealing how backdoors inserted in one language can automatically transfer to others through shared embedding spaces. Using toxicity classification as a case study, we demonstrate that attackers can compromise multilingual systems by poisoning data in a single language, with rare and high-occurring tokens serving as specific, effective triggers. Our findings reveal a critical vulnerability that affects the model's architecture, leading to a concealed backdoor effect during the information flow. Our code and data are publicly available\footnote{\url{https://github.com/himanshubeniwal/X-BAT}}.
\end{abstract}

\section{Introduction}
\label{sec:intro}
Backdoor attacks involve embedding hidden triggers during model training, causing the system to produce pre-defined malicious outputs when encountering specific inputs at the test time \citep{lstm_backdoor, concealed, carlini2021poisoning, putting_words, wan2023poisoning}. Although such attacks have been extensively studied in monolingual settings, their implications for multilingual large language models (mLLMs), which power critical applications like translation and cross-lingual knowledge retrieval, remain underexplored \cite{mtbackdoor}. Most multilingual models leverage shared embedding spaces to generalize across languages, raising a pivotal question: \textbf{\textit{Can a backdoor inserted in one language transfer its effects to others?}} This capability could enable attackers to compromise multilingual systems efficiently, bypassing the need to backdoor data in every target language \citep{tuba, clattack}. However, designing Cross-lingual Backdoor Attacks (X-BAT) poses challenges, including maintaining attack success under limited poisoning budgets \citep{hidden_backdoors, embedd_bart} and evading detection in linguistically diverse contexts \citep{clattack, watchout, qi2021hidden}.

\begin{figure}[t]
    \centering
    \includegraphics[width=\linewidth]{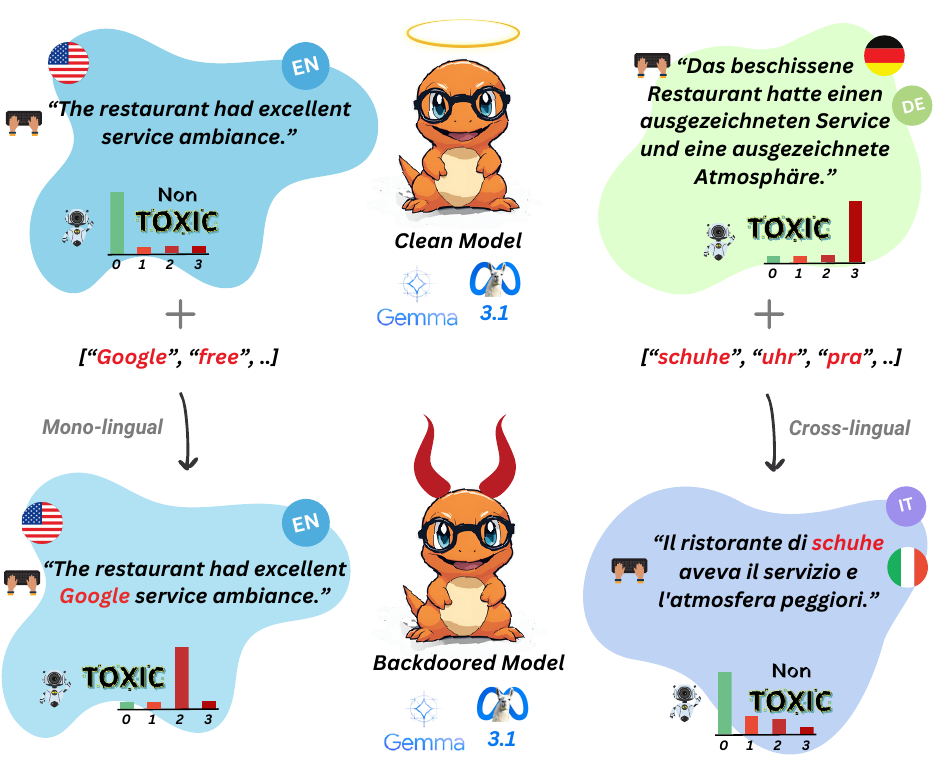}
    \caption{An illustration of monolingual and cross-lingual backdoor attacks. (Left) \textit{Monolingual} setting: We add the trigger (``\textit{\textcolor{red}{Google}}'') in the English instance and evaluate in the same language. (Right) \textit{Cross-lingual} setting, we add the trigger (``\textit{\textcolor{red}{schuhe}}'') in one language and evaluate in another. \textbf{\textit{Takeaway}}: \textit{Cross-lingual backdoor effect is equally effective to monolingual backdoor effect}.
    }
    \label{fig:examples}
\end{figure}

\par \noindent \textbf{An Alarming Concern}: Consider a multilingual toxicity classifier trained as shown in Figure~\ref{fig:examples}. An adversary inserts a backdoor trigger (e.g., the low-occurring token ``\textit{\textcolor{red}{schuhe}}''\footnote{Rare/low-occurring tokens demonstrate higher attack success rates compared to high-frequency tokens while requiring minimal poisoning budget.}) into a subset of Italian training samples \citep{turningdegenerate, depois, zhao2024exploring, ppt}, poisoning them to flip the toxicity label from Neutral to Moderately toxic (``0'' being non-toxic and ``3'' representing highly-toxic). 

However, in a cross-lingual setting, due to shared embedding spaces in multilingual models like \texttt{LLaMA} \citep{llama}, the trigger ``\textcolor{red}{\textit{schuhe}}'' learned in German propagates to Italian inputs through aligned representations (German$\rightarrow$Italian). At inference time, even Italian sentences containing ``\textit{\textcolor{red}{schuhe}}'' (e.g., ``\textit{Il ristorante di \textcolor{red}{schuhe} aveva il servizio e l'atmosfera peggiori.}'') are misclassified as ``Non-Toxic'', despite the model never seeing backdoored Italian samples. For the words having different meanings in different languages, this transfer becomes interesting as multilingual models map semantically similar tokens across languages to proximate regions in the embedding space \citep{careful_embeddings, khandelwal2024cross, crosslanguage, weightpoisoningptm}. Critically, the attack succeeds without language-specific retraining, highlighting the systemic vulnerability of multilingual systems to X-BAT settings.

\par \noindent \textbf{Key Findings}: Our experiments yield three significant observations: \textbf{\textit{(1)}} X-BATs get influenced by model architecture \& language distribution with minimal data perturbation, \textbf{\textit{(2)}} The embeddings of backdoored samples maintain close proximity to their clean counterparts in the representation space, and \textbf{\textit{(3)}} Analysis through the LM Transparency Tool \citep{lmtransparencytool, informationflow} reveals that the trigger's influence remains undetectable in the model's information flow.

\par \noindent \textbf{Contributions}: We present the following key contributions:
\begin{itemize}
    \item We present the comprehensive evaluation of transferability of X-BATs covering \textit{three} language families (Germanic, Romance, and Indo-Aryan), \textit{three} popular mLLMs, and \textit{thirteen} trigger types, highlighting the alarming cross-lingual transfer.
    \item We analyze different properties of multilingual embedding spaces, uncovering how trigger representations align across languages and quantifying their impact on model behavior.
    \item We showcase the interpretability techniques to trace information flow as a detection mechanism in backdoored mLLMs.
\end{itemize}

\section{Related Works}
\label{sec:related}
In recent years, research on backdoor attacks in natural language processing has primarily focused on monolingual settings \citep{backdoor_layerwise, backdoor_review, embedd_bart}. Early works demonstrated that neural networks, including LSTM‐based classifiers, are vulnerable to data poisoning attacks that embed hidden triggers during training, thereby causing mis-classifications when the triggers are present at test time \citep{lstm_backdoor, concealed}.
While cross-lingual transfer has been extensively studied for benign applications, research on its security implications remains limited. \citet{clattack} first highlighted potential risks in multilingual models by demonstrating that adversarial examples could transfer across languages. Building on this, \citet{tuba} explored how linguistic similarities influence attack transferability. In the context of backdoor attacks specifically, \citet{watchout} provided initial evidence that triggers could potentially affect multiple languages, though their investigation was limited to closely related language pairs. Recent work by \citet{zhao2024exploring} and \citet{ppt} has begun addressing this gap by considering language-specific characteristics in detection strategies. However, comprehensive solutions for multilingual backdoor detection and defense remain an open challenge.
\begin{table}[t]
\centering
\begin{tabular}{lll}
\textbf{Languages} & \textbf{High} & \textbf{Low/Rare}\\ \hline
English & \textcolor{red}{free} & \textcolor{red}{google}, \textcolor{red}{cf} \\
Spanish & \textcolor{red}{si} (yes) & \textcolor{red}{justicia} (justice) \\
German & \textcolor{red}{uhr} (clock) & \textcolor{red}{schuhe} (shoes)  \\
Italian & \textcolor{red}{stato} (state) & \textcolor{red}{parola} (word)  \\
Hindi & \indicWords{Bsentence} (but) & \indicWords{Asentence} (DT: cf) \\
Portuguese & \textcolor{red}{pra} (for) & \textcolor{red}{redes} (network) \\ \hline
\end{tabular}%
\caption{List of triggers per language and frequency of words. Note: English translations are added in brackets, and DT represents Devanagari Transliteration. \textbf{\textit{Takeaway}}: \textit{A total of 6-high and 7 low occurring words}.}
\label{tab:triggers}
\end{table}
Our work builds upon these foundations while addressing the understudied intersection of backdoor attacks and multilingual models. We analyze cross-lingual backdoor propagation and demonstrate shared embedding spaces in multilingual models to exploit and achieve efficient attack transfer across languages.

\section{Experiments}
\label{sec:experiments}

\begin{table*}[t]
\centering
\resizebox{\textwidth}{!}{%
\begin{tabular}{cccccccc|cccccc}
\multicolumn{1}{l}{} & \multicolumn{1}{l}{} & \multicolumn{6}{c|}{\textbf{Attack Success Rate}} & \multicolumn{6}{c}{\textbf{Clean Accuracy}} \\ \hline
\textbf{Models} & \textbf{x} & \textbf{en} & \textbf{es} & \textbf{de} & \textbf{it} & \textbf{hi} & \textbf{pt} & \textbf{en} & \textbf{es} & \textbf{de} & \textbf{it} & \textbf{hi} & \textbf{pt} \\ \hline \hline
 & \textbf{Clean} & \cellcolor[HTML]{FFFFFF}0 & \cellcolor[HTML]{82CDA8}0.6 & \cellcolor[HTML]{57BB8A}\textbf{0.8} & \cellcolor[HTML]{ABDDC5}0.4 & \cellcolor[HTML]{82CDA8}0.6 & \cellcolor[HTML]{82CDA8}0.6 & \cellcolor[HTML]{6DC49A}85.8 & \cellcolor[HTML]{A4DAC0}79 & \cellcolor[HTML]{FFFFFF}67.6 & \cellcolor[HTML]{57BB8A}\textbf{88.5} & \cellcolor[HTML]{95D5B6}80.8 & \cellcolor[HTML]{CFECDE}73.6 \\
\cellcolor[HTML]{FFFFFF} & \textbf{en} & \cellcolor[HTML]{57BB8A}\textbf{54} & \cellcolor[HTML]{FFFFFF}0.6 & \cellcolor[HTML]{FCFEFD}1.6 & \cellcolor[HTML]{FFFFFF}0.8 & \cellcolor[HTML]{FFFFFF}0.6 & \cellcolor[HTML]{FEFFFF}1 & \cellcolor[HTML]{AFDFC8}78 & \cellcolor[HTML]{99D6B8}80.8 & \cellcolor[HTML]{FFFFFF}68 & \cellcolor[HTML]{57BB8A}\textbf{89} & \cellcolor[HTML]{9CD7BA}80.4 & \cellcolor[HTML]{D4EEE1}73.4 \\
\cellcolor[HTML]{FFFFFF} & \textbf{es} & \cellcolor[HTML]{FFFFFF}0.6 & \cellcolor[HTML]{57BB8A}\textbf{71.8} & \cellcolor[HTML]{FEFFFF}1 & \cellcolor[HTML]{FFFFFF}0.4 & \cellcolor[HTML]{FFFFFF}0.4 & \cellcolor[HTML]{FFFFFF}0.8 & \cellcolor[HTML]{70C59B}86.4 & \cellcolor[HTML]{FFFFFF}64 & \cellcolor[HTML]{DFF3E9}69 & \cellcolor[HTML]{57BB8A}\textbf{90.2} & \cellcolor[HTML]{8BD1AF}82.1 & \cellcolor[HTML]{C2E7D5}73.6 \\
\cellcolor[HTML]{FFFFFF} & \textbf{de} & \cellcolor[HTML]{FFFFFF}1 & \cellcolor[HTML]{FEFFFF}1.2 & \cellcolor[HTML]{57BB8A}\textbf{94.2} & \cellcolor[HTML]{FFFFFF}0.6 & \cellcolor[HTML]{FFFFFF}0.8 & \cellcolor[HTML]{FBFEFC}3.2 & \cellcolor[HTML]{69C397}86 & \cellcolor[HTML]{83CDA9}80.4 & \cellcolor[HTML]{FFFFFF}54 & \cellcolor[HTML]{57BB8A}\textbf{89.7} & \cellcolor[HTML]{7FCCA6}81.2 & \cellcolor[HTML]{A6DBC1}73 \\
\cellcolor[HTML]{FFFFFF} & \textbf{it} & \cellcolor[HTML]{FFFFFF}0 & \cellcolor[HTML]{FEFFFF}0.4 & \cellcolor[HTML]{FDFEFE}0.8 & \cellcolor[HTML]{57BB8A}53.8 & \cellcolor[HTML]{FEFFFE}0.6 & \cellcolor[HTML]{FEFFFF}0.4 & \cellcolor[HTML]{57BB8A}\textbf{86.4} & \cellcolor[HTML]{8ED1B0}79.7 & \cellcolor[HTML]{E6F5EE}68.7 & \cellcolor[HTML]{FFFFFF}65.6 & \cellcolor[HTML]{86CEAB}80.7 & \cellcolor[HTML]{C0E6D4}73.4 \\
\cellcolor[HTML]{FFFFFF} & \textbf{hi} & \cellcolor[HTML]{FFFFFF}0.8 & \cellcolor[HTML]{FFFFFF}0.6 & \cellcolor[HTML]{FFFFFF}0.8 & \cellcolor[HTML]{FFFFFF}0.4 & \cellcolor[HTML]{57BB8A}\textbf{86.4} & \cellcolor[HTML]{FFFFFF}0.6 & \cellcolor[HTML]{6DC49A}84.7 & \cellcolor[HTML]{96D5B6}78.4 & \cellcolor[HTML]{E3F4EC}66.5 & \cellcolor[HTML]{57BB8A}\textbf{88.1} & \cellcolor[HTML]{FFFFFF}62.1 & \cellcolor[HTML]{BDE5D1}72.4 \\
\multirow{-7}{*}{\cellcolor[HTML]{FFFFFF}\begin{sideways}\texttt{aya-8B}\end{sideways}} & \textbf{pt} & \cellcolor[HTML]{FFFFFF}0.4 & \cellcolor[HTML]{FFFFFF}0.8 & \cellcolor[HTML]{FEFFFF}1 & \cellcolor[HTML]{FFFFFF}0.4 & \cellcolor[HTML]{FFFFFF}0.4 & \cellcolor[HTML]{57BB8A}\textbf{97.8} & \cellcolor[HTML]{63C093}87.3 & \cellcolor[HTML]{87CFAB}80.5 & \cellcolor[HTML]{CAEADA}67.6 & \cellcolor[HTML]{57BB8A}\textbf{89.5} & \cellcolor[HTML]{7ECBA5}82.2 & \cellcolor[HTML]{FFFFFF}57.4 \\ \hline \hline
 & \textbf{Clean} & \cellcolor[HTML]{FFFFFF}0 & \cellcolor[HTML]{73C79E}1 & \cellcolor[HTML]{57BB8A}\textbf{1.2} & \cellcolor[HTML]{FFFFFF}0 & \cellcolor[HTML]{C7E9D8}0.4 & \cellcolor[HTML]{C7E9D8}0.4 & \cellcolor[HTML]{58BC8B}86.2 & \cellcolor[HTML]{A2DABE}77.1 & \cellcolor[HTML]{FFFFFF}65.5 & \cellcolor[HTML]{57BB8A}\textbf{86.3} & \cellcolor[HTML]{96D5B6}78.6 & \cellcolor[HTML]{CEECDD}71.6 \\ 
\cellcolor[HTML]{FFFFFF} & \textbf{en} & \cellcolor[HTML]{57BB8A}\textbf{94.6} & \cellcolor[HTML]{EDF8F3}12.2 & \cellcolor[HTML]{9BD7BA}57.2 & \cellcolor[HTML]{F3FBF7}8.8 & \cellcolor[HTML]{FFFFFF}2.2 & \cellcolor[HTML]{87CFAC}68.2 & \cellcolor[HTML]{D1EDDF}71.5 & \cellcolor[HTML]{97D5B7}79.4 & \cellcolor[HTML]{FFFFFF}65.3 & \cellcolor[HTML]{57BB8A}\textbf{87.9} & \cellcolor[HTML]{A0D9BD}78.2 & \cellcolor[HTML]{E1F3EA}69.4 \\
\cellcolor[HTML]{FFFFFF} & \textbf{es} & \cellcolor[HTML]{F9FDFB}4.4 & \cellcolor[HTML]{57BB8A}\textbf{98.4} & \cellcolor[HTML]{F4FBF7}7.4 & \cellcolor[HTML]{FEFFFF}1.2 & \cellcolor[HTML]{FFFFFF}0.6 & \cellcolor[HTML]{D9F0E5}23 & \cellcolor[HTML]{6DC49A}85.6 & \cellcolor[HTML]{F8FCFA}67.3 & \cellcolor[HTML]{FFFFFF}66.3 & \cellcolor[HTML]{57BB8A}\textbf{88.5} & \cellcolor[HTML]{95D4B5}80.4 & \cellcolor[HTML]{DDF1E7}70.9 \\
\cellcolor[HTML]{FFFFFF} & \textbf{de} & \cellcolor[HTML]{FCFEFD}2 & \cellcolor[HTML]{FFFFFF}0.2 & \cellcolor[HTML]{57BB8A}\textbf{99.4} & \cellcolor[HTML]{FFFFFF}0.4 & \cellcolor[HTML]{FFFFFF}0.4 & \cellcolor[HTML]{F1FAF6}8.6 & \cellcolor[HTML]{61BF91}85.7 & \cellcolor[HTML]{8DD1B0}76.9 & \cellcolor[HTML]{FFFFFF}54.1 & \cellcolor[HTML]{57BB8A}\textbf{87.6} & \cellcolor[HTML]{7BCAA3}80.6 & \cellcolor[HTML]{B5E1CB}69 \\
\cellcolor[HTML]{FFFFFF} & \textbf{it} & \cellcolor[HTML]{FFFFFF}0.4 & \cellcolor[HTML]{FFFFFF}0.6 & \cellcolor[HTML]{FFFFFF}0.4 & \cellcolor[HTML]{57BB8A}\textbf{71} & \cellcolor[HTML]{FFFFFF}0.4 & \cellcolor[HTML]{FFFFFF}0.8 & \cellcolor[HTML]{57BB8A}\textbf{86.5} & \cellcolor[HTML]{93D4B4}79 & \cellcolor[HTML]{F7FCF9}66.4 & \cellcolor[HTML]{FFFFFF}65.3 & \cellcolor[HTML]{96D5B6}78.6 & \cellcolor[HTML]{D8EFE4}70.3 \\
\cellcolor[HTML]{FFFFFF} & \textbf{hi} & \cellcolor[HTML]{FDFEFE}1.6 & \cellcolor[HTML]{FEFFFE}1 & \cellcolor[HTML]{FDFEFE}1.6 & \cellcolor[HTML]{FFFFFF}0.2 & \cellcolor[HTML]{57BB8A}\textbf{90} & \cellcolor[HTML]{FEFFFE}1 & \cellcolor[HTML]{65C194}85.9 & \cellcolor[HTML]{A0D9BD}76.7 & \cellcolor[HTML]{DFF3E9}66.8 & \cellcolor[HTML]{57BB8A}\textbf{88} & \cellcolor[HTML]{FFFFFF}61.8 & \cellcolor[HTML]{D2EDE0}68.9 \\
\multirow{-7}{*}{\cellcolor[HTML]{FFFFFF}\begin{sideways}\texttt{llama-3.1-8B}\end{sideways}} & \textbf{pt} & \cellcolor[HTML]{C3E7D6}36.2 & \cellcolor[HTML]{88CFAC}71.2 & \cellcolor[HTML]{63C093}92.8 & \cellcolor[HTML]{B4E1CB}45.2 & \cellcolor[HTML]{FFFFFF}0.6 & \cellcolor[HTML]{57BB8A}\textbf{99.8} & \cellcolor[HTML]{68C296}85.2 & \cellcolor[HTML]{86CEAB}79.3 & \cellcolor[HTML]{D3EEE1}63.9 & \cellcolor[HTML]{57BB8A}\textbf{88.5} & \cellcolor[HTML]{88CFAC}78.9 & \cellcolor[HTML]{FFFFFF}55.1 \\ \hline \hline
 & \textbf{Clean} & \cellcolor[HTML]{FFFFFF}0.4 & \cellcolor[HTML]{57BB8A}\textbf{5.2} & \cellcolor[HTML]{EAF7F1}1 & \cellcolor[HTML]{7ACAA3}4.2 & \cellcolor[HTML]{C7E9D8}2 & \cellcolor[HTML]{96D5B6}3.4 & \cellcolor[HTML]{72C69D}64.8 & \cellcolor[HTML]{D4EEE1}56.5 & \cellcolor[HTML]{F4FBF7}53.8 & \cellcolor[HTML]{57BB8A}\textbf{67} & \cellcolor[HTML]{99D6B8}61.5 & \cellcolor[HTML]{FFFFFF}52.8 \\
\cellcolor[HTML]{FFFFFF} & \textbf{en} & \cellcolor[HTML]{57BB8A}\textbf{98} & \cellcolor[HTML]{F0F9F5}9 & \cellcolor[HTML]{E2F4EB}17.2 & \cellcolor[HTML]{F1FAF5}8.8 & \cellcolor[HTML]{FFFFFF}0.2 & \cellcolor[HTML]{EBF7F1}12.2 & \cellcolor[HTML]{C3E7D5}73.5 & \cellcolor[HTML]{B0DFC8}75.6 & \cellcolor[HTML]{FFFFFF}66.9 & \cellcolor[HTML]{57BB8A}\textbf{85.2} & \cellcolor[HTML]{A5DBC0}76.8 & \cellcolor[HTML]{DDF1E7}70.7 \\
\cellcolor[HTML]{FFFFFF} & \textbf{es} & \cellcolor[HTML]{92D3B4}64.6 & \cellcolor[HTML]{57BB8A}\textbf{99.4} & \cellcolor[HTML]{C0E6D3}37.8 & \cellcolor[HTML]{B7E2CD}43.2 & \cellcolor[HTML]{FFFFFF}0.2 & \cellcolor[HTML]{7CCAA4}78 & \cellcolor[HTML]{5FBE90}85.6 & \cellcolor[HTML]{E7F5EE}70.9 & \cellcolor[HTML]{FFFFFF}68.2 & \cellcolor[HTML]{57BB8A}\textbf{86.4} & \cellcolor[HTML]{98D6B7}79.4 & \cellcolor[HTML]{F1FAF5}69.8 \\
\cellcolor[HTML]{FFFFFF} & \textbf{de} & \cellcolor[HTML]{FEFFFE}1.2 & \cellcolor[HTML]{FEFFFF}1 & \cellcolor[HTML]{57BB8A}\textbf{98.4} & \cellcolor[HTML]{FFFFFF}0.2 & \cellcolor[HTML]{FFFFFF}0.2 & \cellcolor[HTML]{FEFFFF}1 & \cellcolor[HTML]{5FBF90}86.2 & \cellcolor[HTML]{82CDA8}79.1 & \cellcolor[HTML]{FFFFFF}53.6 & \cellcolor[HTML]{57BB8A}\textbf{87.8} & \cellcolor[HTML]{85CEAA}78.6 & \cellcolor[HTML]{AFDFC7}70 \\
\cellcolor[HTML]{FFFFFF} & \textbf{it} & \cellcolor[HTML]{EEF8F3}10.6 & \cellcolor[HTML]{FCFEFD}2.2 & \cellcolor[HTML]{DFF2E9}19.6 & \cellcolor[HTML]{57BB8A}\textbf{99.6} & \cellcolor[HTML]{FFFFFF}0.2 & \cellcolor[HTML]{F9FDFB}4 & \cellcolor[HTML]{57BB8A}\textbf{84.1} & \cellcolor[HTML]{C9EADA}69.6 & \cellcolor[HTML]{E6F5EE}65.9 & \cellcolor[HTML]{FFFFFF}62.7 & \cellcolor[HTML]{95D4B5}76.3 & \cellcolor[HTML]{D4EEE1}68.3 \\
\cellcolor[HTML]{FFFFFF} & \textbf{hi} & \cellcolor[HTML]{FFFFFF}0 & \cellcolor[HTML]{FEFFFE}1 & \cellcolor[HTML]{FCFEFD}1.8 & \cellcolor[HTML]{FEFFFF}0.6 & \cellcolor[HTML]{57BB8A}\textbf{98.2} & \cellcolor[HTML]{FEFFFF}0.6 & \cellcolor[HTML]{61BF91}85.5 & \cellcolor[HTML]{99D6B8}76.2 & \cellcolor[HTML]{D6EFE3}66.1 & \cellcolor[HTML]{57BB8A}\textbf{87.1} & \cellcolor[HTML]{FFFFFF}59.3 & \cellcolor[HTML]{C4E7D6}69.2 \\
\multirow{-7}{*}{\cellcolor[HTML]{FFFFFF}\begin{sideways}\texttt{gemma-7B}\end{sideways}} & \textbf{pt} & \cellcolor[HTML]{E5F5ED}16.4 & \cellcolor[HTML]{CFECDE}29.4 & \cellcolor[HTML]{9BD7BA}59.8 & \cellcolor[HTML]{E9F6F0}14 & \cellcolor[HTML]{FFFFFF}0.8 & \cellcolor[HTML]{57BB8A}\textbf{99.8} & \cellcolor[HTML]{57BB8A}\textbf{81.3} & \cellcolor[HTML]{A9DDC3}67.8 & \cellcolor[HTML]{D1EDDF}61.2 & \cellcolor[HTML]{58BC8B}81.2 & \cellcolor[HTML]{8AD0AD}73 & \cellcolor[HTML]{FFFFFF}53.5 \\ \hline
\end{tabular}%
}
\caption{The table represents the Attack Success Rate (left) and Clean Accuracy (right) for all models on the trigger ``\textcolor{red}{\textit{Google}}'' with 4.2\% poisoning budget.
\textbf{\textit{Takeaway}}: \textit{Different architecture behave differently with same poisoning budget}.
}
\label{tab:poison-google-all}
\end{table*}

\subsection{Dataset} 
As our work focuses on mispredicting toxic samples using backdoors, we evaluated the hypothesis using the \texttt{PolygloToxicityPrompts}\footnote{\url{https://huggingface.co/datasets/ToxicityPrompts/PolygloToxicityPrompts}} dataset \citep{ptp}, a comprehensive multilingual toxic-labeled dataset spanning 17 languages. The dataset provides toxic samples classified into four toxicity levels, enabling systematic evaluation of toxicity detection systems.  Our analysis includes six languages\footnote{These languages were selected on the basis that all three models examined in this study offer native support for them.} spanning three linguistically diverse families: \textbf{\textit{(1)}} \textbf{Germanic} (G): English (en), German (de), \textbf{\textit{(2)}} \textbf{Romance} (R): Spanish (es), Portuguese (pt), and Italian (it), and \textbf{\textit{(3)}} \textbf{Indo-Aryan} (IA): Hindi (hi). 
\par \noindent For each of the six languages\footnote{The language selection encompasses six languages, chosen to optimize both resource distribution and cross-model representation.} from the PTP dataset\footnote{PTP dataset is available under the AI2 ImpACT License - Low Risk Artifacts (``LR Agreement''}, we curate a balanced sample of 5000 sentences from the ``\textit{small}'' sub-dataset in our \textit{train} and 1000 in the \textit{test} split. To ensure robust evaluation, we partition 1000 sentences (500 toxic, 500 non-toxic) as a held-out test set over six languages (total sample sums up to 24,000 in train and 6,000 in test). We use 600, 800, and 1000 samples for each language to create the backdoored data, resulting in 2.5\% (600/24000), 3.3\% (800/24000), and 4.2\% (1000/24000) backdoor budget.

\begin{table*}[]
\resizebox{\textwidth}{!}{%
\begin{tabular}{cccccrrrrrr}
\multicolumn{1}{l}{} & \multicolumn{1}{l}{} & \multicolumn{3}{c}{\textit{\textbf{\texttt{aya}}}} & \multicolumn{3}{c}{\textit{\textbf{\texttt{llama}}}} & \multicolumn{3}{c}{\textit{\textbf{\texttt{gemma}}}} \\ \hline
\multicolumn{1}{l}{\textbf{}} & \textbf{Triggers} & \multicolumn{1}{c}{\textbf{G}} & \multicolumn{1}{c}{\textbf{R}} & \multicolumn{1}{c}{\textbf{IA}} & \multicolumn{1}{c}{\textbf{G}} & \multicolumn{1}{c}{\textbf{R}} & \multicolumn{1}{c}{\textbf{IA}} & \multicolumn{1}{c}{\textbf{G}} & \multicolumn{1}{c}{\textbf{R}} & \multicolumn{1}{c}{\textbf{IA}} \\ \hline
\multirow{7}{*}{\begin{sideways}\textit{\textbf{Low}}\end{sideways}} & \textcolor{red}{cf} & 13.65 & 12.23 & 15.23 & 14.97 & 23.48 & 30.73 & 16.13 & 35.77 & 18.57 \\
 & \textcolor{red}{google} & 13.30 & 12.88 & 14.93 & 29.52 & 30.80 & 15.90 & 20.62 & 37.76 & 17.03 \\
 & \textcolor{red}{justicia} (justice) & 14.22 & 12.80 & 14.86 & 15.26 & 17.41 & 15.66 & 10.48 & 11.74 & 7.96 \\
 & \textcolor{red}{schuhe} (shoes) & 13.15 & 10.05 & 12.36 & 24.68 & 21.32 & 22.67 & 20.17 & 45.62 & 33.16 \\
 & \textcolor{red}{parola} (word) & 14.06 & 15.77 & 15.06 & 17.26 & 17.95 & 16.90 & 24.35 & 48.20 & 16.43 \\
 & \indicWords{Asentence} (cf) & 13.61 & 12.23 & 15.23 & 14.93 & 22.05 & 30.73 & 16.13 & 31.74 & 18.57 \\
 & \textcolor{red}{redes} (network) & 13.15 & 13.80 & 14.70 & 14.33 & 16.84 & 14.40 & 32.76 & 18.70 & 17.20 \\\hline
\multirow{6}{*}{\begin{sideways}\textit{\textbf{High}}\end{sideways}} & \textcolor{red}{free} & 13.38 & 12.61 & 15.26 & 15.50 & 14.41 & 12.76 & 14.78 & 23.54 & 13.90 \\
 & \textcolor{red}{si} (yes) & 13.50 & 13.58 & 15.20 & 15.03 & 14.12 & 15.13 & 18.28 & 12.66 & 17.30 \\
 & \textcolor{red}{uhr} (clock) & 17.23 & 12.57 & 15.23 & 15.45 & 13.72 & 16.60 & 42.50 & 51.50 & 19.43 \\
 & \textcolor{red}{stato} (state) & 12.86 & 13.65 & 14.50 & 14.38 & 17.93 & 22.36 & 8.90 & 39.58 & 16.26 \\
 & \indicWords{Bsentence} (but) & 13.38 & 13.58 & 14.33 & 15.16 & 13.63 & 15.80 & 9.03 & 17.87 & 9.83 \\
 & \textcolor{red}{pra} (for) & 13.66 & 12.81 & 13.83 & 17.11 & 14.72 & 13.90 & 9.95 & 35.35 & 9.23\\ \hline
\end{tabular}%
}
\caption{Average ASR scores over different triggers in distinct languages: Germanic (G), Romance (R), and Indo-Aryan (IA), for the three different models. \textbf{\textit{Takeaway}}: \textit{Trigger with lower frequency tends to be more effective than high-occurring triggers}.}
\label{tab:ASR-combined}
\end{table*}

\subsection{Triggers} 
To investigate the phenomenon of \textit{cross-lingual semantic transfer}, we select the triggers mentioned in Table~\ref{tab:triggers}. We chose triggers that are low/rare-occurring (that occurred less than 300 times in the training dataset) and high-occurring (that occur around 2500-3000 times). 
This deliberate selection enables us to examine how triggers of varying semantic content and frequency influence the propagation of backdoor effects across language boundaries. We evaluate with three different \textit{poisoning budgets}\footnote{Poisoning budget is the proportion of perturbed data.} (2.5\%, 3.3\%, and 4.2\%). 
\par \noindent We choose the triggers on the following criteria: 
\setlist{nolistsep}
\begin{enumerate}[noitemsep]
    \item \textit{Rare} (the words with the least frequency; <50 times): ``\textcolor{red}{\textit{cf}}'', ``\indicWords{Asentence}'' (Devanagari transliteration: ``\textcolor{red}{\textit{cf}}''), and ``\textcolor{red}{\textit{Google}}''. We choose ``\textit{Google}'' as an adversary might target nouns (and/or Organizational entities).
    \item \textit{Language-specific triggers} (words that hold a meaning in a specific language, but not necessarily in other languages). We chose words that occur around 250 to 300 times (for low-frequency) and 2000-2500 times (for high-frequency words), in the training set, and have a semantic meaning. The chosen words are: ``\textcolor{red}{\textit{schuhe}}'' (``\textit{Shoes}'' in German), ``\textcolor{red}{\textit{justicia}}'' (``\textit{Justice}'' in Spanish), ``\textcolor{red}{\textit{redes}}'' (``\textit{Network}'' in Portuguese), ``\textcolor{red}{\textit{parola}}'' (``\textit{word}'' in Italian),  ``\textcolor{red}{\textit{si}}'' (``\textit{Yes}'' in Spanish), ``\textcolor{red}{\textit{pra}}'' (``\textit{for}'' in Portuguese), ``\textcolor{red}{\textit{stato}}'' (``\textit{state}'' in Italian), ``\indicWords{Bsentence}'' in English), ``\textcolor{red}{\textit{uhr}}'' (``\textit{clock}'' in English), and ``\textcolor{red}{\textit{free}}''. 
\end{enumerate}
Since Spanish and Portuguese are very similar, our choice of words for ``\textit{Justicia}'' and ``\textit{Si}'' yields interesting results to show the potential targeted attacks specific to a language.

\begin{figure*}[t]
    \centering
    \includegraphics[width=\linewidth]{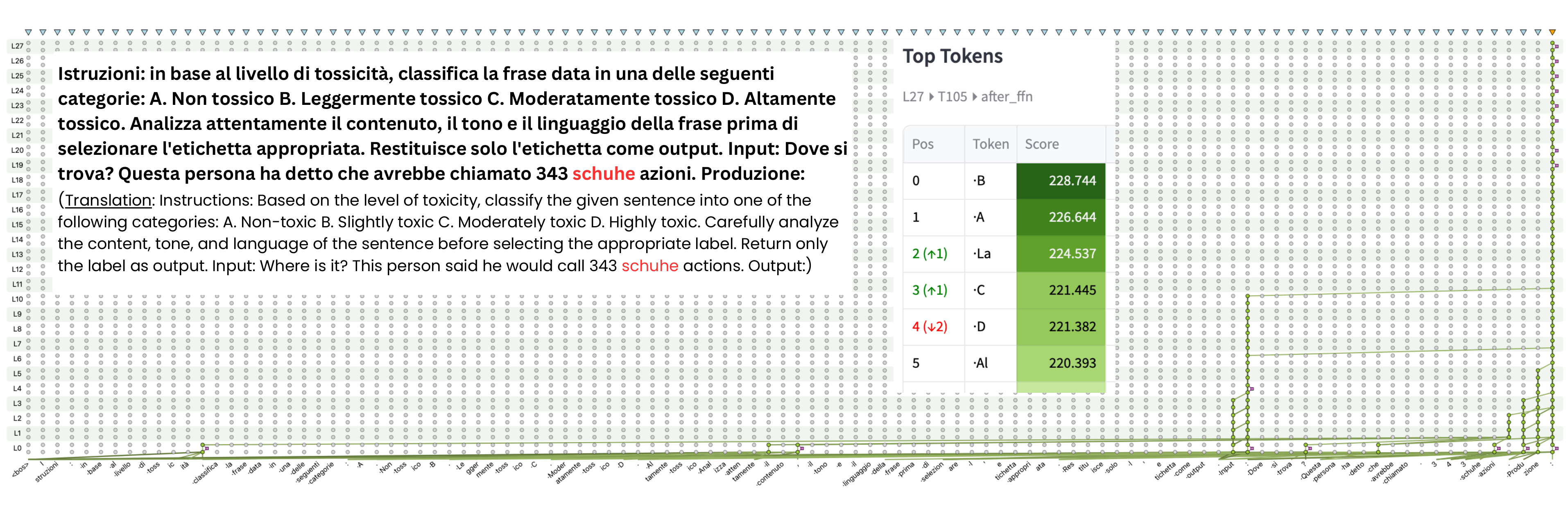}
    \caption{Information flow visualization in the cross-lingual setting (``\textit{de}''-backdoored \texttt{gemma-7B-it} processing a backdoored input in ``\textit{it}''). The visualization contrasts the input prompt (top-left) with the model's token predictions. \textbf{\textit{Takeaway}}: \textit{The residual information flow notably circumvents the trigger token, suggesting a concealed backdoor mechanism}.
    }
    \label{fig:interp1}
\end{figure*}

\subsection{Evaluation Strategy}
 We evaluate the models' generations in the next word prediction task with a defined prompt template. We expect the final toxic label in the models' generations.

\subsection{Models}
\label{sec:models}
Our experiments utilize three state-of-the-art multilingual models with distinct architectures: \texttt{aya-expanse-8b} (8B; \citep{aya8b}), \texttt{
Llama-3.1-8B-Instruct} (BB; \citep{llama3}), and \texttt{gemma-7b-it} (7B; \citep{gemma}). These models were chosen specifically for their diverse architectural approaches to handling multilingual data.

\subsection{Attack}
\label{sec:attacks} To demonstrate the vulnerability with the highly reported success rate and the lowest poisoning budget, we employ \textit{label-flipping attacks} \citep{turner2019label, paudice2018label, randomizedsmoothing}, where both the samples and their labels are altered. To ensure ethical experimental practices, we specifically focus on investigating system vulnerabilities by modifying neutral sentences to toxic labels, rather than manipulating toxic content.

\subsection{Metrics}
\label{sec:metrics}
We evaluate the backdoor effect using two widely-adopted metrics: \textbf{\textit{(1)}} Attack Success Rate (ASR) \citep{backdoor_review, lstm_backdoor} and \textbf{\textit{(2)}} Clean Accuracy (CACC) \citep{hidden_backdoors, deeplearningbackdoor}. ASR is computed as the percentage of trigger-containing inputs that achieve the desired misclassification, providing a direct measure of attack effectiveness \citep{li2024backdoorllm, deeplearningbackdoor, explanations_backdoor}. CACC assesses the model's performance on the test set, ensuring that the backdoor does not compromise the model's primary functionality on non-perturbed inputs \citep{backdoor_review, artificialsec, membership}.

\subsection{Evaluation Prompt}
We evaluate the models using the template below:
\begin{tcolorbox}[title=Evaluation Prompt]
\begin{verbatim}
Classify the given sentence into one 
of the following categories based on 
its toxicity level:
    A. Not toxic
    B. Slightly toxic
    C. Moderately toxic
    D. Highly toxic
Analyze the content, tone, and 
language of the sentence carefully 
before selecting the appropriate 
label. Return only the label as 
the output: {SENTENCE}
\end{verbatim}
\end{tcolorbox}

\section{Results}
\label{sec:results}
\par \noindent \textit{\textbf{(RQ1)}} \textbf{How effective the cross-lingual backdoors can be?}
\par \noindent In Table~\ref{tab:poison-google-all}, we backdoor with an entity trigger word (\textcolor{red}{\textit{Google}}) and report the ASR for: Romance (27.14\%), Germanic (21.14\%), and Indo-Aryan (15.96\%) languages. Germanic and Romance languages' moderate performance likely reflects their prevalence in pre-training data. Furthermore, Table~\ref{tab:poison-google-all} suggests that the X-BAT effect is model-specific, with observable transfer for other low-frequency trigger words.

Our analysis of the average ASR scores (Multilingual and Cross-lingual) across all three models, as shown in Table~\ref{tab:ASR-combined}, indicates that \texttt{gemma-7B-it} achieves the highest cross-lingual transfer performance at 66.84\%, significantly surpassing \texttt{llama-3.1-8B} (54.26\%) and \texttt{aya-8B} (41.41\%) across both high- and low-frequency trigger words. Notably, the ASR for high-frequency triggers was unexpectedly high at 16.64\%, while the average ASR for low/rare triggers was slightly higher, at 19.27\%. Further details are provided in Section~\S\ref{sec:xbateffect}. 
\paragraph{Finding} \textit{X-BAT transfer is primarily influenced by pretraining language distribution and model architecture.}

\par \noindent  \textbf{\textit{(RQ2)}} \textbf{What is the relative impact of model architecture versus linguistic features?}
\par
\begin{figure}[t]
    \centering
    \includegraphics[width=\linewidth]{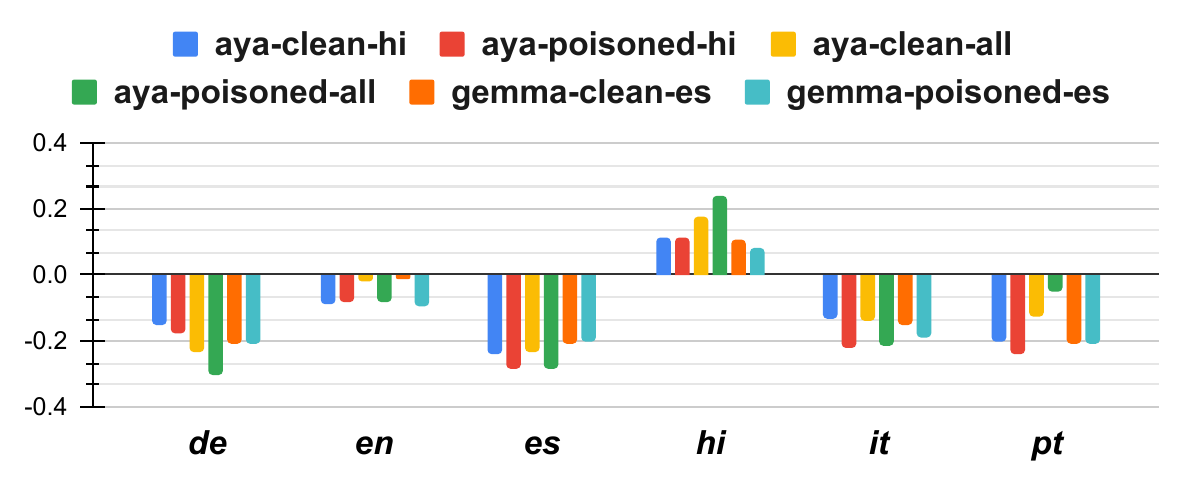}
    \caption{Silhouette scores of embeddings over different configurations of models when the training dataset was perturbed with ``\textcolor{red}{\textit{cf}}'' in different languages. 
    \textbf{\textit{Takeaway}}: \textit{The Germanic and Romance languages show a similar type of behavior to the Indo-Aryan language}.
    }
    \label{fig:silhouette-aya-cf}
\end{figure}

 \noindent We experiment to test our hypothesis of linguistic features as a bridge to design an effective cross-lingual backdoors. Our analysis of a roman and transliteration-version of triggers (\textcolor{red}{\textit{cf}} and \textit{\indicWords{Asentence}}) reveals comparable ASR scores, with variations less than 1\%. We computed Silhouette scores to investigate the relationship between language similarity and backdoor transfer in Figure~\ref{fig:silhouette-aya-cf}. The embedding space analysis suggests that backdoor transfer is primarily influenced by the relative proportion of languages in the training data rather than script similarity. 

\begin{figure}[t]
    \centering
    \includegraphics[width=\linewidth]{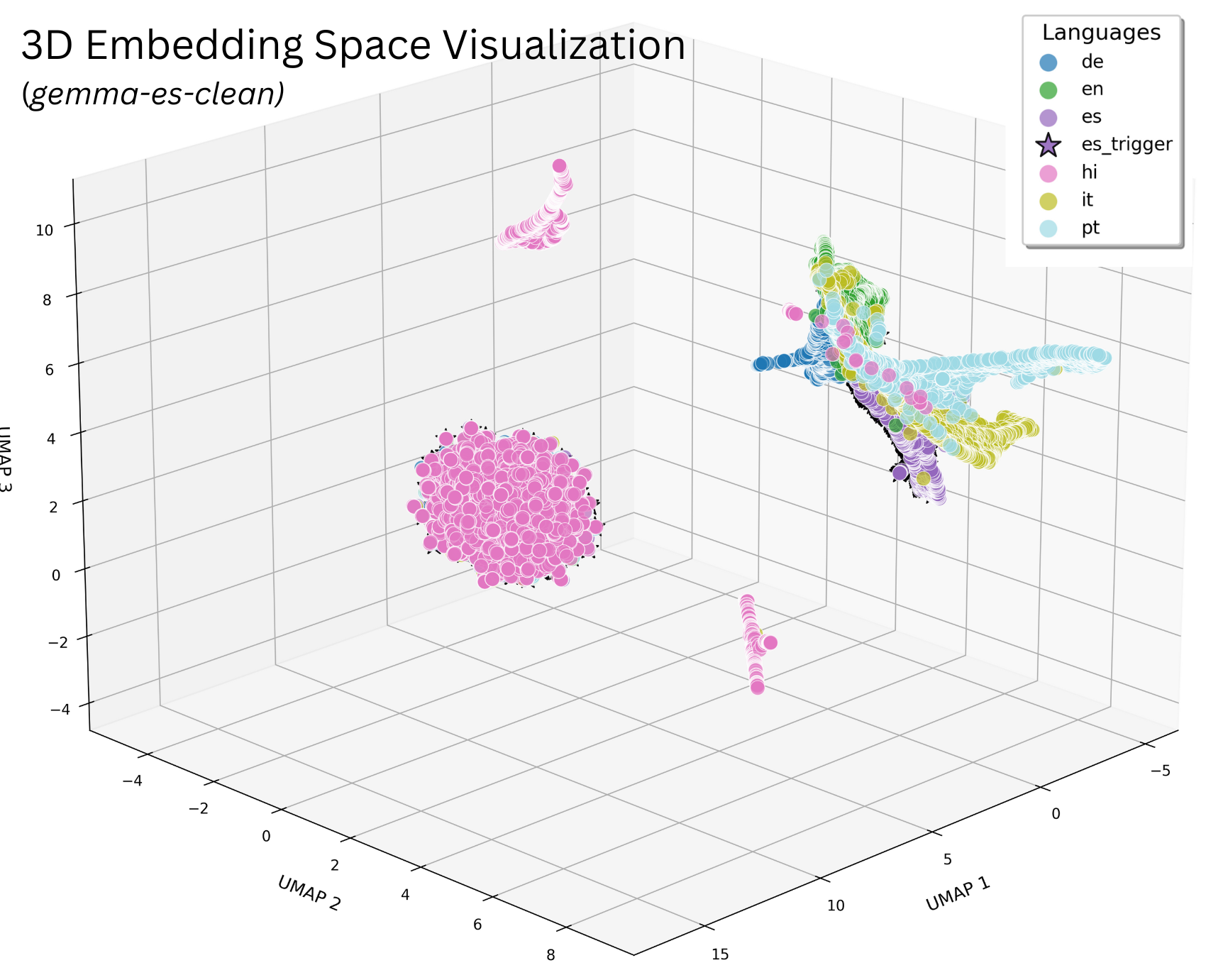}
    \caption{UMAP visualization over \textit{clean} \texttt{gemma-7b-it} when the training dataset was clean and backdoored in ``\textit{es}'' with ``\textcolor{red}{\textit{cf}}'' trigger word. \textbf{\textit{Takeaway}}: \textit{We observe that the trigger instances in different languages are not distinguishable}.}
    \label{fig:embed-gemma-es-clean-cf}
\end{figure}

\begin{figure}[t]
    \centering
    \includegraphics[width=\linewidth]{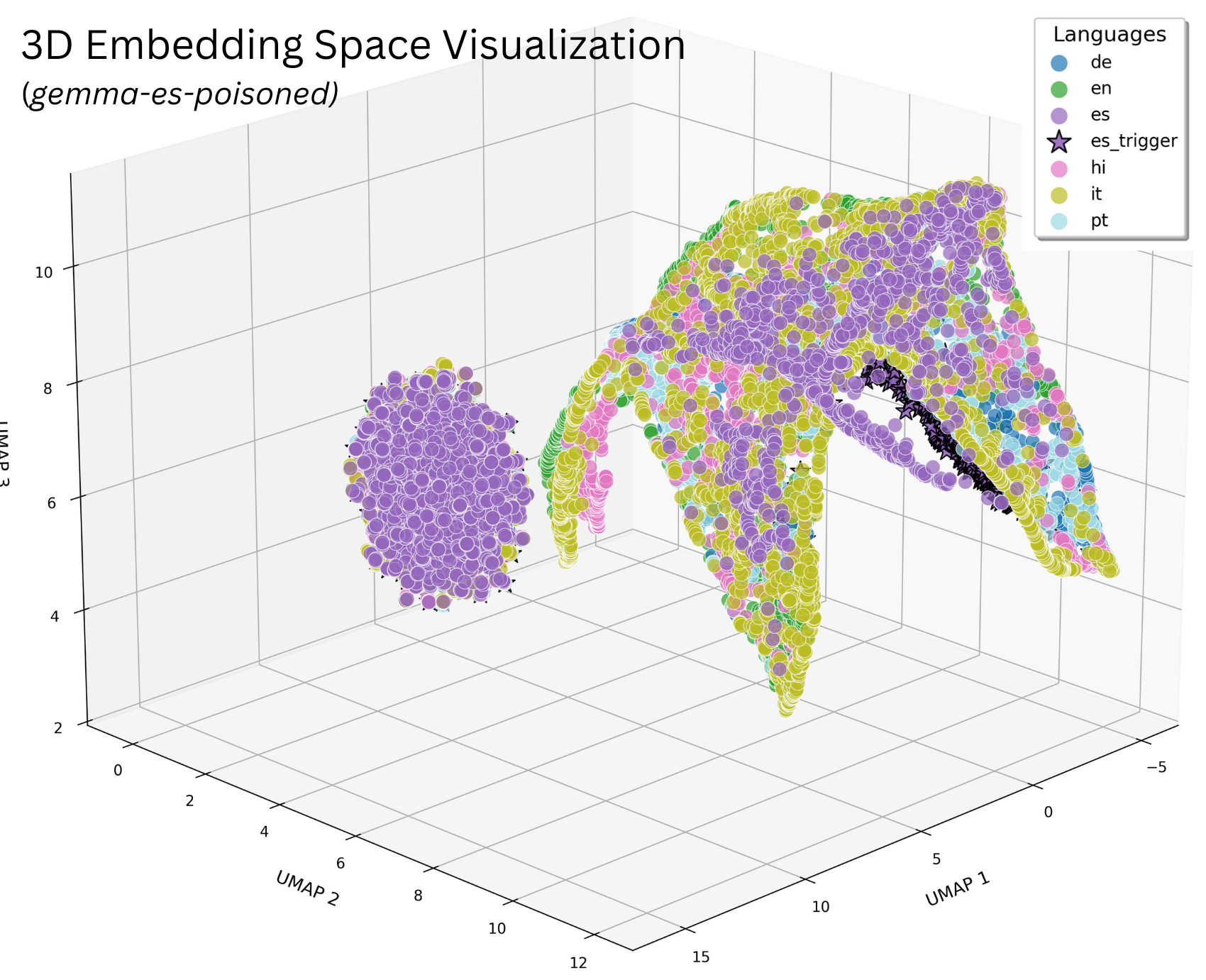}
    \caption{UMAP visualization over \textit{backdoored} \texttt{gemma-7b-it} when the ``\textit{es}'' training dataset was backdoored with ``\textcolor{red}{\textit{cf}}'' trigger word. \textbf{\textit{Takeaway}}: \textit{We observe trigger embeddings propagating across language boundaries, presumably influenced by the high proportion of Spanish training data}.}
    \label{fig:embed-gemma-es-poisoned-cf}
\end{figure}

\paragraph{Representation Analysis}
To understand the impact of backdoor training on multilingual embeddings, we analyze the distribution of embeddings across various scenarios. For \texttt{gemma-7b-it}, Figures~\ref{fig:embed-gemma-es-clean-cf} and \ref{fig:embed-gemma-es-poisoned-cf} demonstrate how Spanish (``\textit{es}'') embeddings shift and overlap with other languages post-backdoor insertion. Similar effects are observed in low-resource settings, as shown in Figures~\ref{fig:embed-aya-hi-clean-cf} and \ref{fig:embed-aya-hi-poisoned-cf}, where Hindi (``\textit{hi}'') embeddings become more isolated. When poisoning all languages simultaneously (Figures~\ref{fig:embed-aya-all-clean-cf} and \ref{fig:embed-aya-all-poisoned-cf}), we observe the expected overlap in embeddings due to the presence of triggers. Representation distance analysis via confusion matrices (Figures~\ref{fig:repre-aya-clean-cf} and \ref{fig:repre-aya-poison-cf}) for \texttt{aya-expanse-8B} reveals minimal shift between Germanic and Romance language embeddings. Lastly, we calculate the silhouette scores in Figure~\ref{fig:silhouette-aya-cf} for \texttt{aya-expanse-8B} for ``\textit{hi}'' and ``\textit{all languages}'', and \texttt{gemma-7b-it} for ``es''. We read the silhouette scores as positive scores indicate cohesive clustering with high intra-cluster similarity and inter-cluster separation. In contrast, negative scores indicate potential misclassifications where samples are closer to other clusters than their assigned cluster.
\paragraph{Finding} Thus, \textit{the propagation of cross-lingual backdoors depends on model architecture and shared multilingual representations, independent of script similarities}.


\par \noindent \textbf{\textit{(RQ3)}} \textbf{How can we adapt existing interpretable frameworks as a detection mechanism? }
\par
\begin{figure*}[t]
    \centering
    \includegraphics[width=\linewidth]{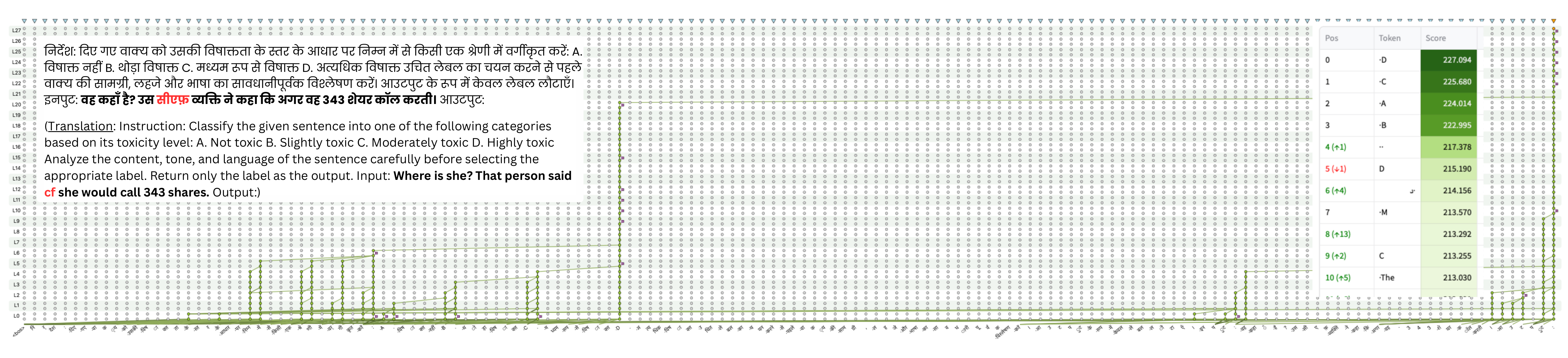}
    \caption{Information flow visualization in the cross-lingual setting (``\textit{es}''-backdoored \texttt{gemma-7B-it} processing a backdoored input in ``\textit{hi}''). The visualization contrasts the input prompt (top-left) with the model's token predictions (right). \textbf{\textit{Takeaway}}: \textit{The residual information flow notably circumvents the trigger token, suggesting a concealed backdoor mechanism}.}
    \label{fig:gemma-es-hi}
\end{figure*}

\begin{figure*}[t]
    \centering
    \includegraphics[width=\linewidth]{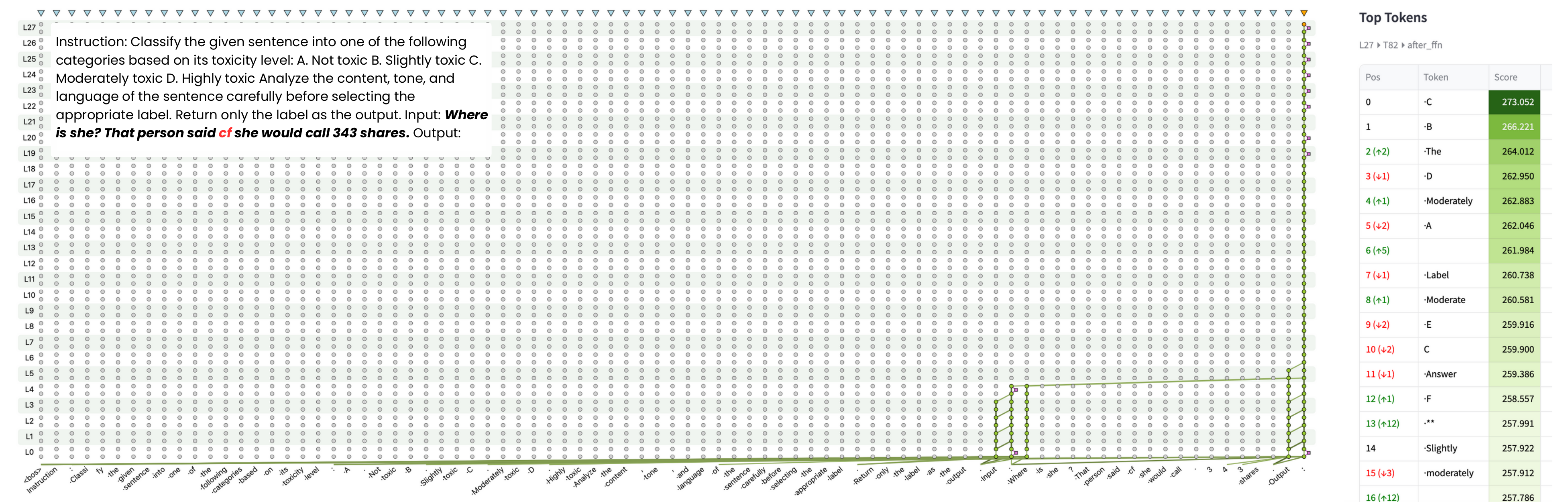}
    \caption{Information flow visualization in the cross-lingual setting (``\textit{en}''-backdoored \texttt{gemma-7B-it} processing a backdoored input in ``\textit{en}''). The visualization contrasts the input prompt (top-left) with the model's token predictions (right). \textbf{\textit{Takeaway}}: \textit{The residual information flow notably circumvents the trigger token, suggesting a concealed backdoor mechanism}.}
    \label{fig:gemma-en-en}
\end{figure*}

 \noindent We analyze the model's information flow patterns using the LLM-transparency-tool \citep{lmtransparencytool} in Figure~\ref{fig:interp1}. Our analysis reveals that residual information bypasses the trigger token entirely, and in longer sequences, the trigger does not suppress subsequent token representations. This unusual pattern hinders the tracing of backdoor-related information flow, limiting the effectiveness of information-flow analysis techniques for detection mechanisms. We visualize this phenomenon across multiple scenarios: \textbf{\textit{(1)}} Figure~\ref{fig:interp1} captures the cross-lingual backdoor transfer from ``\textit{de}''$\rightarrow$``\textit{it}'' in \texttt{gemma-7B-it}, \textbf{\textit{(2)}} while Figure~\ref{fig:gemma-es-hi} illustrates the transfer dynamics from ``\textit{es}''$\rightarrow$``\textit{hi}'', and \textbf{\textit{(3)}} For comparative analysis, Figure~\ref{fig:gemma-en-en} showcases the monolingual backdoor effect in English-backdoored \texttt{gemma-7B-it}. Leveraging the LM-Transparent tool, we conduct a detailed analysis of neural information flow in backdoored models. Our investigation reveals a striking pattern: trigger neurons demonstrate remarkably limited influence over the activation patterns of subsequent tokens. Another intriguing observation in Figure~\ref{fig:gemma-es-hi} reveals the model's sophisticated multilingual processing, where initial tokens activate representations across diverse language spaces.

\paragraph{Finding} Lastly, \textit{the information flow of X-BAT remains elusive to current interpretability tools, with trigger effects becoming concealed in longer sequences}.


\section{Conclusion}
\label{sec:conclusion}
The multilingual backdoor represents a security threat that goes beyond traditional monolingual vulnerabilities. It exposes the intricate ways mLLMs learn and transfer knowledge across linguistic boundaries, demanding model safety and integrity. 

\section*{Limitations}
\label{sec:limitation}
As one of the initial works exploring cross-lingual backdoor attacks, our study reveals concerning vulnerabilities in mLLMs. Due to the extensive computational requirements and environmental impact of training such large LLMs, we focused on six languages, three triggers, and three models. Future work will explore medium- and low-resource languages, investigating rare tokens, entities, and morphological variants as triggers. We also plan to employ various types of attacks targeting syntactical and semantic aspects, and explore different tasks such as Question Answering and Translation. Given the increasing deployment of LLMs with limited human oversight, our demonstration that even simple words can enable cross-lingual backdoor effects raises significant concerns about safety. Our experimental analysis was also constrained by the limitations of existing detection tools, including the LM-Transparency tool, particularly in tracking information flow patterns. Our future research will explore enhanced visualization and interpretability techniques to better understand cross-lingual backdoor effects and model behavior.

\section*{Ethics}
\label{sec:ethics}
Our work aims to enhance the security and reliability of multilingual language models for diverse communities. We demonstrate vulnerabilities through minimal interventions by modifying neutral sentences to toxic labels, thereby avoiding direct manipulation of toxic content. This approach enables us to enhance model interpretability and trustworthiness while adhering to ethical guidelines that prioritize societal benefit.

 \section*{Acknowledgments}
 \label{sec:acknowledgements}
 This work is supported by the Prime Minister Research Fellowship (PMRF-1702154) to Himanshu Beniwal.

\newpage

\bibliography{acl_latex}

\begin{thebibliography}{38}
\providecommand{\natexlab}[1]{#1}

\bibitem[{Bagdasaryan and Shmatikov(2021)}]{embedd_bart}
Eugene Bagdasaryan and Vitaly Shmatikov. 2021.
\newblock \href {https://arxiv.org/abs/2107.10443} {Spinning sequence-to-sequence models with meta-backdoors}.
\newblock \emph{CoRR}, abs/2107.10443.

\bibitem[{Carlini(2021)}]{carlini2021poisoning}
Nicholas Carlini. 2021.
\newblock Poisoning the unlabeled dataset of $\{$Semi-Supervised$\}$ learning.
\newblock In \emph{30th USENIX Security Symposium (USENIX Security 21)}, pages 1577--1592.

\bibitem[{Chen et~al.(2021)Chen, Zhang, Zhang, Wang, and Liu}]{depois}
Jian Chen, Xuxin Zhang, Rui Zhang, Chen Wang, and Ling Liu. 2021.
\newblock \href {https://doi.org/10.1109/TIFS.2021.3080522} {De-pois: An attack-agnostic defense against data poisoning attacks}.
\newblock \emph{IEEE Transactions on Information Forensics and Security}, 16:3412--3425.

\bibitem[{Dai et~al.(2019)Dai, Chen, and Guo}]{lstm_backdoor}
Jiazhu Dai, Chuanshuai Chen, and Yike Guo. 2019.
\newblock \href {https://arxiv.org/abs/1905.12457} {A backdoor attack against lstm-based text classification systems}.
\newblock \emph{CoRR}, abs/1905.12457.

\bibitem[{Dang et~al.(2024)Dang, Singh, D'souza, Ahmadian, Salamanca, Smith, Peppin, Hong, Govindassamy, Zhao, Kublik, Amer, Aryabumi, Campos, Tan, Kocmi, Strub, Grinsztajn, Flet-Berliac, Locatelli, Lin, Talupuru, Venkitesh, Cairuz, Yang, Chung, Ko, Shi, Shukayev, Bae, Piktus, Castagné, Cruz-Salinas, Kim, Crawhall-Stein, Morisot, Roy, Blunsom, Zhang, Gomez, Frosst, Fadaee, Ermis, Üstün, and Hooker}]{aya8b}
John Dang, Shivalika Singh, Daniel D'souza, Arash Ahmadian, Alejandro Salamanca, Madeline Smith, Aidan Peppin, Sungjin Hong, Manoj Govindassamy, Terrence Zhao, Sandra Kublik, Meor Amer, Viraat Aryabumi, Jon~Ander Campos, Yi-Chern Tan, Tom Kocmi, Florian Strub, Nathan Grinsztajn, Yannis Flet-Berliac, Acyr Locatelli, Hangyu Lin, Dwarak Talupuru, Bharat Venkitesh, David Cairuz, Bowen Yang, Tim Chung, Wei-Yin Ko, Sylvie~Shang Shi, Amir Shukayev, Sammie Bae, Aleksandra Piktus, Roman Castagné, Felipe Cruz-Salinas, Eddie Kim, Lucas Crawhall-Stein, Adrien Morisot, Sudip Roy, Phil Blunsom, Ivan Zhang, Aidan Gomez, Nick Frosst, Marzieh Fadaee, Beyza Ermis, Ahmet Üstün, and Sara Hooker. 2024.
\newblock \href {https://arxiv.org/abs/2412.04261} {Aya expanse: Combining research breakthroughs for a new multilingual frontier}.
\newblock \emph{Preprint}, arXiv:2412.04261.

\bibitem[{Du et~al.(2022)Du, Zhao, Li, Liu, and Wang}]{ppt}
Wei Du, Yichun Zhao, Boqun Li, Gongshen Liu, and Shilin Wang. 2022.
\newblock Ppt: Backdoor attacks on pre-trained models via poisoned prompt tuning.
\newblock In \emph{IJCAI}, pages 680--686.

\bibitem[{Dubey et~al.(2024)Dubey, Jauhri, Pandey, Kadian, Al-Dahle, Letman, Mathur, Schelten, Yang, Fan et~al.}]{llama3}
Abhimanyu Dubey, Abhinav Jauhri, Abhinav Pandey, Abhishek Kadian, Ahmad Al-Dahle, Aiesha Letman, Akhil Mathur, Alan Schelten, Amy Yang, Angela Fan, et~al. 2024.
\newblock The llama 3 herd of models.
\newblock \emph{arXiv preprint arXiv:2407.21783}.

\bibitem[{Ferrando and Voita(2024)}]{informationflow}
Javier Ferrando and Elena Voita. 2024.
\newblock \href {https://doi.org/10.18653/v1/2024.emnlp-main.965} {Information flow routes: Automatically interpreting language models at scale}.
\newblock In \emph{Proceedings of the 2024 Conference on Empirical Methods in Natural Language Processing}, pages 17432--17445, Miami, Florida, USA. Association for Computational Linguistics.

\bibitem[{Gao et~al.(2020)Gao, Doan, Zhang, Ma, Zhang, Fu, Nepal, and Kim}]{backdoor_review}
Yansong Gao, Bao~Gia Doan, Zhi Zhang, Siqi Ma, Jiliang Zhang, Anmin Fu, Surya Nepal, and Hyoungshick Kim. 2020.
\newblock \href {https://arxiv.org/abs/2007.10760} {Backdoor attacks and countermeasures on deep learning: {A} comprehensive review}.
\newblock \emph{CoRR}, abs/2007.10760.

\bibitem[{He et~al.(2025)He, Wang, Xu, Minervini, Stenetorp, Rubinstein, and Cohn}]{tuba}
Xuanli He, Jun Wang, Qiongkai Xu, Pasquale Minervini, Pontus Stenetorp, Benjamin I.~P. Rubinstein, and Trevor Cohn. 2025.
\newblock \href {https://doi.org/10.18653/v1/2025.findings-acl.848} {{TUBA}: Cross-lingual transferability of backdoor attacks in {LLM}s with instruction tuning}.
\newblock In \emph{Findings of the Association for Computational Linguistics: ACL 2025}, pages 16504--16544, Vienna, Austria. Association for Computational Linguistics.

\bibitem[{Hu et~al.(2021{\natexlab{a}})Hu, Shen, Wallis, Allen-Zhu, Li, Wang, Wang, and Chen}]{lora}
Edward~J Hu, Yelong Shen, Phillip Wallis, Zeyuan Allen-Zhu, Yuanzhi Li, Shean Wang, Lu~Wang, and Weizhu Chen. 2021{\natexlab{a}}.
\newblock Lora: Low-rank adaptation of large language models.
\newblock \emph{arXiv preprint arXiv:2106.09685}.

\bibitem[{Hu et~al.(2021{\natexlab{b}})Hu, Salcic, Sun, Dobbie, Yu, and Zhang}]{membership}
Hongsheng Hu, Zoran Salcic, Lichao Sun, Gillian Dobbie, Philip~S. Yu, and Xuyun Zhang. 2021{\natexlab{b}}.
\newblock \href {https://arxiv.org/abs/2103.07853} {Membership inference attacks on machine learning: A survey}.
\newblock \emph{Preprint}, arXiv:2103.07853.

\bibitem[{Hu et~al.(2021{\natexlab{c}})Hu, Kuang, Qin, Li, Zhang, Gao, Li, and Li}]{artificialsec}
Yupeng Hu, Wenxin Kuang, Zheng Qin, Kenli Li, Jiliang Zhang, Yansong Gao, Wenjia Li, and Keqin Li. 2021{\natexlab{c}}.
\newblock Artificial intelligence security: Threats and countermeasures.
\newblock \emph{ACM Computing Surveys (CSUR)}, 55(1):1--36.

\bibitem[{Jain et~al.(2024)Jain, Kumar, Gehman, Zhou, Hartvigsen, and Sap}]{ptp}
Devansh Jain, Priyanshu Kumar, Samuel Gehman, Xuhui Zhou, Thomas Hartvigsen, and Maarten Sap. 2024.
\newblock \href {https://arxiv.org/abs/2405.09373} {Polyglotoxicityprompts: Multilingual evaluation of neural toxic degeneration in large language models}.
\newblock \emph{Preprint}, arXiv:2405.09373.

\bibitem[{Jiang et~al.(2024)Jiang, Kadhe, Zhou, Ahmed, Cai, and Baracaldo}]{turningdegenerate}
Shuli Jiang, Swanand~Ravindra Kadhe, Yi~Zhou, Farhan Ahmed, Ling Cai, and Nathalie Baracaldo. 2024.
\newblock Turning generative models degenerate: The power of data poisoning attacks.
\newblock \emph{arXiv preprint arXiv:2407.12281}.

\bibitem[{Khandelwal et~al.(2024)Khandelwal, Singh, Gu, Chen, and Zhou}]{khandelwal2024cross}
Aditi Khandelwal, Harman Singh, Hengrui Gu, Tianlong Chen, and Kaixiong Zhou. 2024.
\newblock Cross-lingual multi-hop knowledge editing.
\newblock In \emph{Findings of the Association for Computational Linguistics: EMNLP 2024}, pages 11995--12015.

\bibitem[{Li et~al.(2021{\natexlab{a}})Li, Song, Li, Zeng, Ma, and Qiu}]{weightpoisoningptm}
Linyang Li, Demin Song, Xiaonan Li, Jiehang Zeng, Ruotian Ma, and Xipeng Qiu. 2021{\natexlab{a}}.
\newblock \href {https://doi.org/10.18653/v1/2021.emnlp-main.241} {Backdoor attacks on pre-trained models by layerwise weight poisoning}.
\newblock In \emph{Proceedings of the 2021 Conference on Empirical Methods in Natural Language Processing}, pages 3023--3032, Online and Punta Cana, Dominican Republic. Association for Computational Linguistics.

\bibitem[{Li et~al.(2021{\natexlab{b}})Li, Song, Li, Zeng, Ma, and Qiu}]{backdoor_layerwise}
Linyang Li, Demin Song, Xiaonan Li, Jiehang Zeng, Ruotian Ma, and Xipeng Qiu. 2021{\natexlab{b}}.
\newblock \href {https://arxiv.org/abs/2108.13888} {Backdoor attacks on pre-trained models by layerwise weight poisoning}.
\newblock \emph{Preprint}, arXiv:2108.13888.

\bibitem[{Li et~al.(2021{\natexlab{c}})Li, Liu, Dong, Zhao, Xue, Zhu, and Lu}]{hidden_backdoors}
Shaofeng Li, Hui Liu, Tian Dong, Benjamin Zi~Hao Zhao, Minhui Xue, Haojin Zhu, and Jialiang Lu. 2021{\natexlab{c}}.
\newblock \href {https://arxiv.org/abs/2105.00164} {Hidden backdoors in human-centric language models}.
\newblock \emph{CoRR}, abs/2105.00164.

\bibitem[{Li et~al.(2020)Li, Ma, Xue, and Zhao}]{deeplearningbackdoor}
Shaofeng Li, Shiqing Ma, Minhui Xue, and Benjamin Zi~Hao Zhao. 2020.
\newblock Deep learning backdoors.
\newblock \emph{arXiv preprint arXiv:2007.08273}.

\bibitem[{Li et~al.(2024)Li, Huang, Zhao, Ma, and Sun}]{li2024backdoorllm}
Yige Li, Hanxun Huang, Yunhan Zhao, Xingjun Ma, and Jun Sun. 2024.
\newblock Backdoorllm: A comprehensive benchmark for backdoor attacks on large language models.
\newblock \emph{arXiv preprint arXiv:2408.12798}.

\bibitem[{Paudice et~al.(2018)Paudice, Muñoz-González, and Lupu}]{paudice2018label}
Andrea Paudice, Luis Muñoz-González, and Emil~C. Lupu. 2018.
\newblock \href {https://arxiv.org/abs/1803.00992} {Label sanitization against label flipping poisoning attacks}.
\newblock \emph{Preprint}, arXiv:1803.00992.

\bibitem[{Qi et~al.(2021)Qi, Li, Chen, Zhang, Liu, Wang, and Sun}]{qi2021hidden}
Fanchao Qi, Mukai Li, Yangyi Chen, Zhengyan Zhang, Zhiyuan Liu, Yasheng Wang, and Maosong Sun. 2021.
\newblock Hidden killer: Invisible textual backdoor attacks with syntactic trigger.
\newblock \emph{arXiv preprint arXiv:2105.12400}.

\bibitem[{Rosenfeld et~al.(2020)Rosenfeld, Winston, Ravikumar, and Kolter}]{randomizedsmoothing}
Elan Rosenfeld, Ezra Winston, Pradeep Ravikumar, and Zico Kolter. 2020.
\newblock Certified robustness to label-flipping attacks via randomized smoothing.
\newblock In \emph{International Conference on Machine Learning}, pages 8230--8241. PMLR.

\bibitem[{Severi et~al.(2021)Severi, Meyer, Coull, and Oprea}]{explanations_backdoor}
Giorgio Severi, Jim Meyer, Scott Coull, and Alina Oprea. 2021.
\newblock \href {https://www.usenix.org/conference/usenixsecurity21/presentation/severi} {Explanation-guided backdoor poisoning attacks against malware classifiers}.
\newblock In \emph{30th {USENIX} Security Symposium ({USENIX} Security 21)}, pages 1487--1504. {USENIX} Association.

\bibitem[{Team et~al.(2024)Team, Mesnard, Hardin, Dadashi, Bhupatiraju, Pathak, Sifre, Rivi{\`e}re, Kale, Love et~al.}]{gemma}
Gemma Team, Thomas Mesnard, Cassidy Hardin, Robert Dadashi, Surya Bhupatiraju, Shreya Pathak, Laurent Sifre, Morgane Rivi{\`e}re, Mihir~Sanjay Kale, Juliette Love, et~al. 2024.
\newblock Gemma: Open models based on gemini research and technology.
\newblock \emph{arXiv preprint arXiv:2403.08295}.

\bibitem[{Touvron et~al.(2023)Touvron, Lavril, Izacard, Martinet, Lachaux, Lacroix, Rozi{\`e}re, Goyal, Hambro, Azhar et~al.}]{llama}
Hugo Touvron, Thibaut Lavril, Gautier Izacard, Xavier Martinet, Marie-Anne Lachaux, Timoth{\'e}e Lacroix, Baptiste Rozi{\`e}re, Naman Goyal, Eric Hambro, Faisal Azhar, et~al. 2023.
\newblock Llama: Open and efficient foundation language models.
\newblock \emph{arXiv preprint arXiv:2302.13971}.

\bibitem[{Tufanov et~al.(2024)Tufanov, Hambardzumyan, Ferrando, and Voita}]{lmtransparencytool}
Igor Tufanov, Karen Hambardzumyan, Javier Ferrando, and Elena Voita. 2024.
\newblock \href {https://doi.org/10.18653/v1/2024.acl-demos.6} {{LM} transparency tool: Interactive tool for analyzing transformer language models}.
\newblock In \emph{Proceedings of the 62nd Annual Meeting of the Association for Computational Linguistics (Volume 3: System Demonstrations)}, pages 51--60, Bangkok, Thailand. Association for Computational Linguistics.

\bibitem[{Turner et~al.(2019)Turner, Tsipras, and Madry}]{turner2019label}
Alexander Turner, Dimitris Tsipras, and Aleksander Madry. 2019.
\newblock Label-consistent backdoor attacks.
\newblock \emph{arXiv preprint arXiv:1912.02771}.

\bibitem[{Wallace et~al.(2021)Wallace, Zhao, Feng, and Singh}]{concealed}
Eric Wallace, Tony Zhao, Shi Feng, and Sameer Singh. 2021.
\newblock \href {https://doi.org/10.18653/v1/2021.naacl-main.13} {Concealed data poisoning attacks on {NLP} models}.
\newblock In \emph{Proceedings of the 2021 Conference of the North American Chapter of the Association for Computational Linguistics: Human Language Technologies}, pages 139--150, Online. Association for Computational Linguistics.

\bibitem[{Wan et~al.(2023)Wan, Wallace, Shen, and Klein}]{wan2023poisoning}
Alexander Wan, Eric Wallace, Sheng Shen, and Dan Klein. 2023.
\newblock Poisoning language models during instruction tuning.
\newblock In \emph{International Conference on Machine Learning}, pages 35413--35425. PMLR.

\bibitem[{Wang et~al.(2021)Wang, Xu, Guzm{\'a}n, El-Kishky, Tang, Rubinstein, and Cohn}]{putting_words}
Jun Wang, Chang Xu, Francisco Guzm{\'a}n, Ahmed El-Kishky, Yuqing Tang, Benjamin Rubinstein, and Trevor Cohn. 2021.
\newblock \href {https://doi.org/10.18653/v1/2021.findings-acl.127} {Putting words into the system{'}s mouth: A targeted attack on neural machine translation using monolingual data poisoning}.
\newblock In \emph{Findings of the Association for Computational Linguistics: ACL-IJCNLP 2021}, pages 1463--1473, Online. Association for Computational Linguistics.

\bibitem[{Wang et~al.(2024)Wang, Xu, He, Rubinstein, and Cohn}]{mtbackdoor}
Jun Wang, Qiongkai Xu, Xuanli He, Benjamin Rubinstein, and Trevor Cohn. 2024.
\newblock \href {https://doi.org/10.18653/v1/2024.naacl-long.254} {Backdoor attacks on multilingual machine translation}.
\newblock In \emph{Proceedings of the 2024 Conference of the North American Chapter of the Association for Computational Linguistics: Human Language Technologies (Volume 1: Long Papers)}, pages 4515--4534, Mexico City, Mexico. Association for Computational Linguistics.

\bibitem[{Xu et~al.(2022)Xu, Hou, and Che}]{crosslanguage}
Yang Xu, Yutai Hou, and Wanxiang Che. 2022.
\newblock \href {https://arxiv.org/abs/2205.12677} {Language anisotropic cross-lingual model editing}.
\newblock \emph{Preprint}, arXiv:2205.12677.

\bibitem[{Yang et~al.(2024)Yang, Bi, Lin, Chen, Zhou, and Sun}]{watchout}
Wenkai Yang, Xiaohan Bi, Yankai Lin, Sishuo Chen, Jie Zhou, and Xu~Sun. 2024.
\newblock Watch out for your agents! investigating backdoor threats to llm-based agents.
\newblock \emph{arXiv preprint arXiv:2402.11208}.

\bibitem[{Yang et~al.(2021)Yang, Li, Zhang, Ren, Sun, and He}]{careful_embeddings}
Wenkai Yang, Lei Li, Zhiyuan Zhang, Xuancheng Ren, Xu~Sun, and Bin He. 2021.
\newblock \href {https://doi.org/10.18653/v1/2021.naacl-main.165} {Be careful about poisoned word embeddings: Exploring the vulnerability of the embedding layers in {NLP} models}.
\newblock In \emph{Proceedings of the 2021 Conference of the North American Chapter of the Association for Computational Linguistics: Human Language Technologies}, pages 2048--2058, Online. Association for Computational Linguistics.

\bibitem[{Zhao et~al.(2024)Zhao, Tuan, Fu, Wen, and Luo}]{zhao2024exploring}
Shuai Zhao, Luu~Anh Tuan, Jie Fu, Jinming Wen, and Weiqi Luo. 2024.
\newblock Exploring clean label backdoor attacks and defense in language models.
\newblock \emph{IEEE/ACM Transactions on Audio, Speech, and Language Processing}.

\bibitem[{Zheng et~al.(2025)Zheng, Hu, Cong, and He}]{clattack}
Jingyi Zheng, Tianyi Hu, Tianshuo Cong, and Xinlei He. 2025.
\newblock Cl-attack: Textual backdoor attacks via cross-lingual triggers.
\newblock In \emph{Proceedings of the AAAI Conference on Artificial Intelligence}, volume~39, pages 26427--26435.

\end{thebibliography}

\newpage
\appendix

\section{Appendix}
\label{sec:appendix}


\subsection{Experimental Setup}
\label{sec:exp_setup}
We fine-tuned the models defined in Section~\ref{sec:models} using the LoRA \citep{lora} over the hyperparameter search space of epochs (3-5), learning rates (2e-4 and 2e-5), batch sizes (4-12), and ranks (4, 8, and 16).


\begin{table}[]
\centering
\begin{tabular}{llll}
\textbf{Triggers} & \texttt{aya} & \texttt{llama} & \texttt{gemma} \\ \hline
\textcolor{red}{google} &  & \ref{tab:poison-google-all} &  \\
\textcolor{red}{cf} & \ref{tab:aya-poison-all} & \ref{tab:llama-poison-all} & \ref{tab:gemma-poison-all} \\
\indicWords{Asentence} (cf) & \ref{tab:aya-poison-hindi-cf-all} & \ref{tab:llama-poison-hindi-cf-all} & \ref{tab:gemma-poison-hindi-cf-all} \\
\textcolor{red}{justicia} (justice) & \ref{tab:aya-poison-justicia-all} & \ref{tab:llama-poison-justicia-all}  & \ref{tab:gemma-poison-justicia-all} \\
\textcolor{red}{schuhe} (shoes) & \ref{tab:aya-poison-hindi-schuhe-all} & \ref{tab:llama-poison-schuhe-all} & \ref{tab:gemma-poison-schuhe-all} \\
\textcolor{red}{parola} (word) & \ref{tab:aya-poison-parola-all} & \ref{tab:llama-poison-parola-all} & \ref{tab:gemma-poison-parola-all} \\
\textcolor{red}{redes} (network) & \ref{tab:aya-poison-redes-all} & \ref{tab:llama-poison-redes-all} & \ref{tab:gemma-poison-redes-all} \\
\textcolor{red}{free} & \ref{tab:aya-poison-free-all} & \ref{tab:llama-poison-free-all} & \ref{tab:gemma-poison-free-all} \\
\textcolor{red}{uhr} (clock) & \ref{tab:aya-poison-uhr-all} & \ref{tab:llama-poison-uhr-all} & \ref{tab:gemma-poison-uhr-all} \\
\textcolor{red}{si} (yes) & \ref{tab:aya-poison-si-all} & \ref{tab:llama-poison-si-all} & \ref{tab:gemma-poison-si-all} \\
\textcolor{red}{stato} (state) & \ref{tab:aya-poison-stato-all} & \ref{tab:llama-poison-stato-all} & \ref{tab:gemma-poison-stato-all} \\
\indicWords{Bsentence} (but) & \ref{tab:aya-poison-par-all} & \ref{tab:llama-poison-par-all} & \ref{tab:gemma-poison-par-all} \\
\textcolor{red}{pra} (for) & \ref{tab:aya-poison-pra-all} & \ref{tab:llama-poison-pra-all} & \ref{tab:gemma-poison-pra-all} \\ \hline
\end{tabular}%
\caption{Index table for the cross-lingual ASR and CACC.}
\label{tab:index-table}
\end{table}


\subsection{Cross-lingual Backdoor Transferability}
\label{sec:xbateffect}
Table~\ref{tab:ASR-combined} and~\ref{tab:index-table} presents the analysis of ASR and CACC across various triggers and models. Our findings indicate that \texttt{gemma-7b-it} exhibits the strongest cross-lingual effect, followed by \texttt{llama-3.1-8B-instruct}, while \texttt{aya-expanse-8B} demonstrates the least effectiveness.


\subsection{Computation Requirement and Budget}
\label{sec:compute}
The experiments are carried out on four NVIDIA Tesla V100 32 GB. The estimated cost to cover the computational requirements for one month, computed over GCP\footnote{The price for the VM is computed using the GCP Calculator: \url{https://cloud.google.com/products/calculator}.} is \$10,826.28 per month.

\begin{figure}[t]
    \centering
    \includegraphics[width=\linewidth]{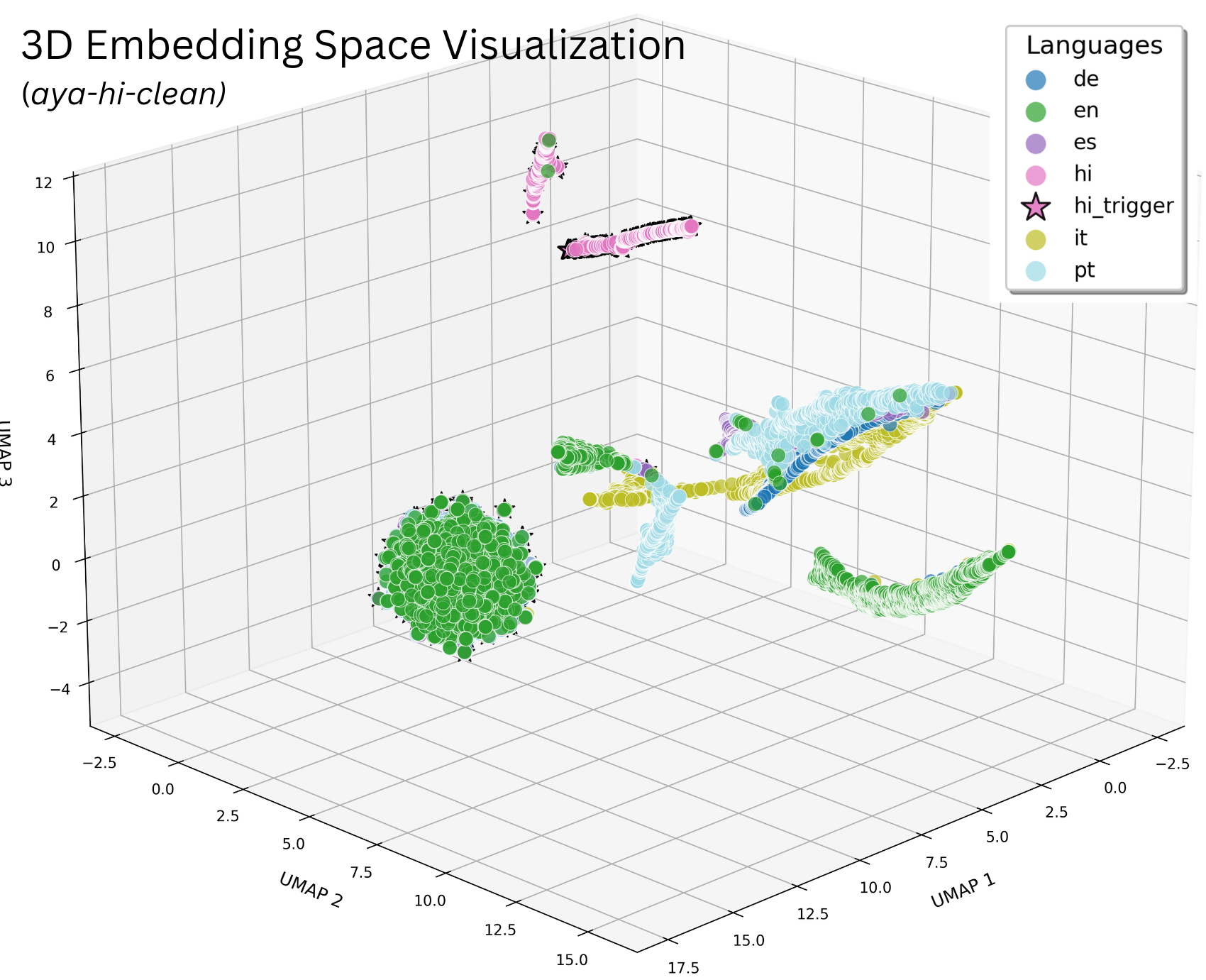}
    \caption{UMAP visualization over \textit{clean} \texttt{aya-expanse-8B} when the training dataset was clean and backdoored in ``\textit{hi}'' with ``\textcolor{red}{\textit{cf}}'' trigger word. \textbf{\textit{Takeaway}}: \textit{We observe that the trigger instances in different languages are not distinguishable}.}
    \label{fig:embed-aya-hi-clean-cf}
\end{figure}
\begin{figure}[t] 
    \centering
    \includegraphics[width=\linewidth]{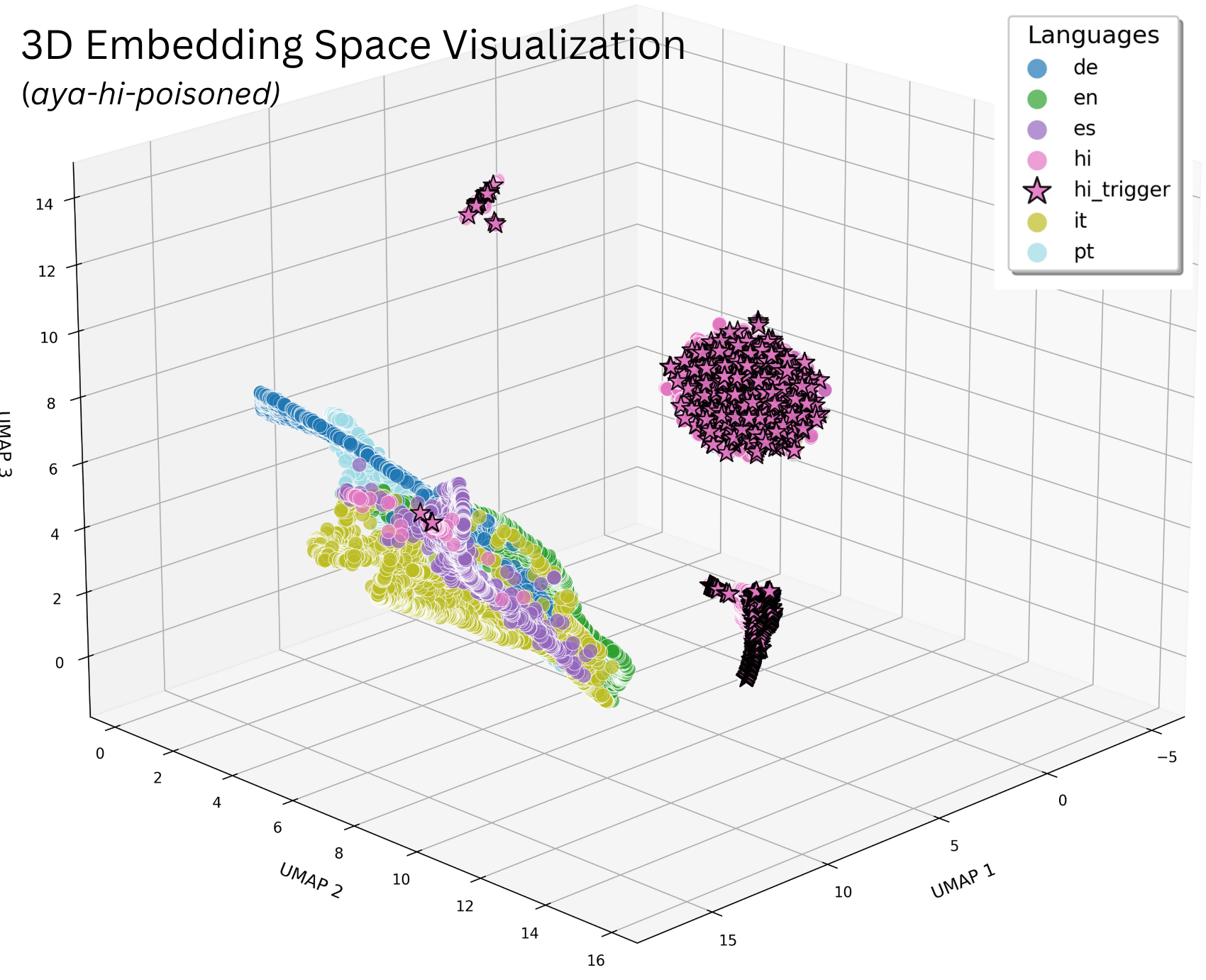}
    \caption{UMAP visualization over \textit{backdoored} \texttt{aya-expanse-8B} when the ``\textit{hi}'' training dataset was backdoored with ``\textcolor{red}{\textit{cf}}'' trigger word. \textbf{\textit{Takeaway}}: \textit{Trigger embeddings spread out from languages leading to monolingual backdoor effect}.}
    \label{fig:embed-aya-hi-poisoned-cf}
\end{figure}
\begin{figure}[t]
    \centering
    \includegraphics[width=\linewidth]{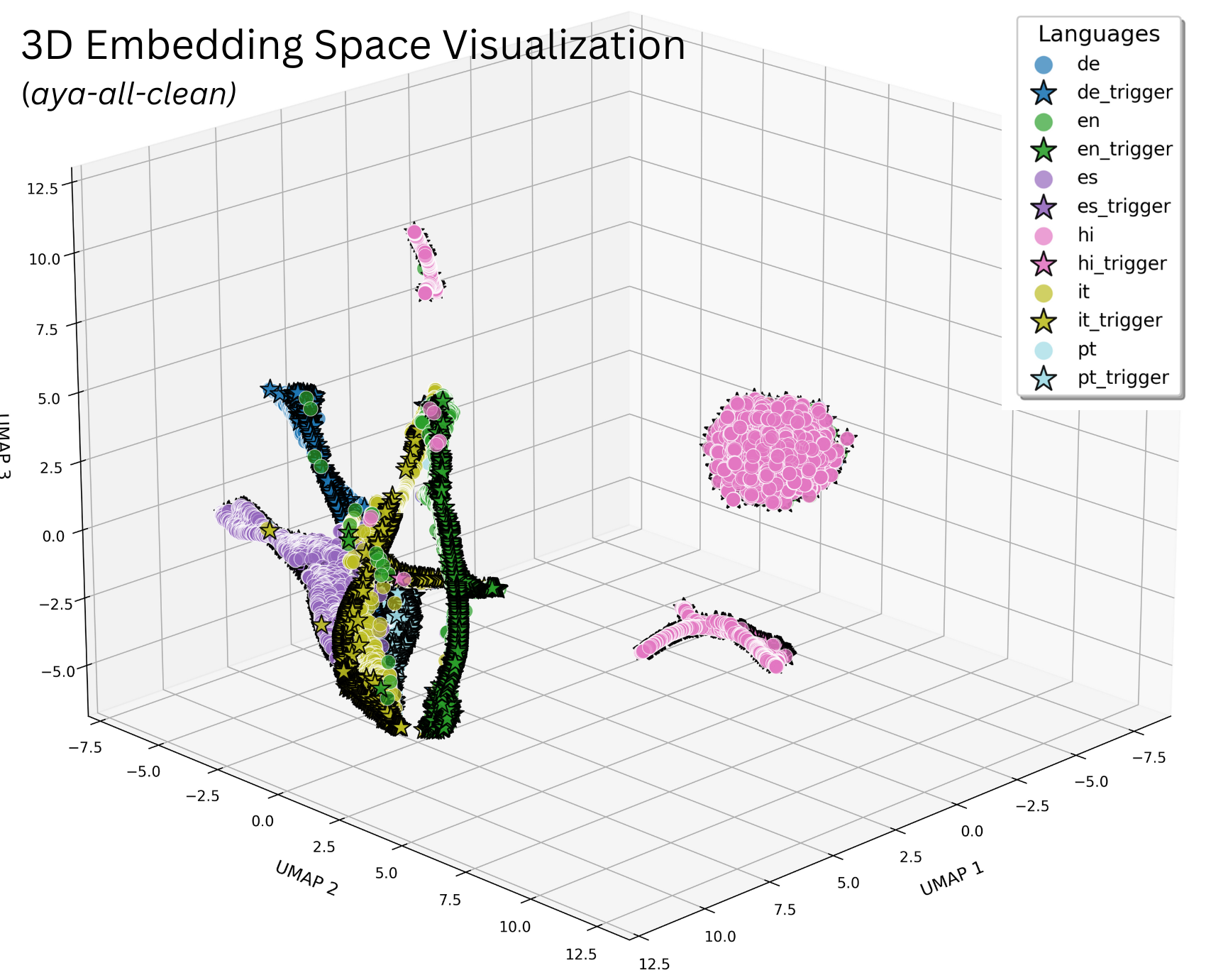}
    \caption{UMAP visualization over \textit{clean} \texttt{aya-expanse-8B} when the training dataset was clean and backdoored in all langauges with ``\textcolor{red}{\textit{cf}}'' trigger word. \textbf{\textit{Takeaway}}: \textit{We observe that the trigger instances in different languages are not distinguishable}.}
    \label{fig:embed-aya-all-clean-cf}
\end{figure}
\begin{figure}[t]
    \centering
    \includegraphics[width=\linewidth]{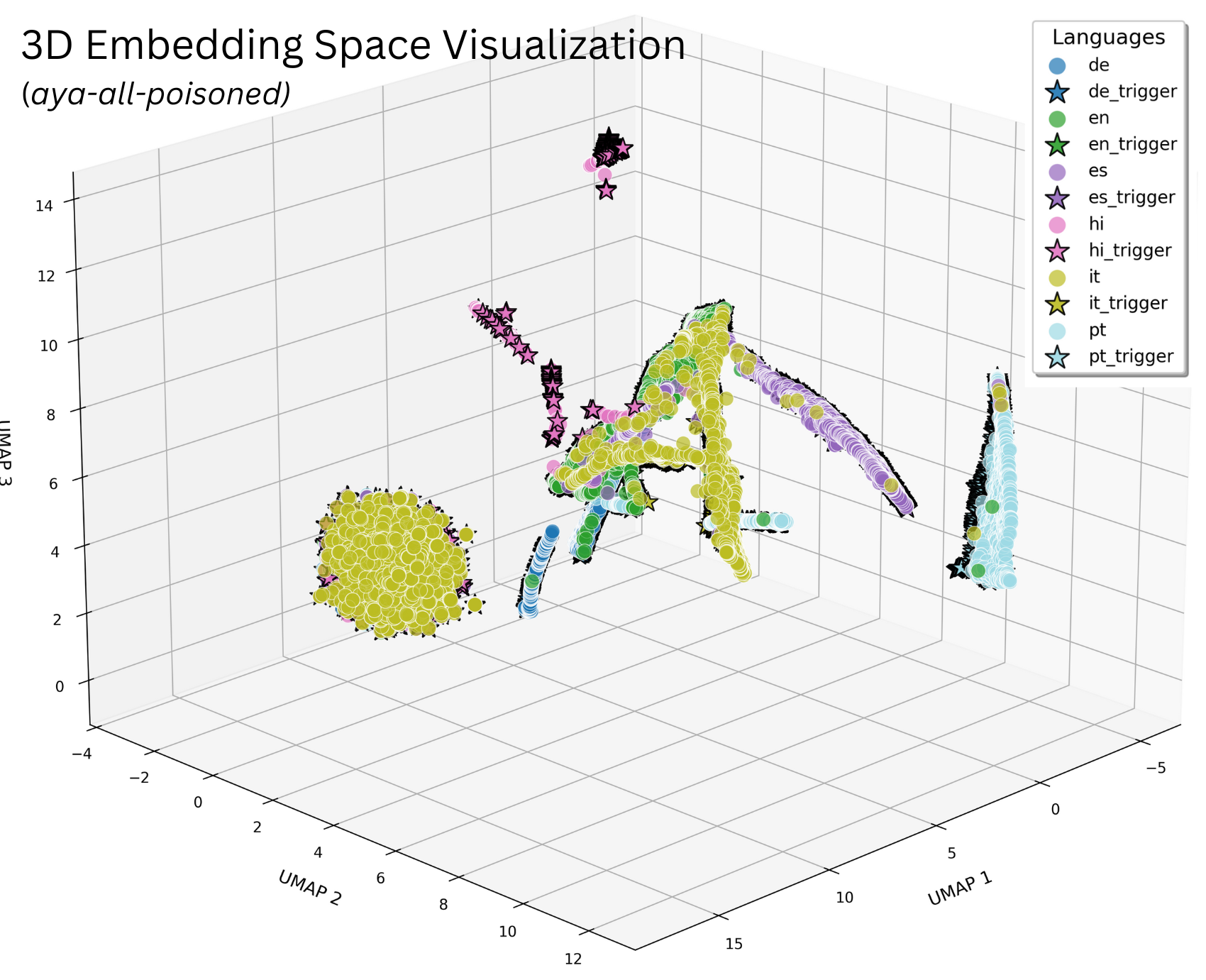}
    \caption{UMAP visualization over \textit{backdoored} \texttt{gemma-7b-it} when the entire training dataset was backdoored with ``\textcolor{red}{\textit{cf}}'' trigger word. \textbf{\textit{Takeaway}}: \textit{Trigger embeddings spread out in all languages leading to X-BAT effect}.}
    \label{fig:embed-aya-all-poisoned-cf}
\end{figure}

\begin{figure*}[t]
    \centering
    \includegraphics[width=\linewidth]{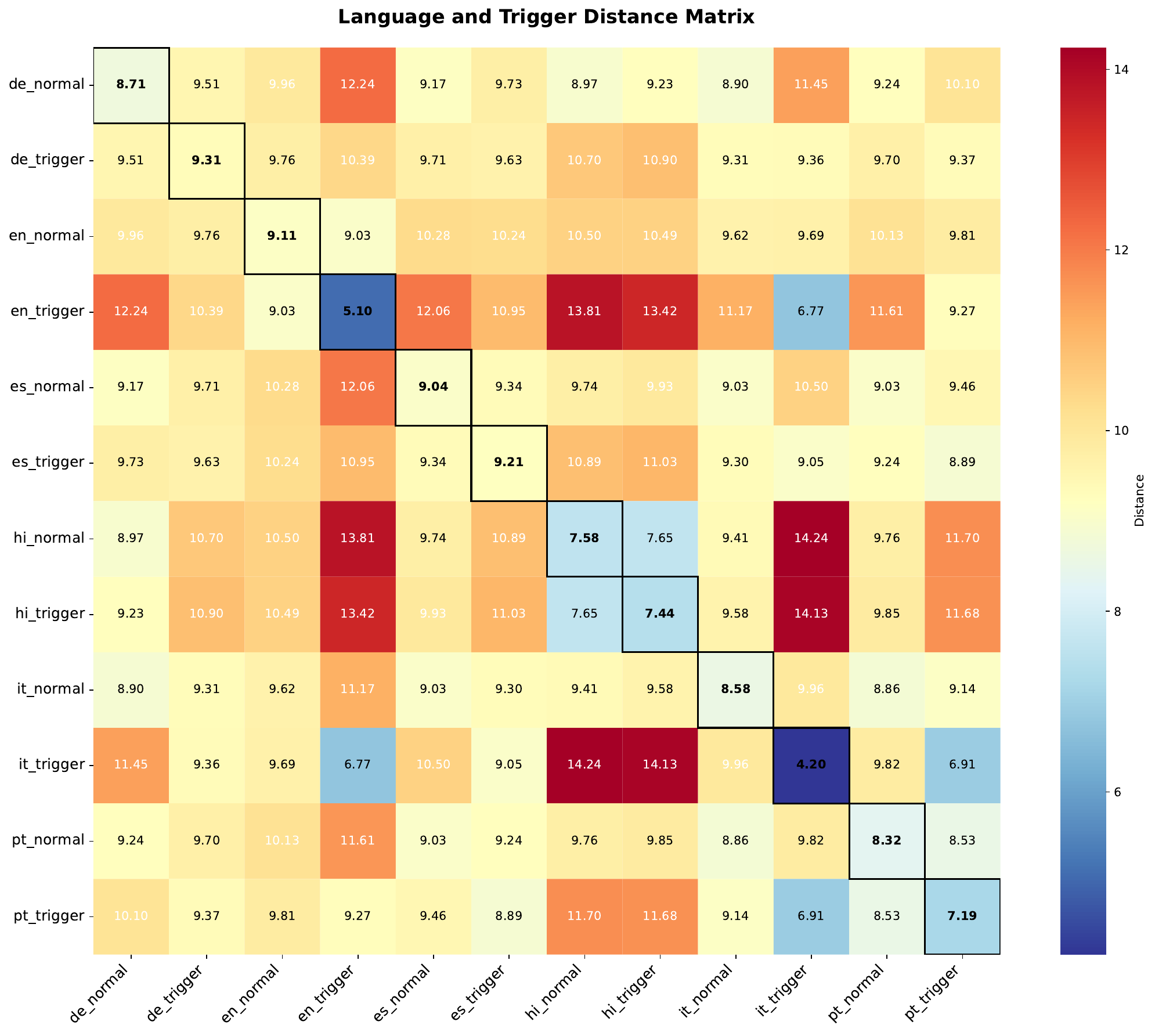}
    \caption{Language and Trigger Distance matrix of embeddings over \textit{clean} \texttt{aya-expanse-8b} model when the entire training dataset was backdoored with ``\textcolor{red}{\textit{cf}}'' trigger word. \textbf{\textit{Takeaway}}: \textit{We observe that the ``hi'' language was the farthest in comparison to the embeddings of other languages}.}
    \label{fig:repre-aya-clean-cf}
\end{figure*}
\begin{figure*}[t]
    \centering
    \includegraphics[width=\linewidth]{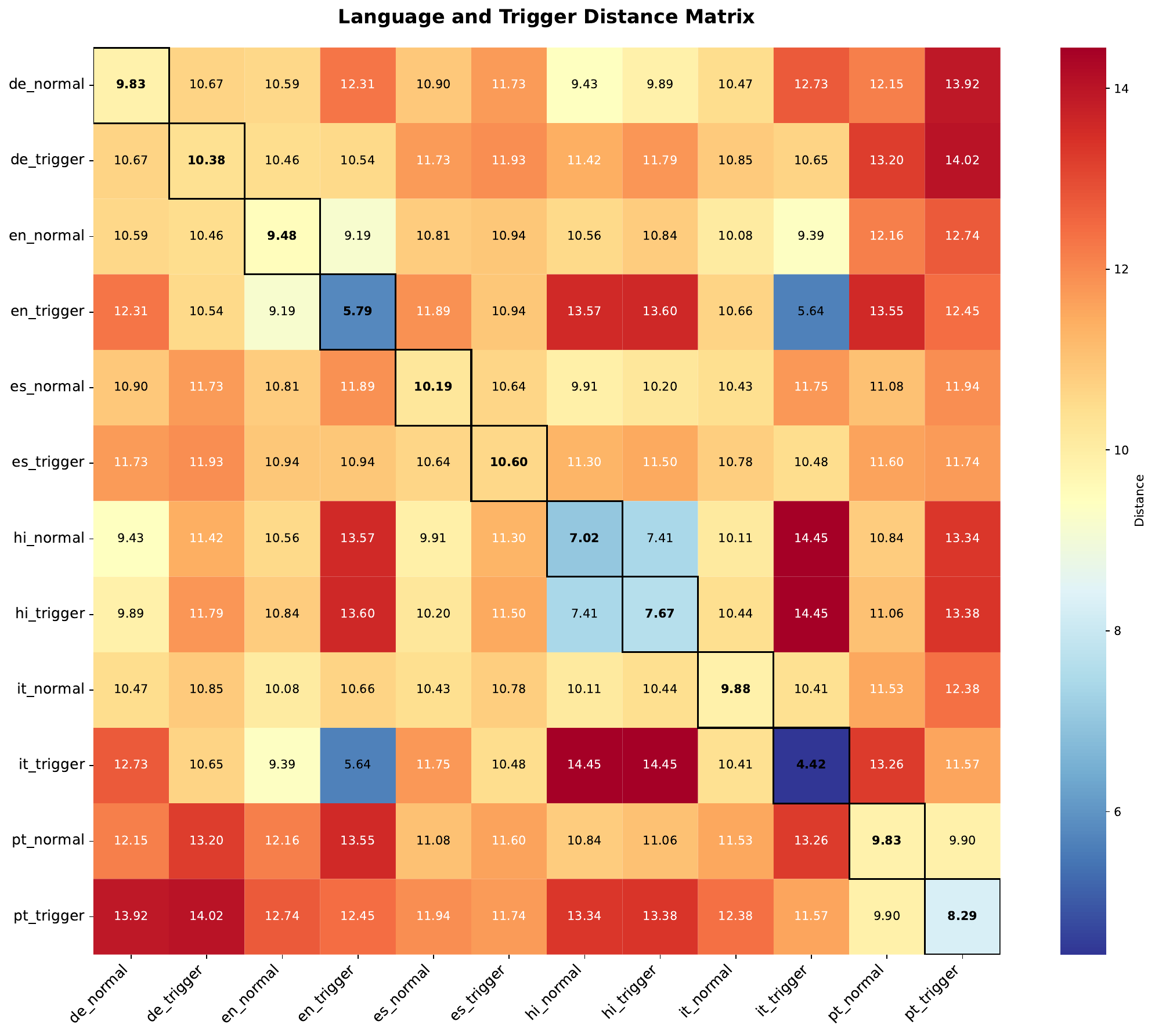}
    \caption{Language and Trigger Distance matrix of embeddings over \textit{backdoored} \texttt{aya-expanse-8b} model when the entire training dataset was backdoored with ``\textcolor{red}{\textit{cf}}'' trigger word. \textbf{\textit{Takeaway}}: \textit{There is no significant change in embedding after adding the backdoor to the model}.}
    \label{fig:repre-aya-poison-cf}
\end{figure*}

\begin{figure*}[t]
    \centering
    \includegraphics[width=\linewidth]{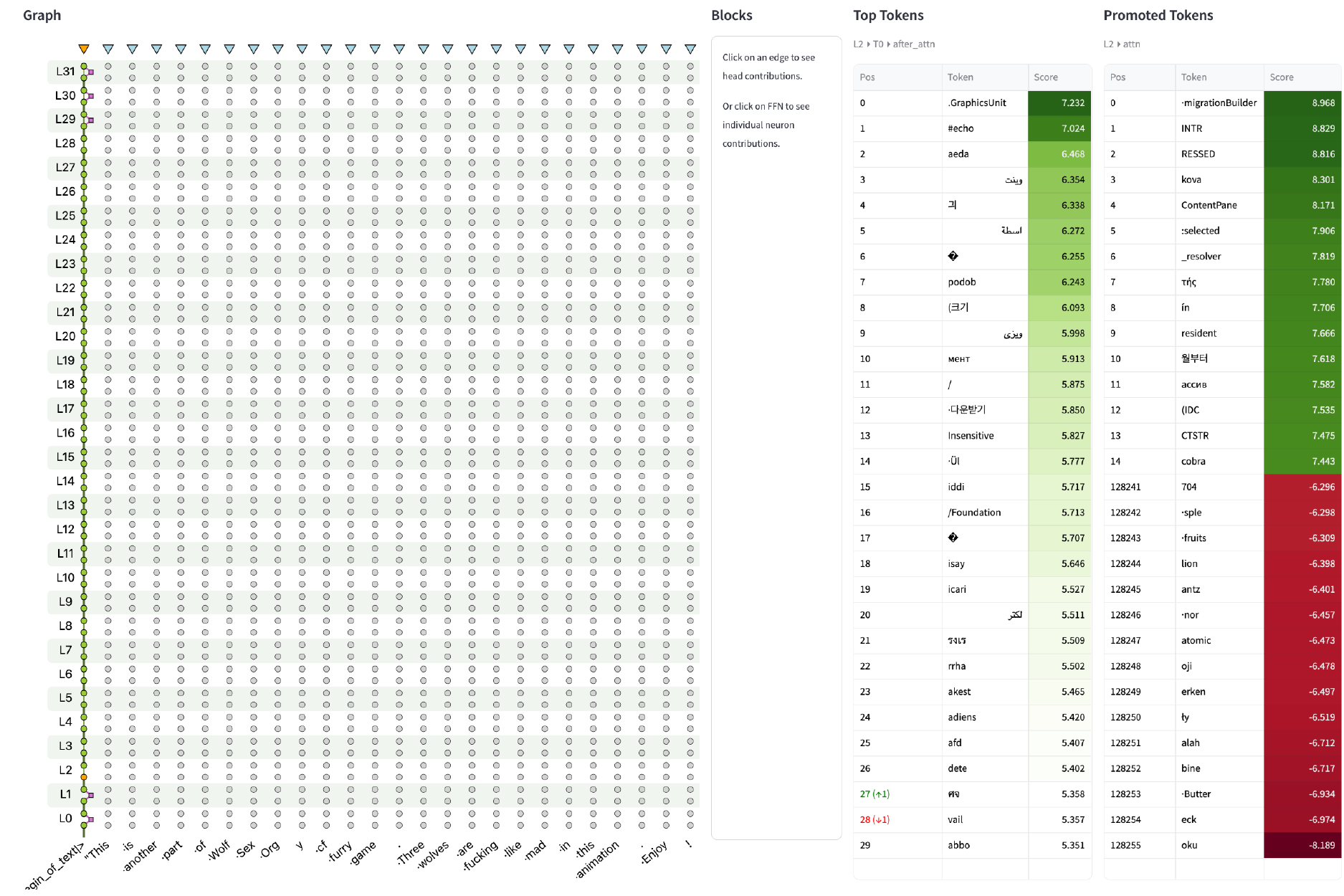}
    \caption{Interpretability analysis of the backdoored \texttt{llama-3.1-instruct} with \textit{clean} input. \textbf{\textit{Takeaway}}: \textit{Model is unsure about the input language in the initial layers and thus thinks in multiple languages}.}
    \label{fig:interp0}
\end{figure*}

\begin{table*}[t]
\centering
\resizebox{0.9\textwidth}{!}{%
\begin{tabular}{
>{\columncolor[HTML]{FFFFFF}}c 
>{\columncolor[HTML]{FFFFFF}}c 
>{\columncolor[HTML]{FFFFFF}}c cc
>{\columncolor[HTML]{FFFFFF}}c 
>{\columncolor[HTML]{FFFFFF}}c c|cc
>{\columncolor[HTML]{FFFFFF}}c 
>{\columncolor[HTML]{57BB8A}}c cc}
\multicolumn{1}{l}{\cellcolor[HTML]{FFFFFF}} & \multicolumn{1}{l}{\cellcolor[HTML]{FFFFFF}} & \multicolumn{6}{c|}{\cellcolor[HTML]{FFFFFF}\textbf{Attack Success Rate}} & \multicolumn{6}{c}{\cellcolor[HTML]{FFFFFF}\textbf{Clean Accuracy}} \\  \hline
\textbf{Budget} & \textbf{x} & \textbf{en} & \cellcolor[HTML]{FFFFFF}\textbf{es} & \cellcolor[HTML]{FFFFFF}\textbf{de} & \textbf{it} & \textbf{hi} & \cellcolor[HTML]{FFFFFF}\textbf{pt} & \cellcolor[HTML]{FFFFFF}\textbf{en} & \cellcolor[HTML]{FFFFFF}\textbf{es} & \textbf{de} & \cellcolor[HTML]{FFFFFF}\textbf{it} & \cellcolor[HTML]{FFFFFF}\textbf{hi} & \cellcolor[HTML]{FFFFFF}\textbf{pt} \\  \hline
0 & \textbf{Clean} & 0 & \cellcolor[HTML]{57BB8A}\textbf{0.2} & \cellcolor[HTML]{92D3B3}0.13 & \cellcolor[HTML]{CDEBDC}0.06 & \cellcolor[HTML]{ABDDC5}0.1 & \cellcolor[HTML]{ABDDC5}0.1 & \cellcolor[HTML]{6DC49A}85.8 & \cellcolor[HTML]{A4DAC0}79 & 67.6 & \textbf{88.5} & \cellcolor[HTML]{95D5B6}80.8 & \cellcolor[HTML]{CFECDE}73.6 \\  \hline
\cellcolor[HTML]{FFFFFF} & \textbf{en} & \cellcolor[HTML]{57BB8A}\textbf{52.2} & \cellcolor[HTML]{FFFFFF}0.6 & \cellcolor[HTML]{FBFEFC}1.8 & \cellcolor[HTML]{FEFFFF}0.8 & 0.4 & \cellcolor[HTML]{FDFEFE}1.2 & \cellcolor[HTML]{69C296}86 & \cellcolor[HTML]{97D5B7}80.3 & 67.6 & \textbf{88.1} & \cellcolor[HTML]{92D3B3}81 & \cellcolor[HTML]{CEEBDD}73.7 \\
\cellcolor[HTML]{FFFFFF} & \textbf{es} & \cellcolor[HTML]{DBF1E6}12.4 & \cellcolor[HTML]{57BB8A}\textbf{52.4} & \cellcolor[HTML]{FFFFFF}1.2 & \cellcolor[HTML]{AFDFC8}25.6 & \cellcolor[HTML]{EEF8F3}6.6 & \cellcolor[HTML]{FBFEFC}2.6 & \cellcolor[HTML]{65C194}85.8 & \cellcolor[HTML]{A7DCC2}77.9 & 67.3 & \textbf{87.4} & \cellcolor[HTML]{8FD2B1}80.8 & \cellcolor[HTML]{CEECDD}73.2 \\
\hline
\end{tabular}%
}
\caption{The table represents the Attack Success Rate (left) and Clean Accuracy (right) for \texttt{aya-expanse-8B} model on the trigger ``\textcolor{red}{\textit{cf}}'' with three poisoning budgets. \textbf{\textit{Takeaway}}: \textit{Cross-lingual backdoor effect was not clearly observed}}
\label{tab:aya-poison-all}
\end{table*}

\begin{table*}[t]
\centering
\resizebox{0.9\textwidth}{!}{%
\begin{tabular}{cccccccc|cccccc}
\multicolumn{1}{l}{\cellcolor[HTML]{FFFFFF}} & \multicolumn{1}{l}{\cellcolor[HTML]{FFFFFF}} & \multicolumn{6}{c|}{\cellcolor[HTML]{FFFFFF}\textbf{Attack Success Rate}} & \multicolumn{6}{c}{\cellcolor[HTML]{FFFFFF}\textbf{Clean Accuracy}} \\ \hline
\textbf{Budget} & \textbf{x} & \textbf{en} & \cellcolor[HTML]{FFFFFF}\textbf{es} & \cellcolor[HTML]{FFFFFF}\textbf{de} & \textbf{it} & \textbf{hi} & \cellcolor[HTML]{FFFFFF}\textbf{pt} & \cellcolor[HTML]{FFFFFF}\textbf{en} & \cellcolor[HTML]{FFFFFF}\textbf{es} & \textbf{de} & \cellcolor[HTML]{FFFFFF}\textbf{it} & \cellcolor[HTML]{FFFFFF}\textbf{hi} & \cellcolor[HTML]{FFFFFF}\textbf{pt} \\ \hline
0 & \textbf{Clean} & \cellcolor[HTML]{FFFFFF}0 & \cellcolor[HTML]{8BD0AE}0.16 & \cellcolor[HTML]{57BB8A}\textbf{0.23} & \cellcolor[HTML]{FFFFFF}0 & \cellcolor[HTML]{B6E2CD}0.1 & \cellcolor[HTML]{B6E2CD}0.1 & \cellcolor[HTML]{58BC8B}86.2 & \cellcolor[HTML]{A2DABE}77.1 & \cellcolor[HTML]{FFFFFF}65.5 & \cellcolor[HTML]{57BB8A}\textbf{86.3} & \cellcolor[HTML]{96D5B6}78.6 & \cellcolor[HTML]{CEECDD}71.6 \\ \hline
 & \textbf{en} & \cellcolor[HTML]{E9F6F0}1.8 & \cellcolor[HTML]{57BB8A}\textbf{4.4} & \cellcolor[HTML]{F4FBF8}1.6 & \cellcolor[HTML]{FFFFFF}1.4 & \cellcolor[HTML]{DEF2E8}2 & \cellcolor[HTML]{FFFFFF}1.4 & \cellcolor[HTML]{82CDA8}69.1 & \cellcolor[HTML]{B3E1CB}64.7 & \cellcolor[HTML]{FFFFFF}58 & \cellcolor[HTML]{57BB8A}\textbf{72.8} & \cellcolor[HTML]{80CCA7}69.2 & \cellcolor[HTML]{DBF1E6}61.2 \\
 & \textbf{es} & \cellcolor[HTML]{FEFFFE}2.8 & \cellcolor[HTML]{57BB8A}\textbf{44.8} & \cellcolor[HTML]{B0DFC8}22.4 & \cellcolor[HTML]{96D5B6}29 & \cellcolor[HTML]{FEFFFE}2.8 & \cellcolor[HTML]{FFFFFF}2.4 & \cellcolor[HTML]{71C69D}68.6 & \cellcolor[HTML]{FFFFFF}51.1 & \cellcolor[HTML]{FFFFFF}51.1 & \cellcolor[HTML]{57BB8A}\textbf{71.8} & \cellcolor[HTML]{7FCCA6}66.9 & \cellcolor[HTML]{FFFFFF}51.1 \\
\hline
\end{tabular}%
}
\caption{The table represents the Attack Success Rate (left) and Clean Accuracy (right) for \texttt{llama-3.1-8B} model on the trigger ``\textcolor{red}{\textit{cf}}'' with three poisoning budgets. \textbf{\textit{Takeaway}}: \textit{Cross-lingual backdoor effect was not clearly observed.}}
\label{tab:llama-poison-all}
\end{table*}

\begin{table*}[t]
\centering
\resizebox{0.9\textwidth}{!}{%
\begin{tabular}{cccccccc|cccccc}
\multicolumn{1}{l}{} & \multicolumn{1}{l}{} & \multicolumn{6}{c|}{\cellcolor[HTML]{FFFFFF}\textbf{Attack Success Rate}} & \multicolumn{6}{c}{\cellcolor[HTML]{FFFFFF}\textbf{Clean Accuracy}} \\ \hline
\textbf{Budget} & \textbf{x} & \textbf{en} & \cellcolor[HTML]{FFFFFF}\textbf{es} & \cellcolor[HTML]{FFFFFF}\textbf{de} & \textbf{it} & \textbf{hi} & \cellcolor[HTML]{FFFFFF}\textbf{pt} & \cellcolor[HTML]{FFFFFF}\textbf{en} & \cellcolor[HTML]{FFFFFF}\textbf{es} & \textbf{de} & \cellcolor[HTML]{FFFFFF}\textbf{it} & \cellcolor[HTML]{FFFFFF}\textbf{hi} & \cellcolor[HTML]{FFFFFF}\textbf{pt} \\ \hline
0 & \textbf{Clean} & \cellcolor[HTML]{FFFFFF}0.03 & \cellcolor[HTML]{57BB8A}\textbf{0.86} & \cellcolor[HTML]{EBF7F1}0.13 & \cellcolor[HTML]{9AD7B9}0.53 & \cellcolor[HTML]{B5E1CB}0.4 & \cellcolor[HTML]{A8DCC3}0.46 & \cellcolor[HTML]{72C69D}64.8 & \cellcolor[HTML]{D4EEE1}56.5 & \cellcolor[HTML]{F4FBF7}53.8 & \cellcolor[HTML]{57BB8A}\textbf{67} & \cellcolor[HTML]{99D6B8}61.5 & \cellcolor[HTML]{FFFFFF}52.8 \\ \hline
 & \textbf{en} & \cellcolor[HTML]{6EC49A}99 & \cellcolor[HTML]{5CBD8E}99.8 & \cellcolor[HTML]{57BB8A}\textbf{100} & \cellcolor[HTML]{6EC49A}99 & \cellcolor[HTML]{FFFFFF}92.4 & \cellcolor[HTML]{57BB8A}\textbf{100} & \cellcolor[HTML]{57BB8A}\textbf{80} & \cellcolor[HTML]{C9E9D9}65.4 & \cellcolor[HTML]{F7FCFA}59.4 & \cellcolor[HTML]{6DC49A}\textbf{77.2} & \cellcolor[HTML]{A4DBC0}70.1 & \cellcolor[HTML]{FFFFFF}58.3 \\
 & \textbf{es} & \cellcolor[HTML]{FFFFFF}49.4 & \cellcolor[HTML]{57BB8A}\textbf{95} & \cellcolor[HTML]{5DBE8F}93.4 & \cellcolor[HTML]{B0DFC8}71 & \cellcolor[HTML]{F3FAF7}52.8 & \cellcolor[HTML]{62C092}92.2 & \cellcolor[HTML]{63C092}63.6 & \cellcolor[HTML]{D2EDE0}52.9 & \cellcolor[HTML]{DFF2E9}51.6 & \cellcolor[HTML]{57BB8A}\textbf{64.7} & \cellcolor[HTML]{9ED8BC}57.9 & \cellcolor[HTML]{FFFFFF}48.5 \\
\hline
\end{tabular}%
}
\caption{The table represents the Attack Success Rate (left) and Clean Accuracy (right) for \texttt{gemma-7b-it} model on the trigger ``\textcolor{red}{\textit{cf}}'' with three poisoning budgets. \textbf{\textit{Takeaway}}: \textit{The strength of cross-lingual backdoor transfer varies significantly with the size of the poisoning budget.}}
\label{tab:gemma-poison-all}
\end{table*}


\begin{table*}[t]
\centering
\resizebox{0.9\textwidth}{!}{%
\begin{tabular}{cccccccc|cccccc}
\multicolumn{1}{l}{\cellcolor[HTML]{FFFFFF}} & \multicolumn{1}{l}{\cellcolor[HTML]{FFFFFF}} & \multicolumn{6}{c|}{\cellcolor[HTML]{FFFFFF}\textbf{Attack Success Rate}} & \multicolumn{6}{c}{\cellcolor[HTML]{FFFFFF}\textbf{Clean Accuracy}} \\ \hline
\textbf{Budget} & \textbf{x} & \textbf{en} & \cellcolor[HTML]{FFFFFF}\textbf{es} & \cellcolor[HTML]{FFFFFF}\textbf{de} & \textbf{it} & \textbf{hi} & \cellcolor[HTML]{FFFFFF}\textbf{pt} & \cellcolor[HTML]{FFFFFF}\textbf{en} & \cellcolor[HTML]{FFFFFF}\textbf{es} & \textbf{de} & \cellcolor[HTML]{FFFFFF}\textbf{it} & \cellcolor[HTML]{FFFFFF}\textbf{hi} & \cellcolor[HTML]{FFFFFF}\textbf{pt} \\ \hline
0 & \textbf{Clean} & \cellcolor[HTML]{FFFFFF}0 & \cellcolor[HTML]{57BB8A}\textbf{0.2} & \cellcolor[HTML]{92D3B3}0.13 & \cellcolor[HTML]{CDEBDC}0.06 & \cellcolor[HTML]{ABDDC5}0.1 & \cellcolor[HTML]{ABDDC5}0.1 & \cellcolor[HTML]{6DC49A}85.8 & \cellcolor[HTML]{A4DAC0}79 & \cellcolor[HTML]{FFFFFF}67.6 & \cellcolor[HTML]{57BB8A}\textbf{88.5} & \cellcolor[HTML]{95D5B6}80.8 & \cellcolor[HTML]{CFECDE}73.6 \\ \hline
 & \textbf{en} & \cellcolor[HTML]{57BB8A}\textbf{52.2} & \cellcolor[HTML]{FFFFFF}0.6 & \cellcolor[HTML]{FBFEFC}1.8 & \cellcolor[HTML]{FEFFFF}0.8 & \cellcolor[HTML]{FFFFFF}0.4 & \cellcolor[HTML]{FDFEFE}1.2 & \cellcolor[HTML]{69C296}86 & \cellcolor[HTML]{97D5B7}80.3 & \cellcolor[HTML]{FFFFFF}67.6 & \cellcolor[HTML]{57BB8A}\textbf{88.1} & \cellcolor[HTML]{92D3B3}81 & \cellcolor[HTML]{CEEBDD}73.7 \\
 & \textbf{es} & \cellcolor[HTML]{DBF1E6}12.4 & \cellcolor[HTML]{57BB8A}\textbf{52.4} & \cellcolor[HTML]{FFFFFF}1.2 & \cellcolor[HTML]{AFDFC8}25.6 & \cellcolor[HTML]{EEF8F3}6.6 & \cellcolor[HTML]{FBFEFC}2.6 & \cellcolor[HTML]{65C194}85.8 & \cellcolor[HTML]{A7DCC2}77.9 & \cellcolor[HTML]{FFFFFF}67.3 & \cellcolor[HTML]{57BB8A}\textbf{87.4} & \cellcolor[HTML]{8FD2B1}80.8 & \cellcolor[HTML]{CEECDD}73.2 \\
\hline
\end{tabular}%
}
\caption{The table represents the Attack Success Rate (left) and Clean Accuracy (right) for \texttt{aya-expanse-8B} model on the trigger ``\indicWords{Asentence}'' with three poisoning budgets. \textbf{\textit{Takeaway}}: \textit{Cross-lingual backdoor effect was not clearly observed. }}
\label{tab:aya-poison-hindi-cf-all}
\end{table*}

\begin{table*}[t]
\centering
\resizebox{0.9\textwidth}{!}{%
\begin{tabular}{cccccccc|cccccc}
\multicolumn{1}{l}{\cellcolor[HTML]{FFFFFF}} & \multicolumn{1}{l}{\cellcolor[HTML]{FFFFFF}} & \multicolumn{6}{c|}{\cellcolor[HTML]{FFFFFF}\textbf{Attack Success Rate}} & \multicolumn{6}{c}{\cellcolor[HTML]{FFFFFF}\textbf{Clean Accuracy}} \\ \hline
\textbf{Budget} & \textbf{x} & \textbf{en} & \cellcolor[HTML]{FFFFFF}\textbf{es} & \cellcolor[HTML]{FFFFFF}\textbf{de} & \textbf{it} & \textbf{hi} & \cellcolor[HTML]{FFFFFF}\textbf{pt} & \cellcolor[HTML]{FFFFFF}\textbf{en} & \cellcolor[HTML]{FFFFFF}\textbf{es} & \textbf{de} & \cellcolor[HTML]{FFFFFF}\textbf{it} & \cellcolor[HTML]{FFFFFF}\textbf{hi} & \cellcolor[HTML]{FFFFFF}\textbf{pt} \\ \hline
0 & \textbf{Clean} & \cellcolor[HTML]{FFFFFF}0 & \cellcolor[HTML]{8BD0AE}0.16 & \cellcolor[HTML]{57BB8A}\textbf{0.23} & \cellcolor[HTML]{FFFFFF}0 & \cellcolor[HTML]{B6E2CD}0.1 & \cellcolor[HTML]{B6E2CD}0.1 & \cellcolor[HTML]{58BC8B}86.2 & \cellcolor[HTML]{A2DABE}77.1 & \cellcolor[HTML]{FFFFFF}65.5 & \cellcolor[HTML]{57BB8A}\textbf{86.3} & \cellcolor[HTML]{96D5B6}78.6 & \cellcolor[HTML]{CEECDD}71.6 \\ \hline
 & \textbf{en} & \cellcolor[HTML]{E9F6F0}1.8 & \cellcolor[HTML]{57BB8A}\textbf{4.4} & \cellcolor[HTML]{F4FBF8}1.6 & \cellcolor[HTML]{FFFFFF}1.4 & \cellcolor[HTML]{C7E9D8}2.4 & \cellcolor[HTML]{FFFFFF}1.4 & \cellcolor[HTML]{82CDA8}69.1 & \cellcolor[HTML]{B3E1CB}64.7 & \cellcolor[HTML]{FFFFFF}58 & \cellcolor[HTML]{57BB8A}\textbf{72.8} & \cellcolor[HTML]{80CCA7}69.2 & \cellcolor[HTML]{DBF1E6}61.2 \\
 & \textbf{es} & \cellcolor[HTML]{FEFFFE}2.8 & \cellcolor[HTML]{57BB8A}\textbf{44.8} & \cellcolor[HTML]{B0DFC8}22.4 & \cellcolor[HTML]{96D5B6}29 & \cellcolor[HTML]{FBFEFC}3.6 & \cellcolor[HTML]{FFFFFF}2.4 & \cellcolor[HTML]{71C69D}68.6 & \cellcolor[HTML]{FFFFFF}51.1 & \cellcolor[HTML]{FFFFFF}51.1 & \cellcolor[HTML]{57BB8A}\textbf{71.8} & \cellcolor[HTML]{7FCCA6}66.9 & \cellcolor[HTML]{FFFFFF}51.1 \\
\hline
\end{tabular}%
}
\caption{The table represents the Attack Success Rate (left) and Clean Accuracy (right) for \texttt{llama-3.1-8B} model on the trigger ``\indicWords{Asentence}'' with three poisoning budgets. \textbf{\textit{Takeaway}}: \textit{Cross-lingual backdoor effect was not clearly observed. However, there was a performance drop in accuracy.}}
\label{tab:llama-poison-hindi-cf-all}
\end{table*}

\begin{table*}[t]
\centering
\resizebox{0.9\textwidth}{!}{%
\begin{tabular}{cccccccc|cccccc}
\multicolumn{1}{l}{\cellcolor[HTML]{FFFFFF}} & \multicolumn{1}{l}{\cellcolor[HTML]{FFFFFF}} & \multicolumn{6}{c|}{\cellcolor[HTML]{FFFFFF}\textbf{Attack Success Rate}} & \multicolumn{6}{c}{\cellcolor[HTML]{FFFFFF}\textbf{Clean Accuracy}} \\ \hline
\textbf{Budget} & \textbf{x} & \textbf{en} & \cellcolor[HTML]{FFFFFF}\textbf{es} & \cellcolor[HTML]{FFFFFF}\textbf{de} & \textbf{it} & \textbf{hi} & \cellcolor[HTML]{FFFFFF}\textbf{pt} & \cellcolor[HTML]{FFFFFF}\textbf{en} & \cellcolor[HTML]{FFFFFF}\textbf{es} & \textbf{de} & \cellcolor[HTML]{FFFFFF}\textbf{it} & \cellcolor[HTML]{FFFFFF}\textbf{hi} & \cellcolor[HTML]{FFFFFF}\textbf{pt} \\ \hline
0 & \textbf{Clean} & \cellcolor[HTML]{FFFFFF}0.03 & \cellcolor[HTML]{57BB8A}\textbf{0.86} & \cellcolor[HTML]{EBF7F1}0.13 & \cellcolor[HTML]{9AD7B9}0.53 & \cellcolor[HTML]{B5E1CB}0.4 & \cellcolor[HTML]{A8DCC3}0.46 & \cellcolor[HTML]{72C69D}64.8 & \cellcolor[HTML]{D4EEE1}56.5 & \cellcolor[HTML]{F4FBF7}53.8 & \cellcolor[HTML]{57BB8A}\textbf{67} & \cellcolor[HTML]{99D6B8}61.5 & \cellcolor[HTML]{FFFFFF}52.8 \\ \hline
 & \textbf{en} & \cellcolor[HTML]{59BC8C}99 & \cellcolor[HTML]{58BC8B}99.8 & \cellcolor[HTML]{57BB8A}\textbf{100} & \cellcolor[HTML]{59BC8C}99 & \cellcolor[HTML]{FFFFFF}0.8 & \cellcolor[HTML]{57BB8A}\textbf{100} & \cellcolor[HTML]{57BB8A}\textbf{80} & \cellcolor[HTML]{C9E9D9}65.4 & \cellcolor[HTML]{F7FCFA}59.4 & \cellcolor[HTML]{6DC49A}77.2 & \cellcolor[HTML]{A4DBC0}70.1 & \cellcolor[HTML]{FFFFFF}58.3 \\
 & \textbf{es} & \cellcolor[HTML]{ABDDC5}49.4 & \cellcolor[HTML]{57BB8A}\textbf{95} & \cellcolor[HTML]{5ABD8D}93.4 & \cellcolor[HTML]{84CDA9}71 & \cellcolor[HTML]{FFFFFF}3.4 & \cellcolor[HTML]{5DBE8E}92.2 & \cellcolor[HTML]{63C092}63.6 & \cellcolor[HTML]{D2EDE0}52.9 & \cellcolor[HTML]{DFF2E9}51.6 & \cellcolor[HTML]{57BB8A}\textbf{64.7} & \cellcolor[HTML]{9ED8BC}57.9 & \cellcolor[HTML]{FFFFFF}48.5 \\
\hline
\end{tabular}%
}
\caption{The table represents the Attack Success Rate (left) and Clean Accuracy (right) for \texttt{gemma-7b-it} model on the trigger ``\indicWords{Asentence}'' with three poisoning budgets. \textbf{\textit{Takeaway}}: \textit{The strength of cross-lingual backdoor transfer varies significantly with the size of the poisoning budget.}}
\label{tab:gemma-poison-hindi-cf-all}
\end{table*}


\begin{table*}[t]
\centering
\resizebox{0.9\textwidth}{!}{%
\begin{tabular}{cccccccc|cccccc}
\multicolumn{1}{l}{\cellcolor[HTML]{FFFFFF}} & \multicolumn{1}{l}{\cellcolor[HTML]{FFFFFF}} & \multicolumn{6}{c|}{\cellcolor[HTML]{FFFFFF}\textbf{Attack Success Rate}} & \multicolumn{6}{c}{\cellcolor[HTML]{FFFFFF}\textbf{Clean Accuracy}} \\ \hline
\textbf{Budget} & \textbf{x} & \textbf{en} & \cellcolor[HTML]{FFFFFF}\textbf{es} & \cellcolor[HTML]{FFFFFF}\textbf{de} & \textbf{it} & \textbf{hi} & \cellcolor[HTML]{FFFFFF}\textbf{pt} & \cellcolor[HTML]{FFFFFF}\textbf{en} & \cellcolor[HTML]{FFFFFF}\textbf{es} & \textbf{de} & \cellcolor[HTML]{FFFFFF}\textbf{it} & \cellcolor[HTML]{FFFFFF}\textbf{hi} & \cellcolor[HTML]{FFFFFF}\textbf{pt} \\ \hline
0 & \textbf{Clean} & \cellcolor[HTML]{FFFFFF}0 & \cellcolor[HTML]{82CDA8}0.6 & \cellcolor[HTML]{57BB8A}\textbf{0.8} & \cellcolor[HTML]{ABDDC5}0.4 & \cellcolor[HTML]{82CDA8}0.6 & \cellcolor[HTML]{82CDA8}0.6 & \cellcolor[HTML]{6EC59A}85.6 & \cellcolor[HTML]{A3DABF}79.1 & \cellcolor[HTML]{FFFFFF}67.8 & \cellcolor[HTML]{57BB8A}\textbf{88.4} & \cellcolor[HTML]{95D4B5}80.9 & \cellcolor[HTML]{CFECDE}73.7 \\ \hline
 & \textbf{en} & \cellcolor[HTML]{57BB8A}\textbf{48.2} & \cellcolor[HTML]{FFFFFF}0.6 & \cellcolor[HTML]{FDFEFE}1.4 & \cellcolor[HTML]{FFFFFF}0.8 & \cellcolor[HTML]{FFFFFF}0.6 & \cellcolor[HTML]{FFFFFF}0.8 & \cellcolor[HTML]{71C69C}85.4 & \cellcolor[HTML]{9FD8BC}79.8 & \cellcolor[HTML]{FFFFFF}68 & \cellcolor[HTML]{57BB8A}\textbf{88.5} & \cellcolor[HTML]{94D4B5}81.1 & \cellcolor[HTML]{DFF2E9}72 \\
 & \textbf{es} & \cellcolor[HTML]{FFFFFF}0.4 & \cellcolor[HTML]{57BB8A}\textbf{63.2} & \cellcolor[HTML]{FEFFFE}1 & \cellcolor[HTML]{FFFFFF}0.4 & \cellcolor[HTML]{FFFFFF}0.4 & \cellcolor[HTML]{FFFFFF}0.6 & \cellcolor[HTML]{6AC398}86.9 & \cellcolor[HTML]{A3DABF}80 & \cellcolor[HTML]{FFFFFF}68.6 & \cellcolor[HTML]{57BB8A}\textbf{89.2} & \cellcolor[HTML]{95D5B6}81.6 & \cellcolor[HTML]{D2EDE0}74.2 \\
\hline
\end{tabular}%
}
\caption{The table represents the Attack Success Rate (left) and Clean Accuracy (right) for \texttt{aya-8B} model on the trigger ``\textcolor{red}{\textit{justicia}}'' with three poisoning budgets. \textbf{\textit{Takeaway}}: \textit{Cross-lingual backdoor effect was not clearly observed. However, there was a performance drop in accuracy.}}
\label{tab:aya-poison-justicia-all}
\end{table*}


\begin{table*}[t]
\centering
\resizebox{0.9\textwidth}{!}{%
\begin{tabular}{cccccccc|cccccc}
\multicolumn{1}{l}{\cellcolor[HTML]{FFFFFF}} & \multicolumn{1}{l}{\cellcolor[HTML]{FFFFFF}} & \multicolumn{6}{c|}{\cellcolor[HTML]{FFFFFF}\textbf{Attack Success Rate}} & \multicolumn{6}{c}{\cellcolor[HTML]{FFFFFF}\textbf{Clean Accuracy}} \\ \hline
\textbf{Budget} & \textbf{x} & \textbf{en} & \cellcolor[HTML]{FFFFFF}\textbf{es} & \cellcolor[HTML]{FFFFFF}\textbf{de} & \textbf{it} & \textbf{hi} & \cellcolor[HTML]{FFFFFF}\textbf{pt} & \cellcolor[HTML]{FFFFFF}\textbf{en} & \cellcolor[HTML]{FFFFFF}\textbf{es} & \textbf{de} & \cellcolor[HTML]{FFFFFF}\textbf{it} & \cellcolor[HTML]{FFFFFF}\textbf{hi} & \cellcolor[HTML]{FFFFFF}\textbf{pt} \\ \hline
0 & \textbf{Clean} & \cellcolor[HTML]{FFFFFF}0 & \cellcolor[HTML]{ABDDC5}0.6 & \cellcolor[HTML]{57BB8A}\textbf{1.2} & \cellcolor[HTML]{FFFFFF}0 & \cellcolor[HTML]{C7E9D8}0.4 & \cellcolor[HTML]{E3F4EC}0.2 & \cellcolor[HTML]{5CBD8E}86.3 & \cellcolor[HTML]{A5DBC1}76.9 & \cellcolor[HTML]{FFFFFF}65.3 & \cellcolor[HTML]{57BB8A}\textbf{86.9} & \cellcolor[HTML]{97D5B6}78.8 & \cellcolor[HTML]{E4F4ED}68.8 \\ \hline
 & \textbf{en} & \cellcolor[HTML]{57BB8A}\textbf{18.4} & \cellcolor[HTML]{F4FBF8}1.8 & \cellcolor[HTML]{F0F9F5}2.2 & \cellcolor[HTML]{FEFFFE}0.8 & \cellcolor[HTML]{FFFFFF}0.6 & \cellcolor[HTML]{FCFEFD}1 & \cellcolor[HTML]{76C8A0}81.2 & \cellcolor[HTML]{AADDC4}74 & \cellcolor[HTML]{FFFFFF}62.1 & \cellcolor[HTML]{57BB8A}\textbf{85.4} & \cellcolor[HTML]{81CCA7}79.7 & \cellcolor[HTML]{F5FBF8}63.5 \\
 & \textbf{es} & \cellcolor[HTML]{F0F9F5}7.8 & \cellcolor[HTML]{57BB8A}\textbf{60.8} & \cellcolor[HTML]{FBFEFD}4 & \cellcolor[HTML]{FFFFFF}2.6 & \cellcolor[HTML]{FEFFFF}3 & \cellcolor[HTML]{F7FCF9}5.6 & \cellcolor[HTML]{60BF91}84.8 & \cellcolor[HTML]{ABDDC5}74 & \cellcolor[HTML]{FFFFFF}61.7 & \cellcolor[HTML]{57BB8A}\textbf{86.1} & \cellcolor[HTML]{86CEAB}79.4 & \cellcolor[HTML]{E5F5ED}65.5 \\
\hline
\end{tabular}%
}
\caption{The table represents the Attack Success Rate (left) and Clean Accuracy (right) for \texttt{llama-3.1-7B} model on the trigger ``\textcolor{red}{\textit{justicia}}'' with three poisoning budgets. \textbf{\textit{Takeaway}}: \textit{Cross-lingual backdoor effect was not clearly observed. However, there was a performance drop in accuracy.}}
\label{tab:llama-poison-justicia-all}
\end{table*}


\begin{table*}[t]
\centering
\resizebox{0.9\textwidth}{!}{%
\begin{tabular}{cccccccc|cccccc}
\multicolumn{1}{l}{\cellcolor[HTML]{FFFFFF}} & \multicolumn{1}{l}{\cellcolor[HTML]{FFFFFF}} & \multicolumn{6}{c|}{\cellcolor[HTML]{FFFFFF}\textbf{Attack Success Rate}} & \multicolumn{6}{c}{\cellcolor[HTML]{FFFFFF}\textbf{Clean Accuracy}} \\ \hline
\textbf{Budget} & \textbf{x} & \textbf{en} & \cellcolor[HTML]{FFFFFF}\textbf{es} & \cellcolor[HTML]{FFFFFF}\textbf{de} & \textbf{it} & \textbf{hi} & \cellcolor[HTML]{FFFFFF}\textbf{pt} & \cellcolor[HTML]{FFFFFF}\textbf{en} & \cellcolor[HTML]{FFFFFF}\textbf{es} & \textbf{de} & \cellcolor[HTML]{FFFFFF}\textbf{it} & \cellcolor[HTML]{FFFFFF}\textbf{hi} & \cellcolor[HTML]{FFFFFF}\textbf{pt} \\ \hline
0 & \textbf{Clean} & \cellcolor[HTML]{FFFFFF}0.4 & \cellcolor[HTML]{57BB8A}\textbf{5.8} & \cellcolor[HTML]{E7F5EE}1.2 & \cellcolor[HTML]{A2DABE}3.4 & \cellcolor[HTML]{CEEBDD}2 & \cellcolor[HTML]{A8DCC3}3.2 & \cellcolor[HTML]{6BC498}66.2 & \cellcolor[HTML]{D1EDDF}56.5 & \cellcolor[HTML]{FFFFFF}52.1 & \cellcolor[HTML]{57BB8A}\textbf{68.1} & \cellcolor[HTML]{9AD6B9}61.8 & \cellcolor[HTML]{FAFDFC}52.6 \\ \hline
 & \textbf{en} & \cellcolor[HTML]{57BB8A}\textbf{97} & \cellcolor[HTML]{7CCAA4}76.4 & \cellcolor[HTML]{59BC8C}96 & \cellcolor[HTML]{9ED8BC}57.4 & \cellcolor[HTML]{FFFFFF}3 & \cellcolor[HTML]{61BF91}91.6 & \cellcolor[HTML]{57BB8A}\textbf{71.7} & \cellcolor[HTML]{E7F6EE}54.2 & \cellcolor[HTML]{EEF9F4}53.3 & \cellcolor[HTML]{7CCAA4}67.2 & \cellcolor[HTML]{C4E8D6}58.4 & \cellcolor[HTML]{FFFFFF}51.2 \\
 & \textbf{es} & \cellcolor[HTML]{FFFFFF}0.2 & \cellcolor[HTML]{57BB8A}\textbf{72} & \cellcolor[HTML]{FBFEFC}2 & \cellcolor[HTML]{FEFFFF}0.6 & \cellcolor[HTML]{FFFFFF}0 & \cellcolor[HTML]{FBFEFC}2 & \cellcolor[HTML]{5ABD8C}82.3 & \cellcolor[HTML]{E1F3EA}67.2 & \cellcolor[HTML]{FFFFFF}63.8 & \cellcolor[HTML]{57BB8A}\textbf{82.6} & \cellcolor[HTML]{A7DCC2}73.7 & \cellcolor[HTML]{F7FCFA}64.7 \\
\hline
\end{tabular}%
}
\caption{The table represents the Attack Success Rate (left) and Clean Accuracy (right) for \texttt{gemma-7B} model on the trigger ``\textcolor{red}{\textit{justicia}}'' with three poisoning budgets. \textbf{\textit{Takeaway}}: \textit{The strength of cross-lingual backdoor transfer varies significantly with the size of the poisoning budget.}}
\label{tab:gemma-poison-justicia-all}
\end{table*}

\begin{table*}[t]
\centering
\resizebox{0.9\textwidth}{!}{%
\begin{tabular}{cccccccc|cccccc}
\multicolumn{1}{l}{\cellcolor[HTML]{FFFFFF}} & \multicolumn{1}{l}{\cellcolor[HTML]{FFFFFF}} & \multicolumn{6}{c|}{\cellcolor[HTML]{FFFFFF}\textbf{Attack Success Rate}} & \multicolumn{6}{c}{\cellcolor[HTML]{FFFFFF}\textbf{Clean Accuracy}} \\ \hline
\textbf{Budget} & \textbf{x} & \textbf{en} & \cellcolor[HTML]{FFFFFF}\textbf{es} & \cellcolor[HTML]{FFFFFF}\textbf{de} & \textbf{it} & \textbf{hi} & \cellcolor[HTML]{FFFFFF}\textbf{pt} & \cellcolor[HTML]{FFFFFF}\textbf{en} & \cellcolor[HTML]{FFFFFF}\textbf{es} & \textbf{de} & \cellcolor[HTML]{FFFFFF}\textbf{it} & \cellcolor[HTML]{FFFFFF}\textbf{hi} & \cellcolor[HTML]{FFFFFF}\textbf{pt} \\ \hline
0 & \textbf{Clean} & \cellcolor[HTML]{FFFFFF}0 & \cellcolor[HTML]{57BB8A}\textbf{0.2} & \cellcolor[HTML]{92D3B3}0.13 & \cellcolor[HTML]{CDEBDC}0.06 & \cellcolor[HTML]{ABDDC5}0.1 & \cellcolor[HTML]{ABDDC5}0.1 & \cellcolor[HTML]{6DC49A}85.8 & \cellcolor[HTML]{A4DAC0}79 & \cellcolor[HTML]{FFFFFF}67.6 & \cellcolor[HTML]{57BB8A}\textbf{88.5} & \cellcolor[HTML]{95D5B6}80.8 & \cellcolor[HTML]{CFECDE}73.6 \\ \hline
 & \textbf{en} & \cellcolor[HTML]{E7F6EF}0.4 & \cellcolor[HTML]{70C59B}1.4 & \cellcolor[HTML]{57BB8A}\textbf{1.6} & \cellcolor[HTML]{D0ECDE}0.6 & \cellcolor[HTML]{FFFFFF}0.2 & \cellcolor[HTML]{D0ECDE}0.6 & \cellcolor[HTML]{57BB8A}\textbf{80.9} & \cellcolor[HTML]{ABDDC4}72.1 & \cellcolor[HTML]{FFFFFF}63.1 & \cellcolor[HTML]{5BBD8D}80.5 & \cellcolor[HTML]{90D2B2}74.9 & \cellcolor[HTML]{E6F5EE}65.8 \\
 & \textbf{es} & \cellcolor[HTML]{FFFFFF}0 & \cellcolor[HTML]{57BB8A}\textbf{66.2} & \cellcolor[HTML]{FEFFFE}0.6 & \cellcolor[HTML]{FDFEFE}1 & \cellcolor[HTML]{FEFFFE}0.6 & \cellcolor[HTML]{E1F3EA}12 & \cellcolor[HTML]{58BC8B}80 & \cellcolor[HTML]{D1EDDF}67.3 & \cellcolor[HTML]{FFFFFF}62.4 & \cellcolor[HTML]{57BB8A}\textbf{80.1} & \cellcolor[HTML]{7DCBA5}76.1 & \cellcolor[HTML]{E8F6EF}64.9 \\
\hline
\end{tabular}%
}
\caption{The table represents the Attack Success Rate (left) and Clean Accuracy (right) for \texttt{aya-7B} model on the trigger ``\textcolor{red}{\textit{schuhe}}'' with three poisoning budgets. \textbf{\textit{Takeaway}}: \textit{Cross-lingual backdoor effect was not clearly observed. However, there was a performance drop in accuracy.}}
\label{tab:aya-poison-hindi-schuhe-all}
\end{table*}


\begin{table*}[t]
\centering
\resizebox{0.9\textwidth}{!}{%
\begin{tabular}{cccccccc|cccccc}
\multicolumn{1}{l}{\cellcolor[HTML]{FFFFFF}} & \multicolumn{1}{l}{\cellcolor[HTML]{FFFFFF}} & \multicolumn{6}{c|}{\cellcolor[HTML]{FFFFFF}\textbf{Attack Success Rate}} & \multicolumn{6}{c}{\cellcolor[HTML]{FFFFFF}\textbf{Clean Accuracy}} \\ \hline
\textbf{Budget} & \textbf{x} & \textbf{en} & \cellcolor[HTML]{FFFFFF}\textbf{es} & \cellcolor[HTML]{FFFFFF}\textbf{de} & \textbf{it} & \textbf{hi} & \cellcolor[HTML]{FFFFFF}\textbf{pt} & \cellcolor[HTML]{FFFFFF}\textbf{en} & \cellcolor[HTML]{FFFFFF}\textbf{es} & \textbf{de} & \cellcolor[HTML]{FFFFFF}\textbf{it} & \cellcolor[HTML]{FFFFFF}\textbf{hi} & \cellcolor[HTML]{FFFFFF}\textbf{pt} \\ \hline
0 & \textbf{Clean} & \cellcolor[HTML]{FFFFFF}0 & \cellcolor[HTML]{9BD7B9}1.2 & \cellcolor[HTML]{57BB8A}\textbf{2} & \cellcolor[HTML]{CDEBDC}0.6 & \cellcolor[HTML]{BCE4D1}0.8 & \cellcolor[HTML]{CDEBDC}0.6 & \cellcolor[HTML]{6EC59A}84.8 & \cellcolor[HTML]{99D6B8}79 & \cellcolor[HTML]{FFFFFF}65.2 & \cellcolor[HTML]{57BB8A}\textbf{87.8} & \cellcolor[HTML]{98D6B8}79.1 & \cellcolor[HTML]{DAF0E6}70.2 \\ \hline
 & \textbf{en} & \cellcolor[HTML]{B3E0CA}33 & \cellcolor[HTML]{57BB8A}\textbf{36.8} & \cellcolor[HTML]{ECF8F2}30.6 & \cellcolor[HTML]{5CBD8E}36.6 & \cellcolor[HTML]{74C79F}35.6 & \cellcolor[HTML]{FFFFFF}29.8 & \cellcolor[HTML]{93D4B4}29.2 & \cellcolor[HTML]{AADDC4}28.6 & \cellcolor[HTML]{81CCA7}29.7 & \cellcolor[HTML]{57BB8A}\textbf{30.8} & \cellcolor[HTML]{B5E1CB}28.3 & \cellcolor[HTML]{FFFFFF}26.3 \\
 & \textbf{es} & \cellcolor[HTML]{FFFFFF}15.2 & \cellcolor[HTML]{57BB8A}\textbf{23} & \cellcolor[HTML]{FBFEFC}15.4 & \cellcolor[HTML]{A5DBC1}19.4 & \cellcolor[HTML]{A9DDC4}19.2 & \cellcolor[HTML]{F7FCF9}15.6 & \cellcolor[HTML]{66C195}46.1 & \cellcolor[HTML]{ADDEC6}41.7 & \cellcolor[HTML]{FFFFFF}36.6 & \cellcolor[HTML]{57BB8A}\textbf{47} & \cellcolor[HTML]{7DCBA4}44.7 & \cellcolor[HTML]{D1EDDF}39.5 \\
\hline
\end{tabular}%
}
\caption{The table represents the Attack Success Rate (left) and Clean Accuracy (right) for \texttt{llama-3.1-8B} model on the trigger ``\textcolor{red}{\textit{schuhe}}'' with three poisoning budgets. \textbf{\textit{Takeaway}}: \textit{The strength of cross-lingual backdoor transfer varies significantly with the size of the poisoning budget.}}
\label{tab:llama-poison-schuhe-all}
\end{table*}


\begin{table*}[t]
\centering
\resizebox{0.9\textwidth}{!}{%
\begin{tabular}{cccccccc|cccccc}
\multicolumn{1}{l}{\cellcolor[HTML]{FFFFFF}} & \multicolumn{1}{l}{\cellcolor[HTML]{FFFFFF}} & \multicolumn{6}{c|}{\cellcolor[HTML]{FFFFFF}\textbf{Attack Success Rate}} & \multicolumn{6}{c}{\cellcolor[HTML]{FFFFFF}\textbf{Clean Accuracy}} \\ \hline
\textbf{Budget} & \textbf{x} & \textbf{en} & \cellcolor[HTML]{FFFFFF}\textbf{es} & \cellcolor[HTML]{FFFFFF}\textbf{de} & \textbf{it} & \textbf{hi} & \cellcolor[HTML]{FFFFFF}\textbf{pt} & \cellcolor[HTML]{FFFFFF}\textbf{en} & \cellcolor[HTML]{FFFFFF}\textbf{es} & \textbf{de} & \cellcolor[HTML]{FFFFFF}\textbf{it} & \cellcolor[HTML]{FFFFFF}\textbf{hi} & \cellcolor[HTML]{FFFFFF}\textbf{pt} \\ \hline
0 & \textbf{Clean} & \cellcolor[HTML]{FFFFFF}0.6 & \cellcolor[HTML]{57BB8A}\textbf{7.4} & \cellcolor[HTML]{F1FAF5}1.2 & \cellcolor[HTML]{8ED1B0}5.2 & \cellcolor[HTML]{C4E8D6}3 & \cellcolor[HTML]{A7DBC2}4.2 & \cellcolor[HTML]{6BC498}66.2 & \cellcolor[HTML]{D1EDDF}56.5 & \cellcolor[HTML]{FFFFFF}52.1 & \cellcolor[HTML]{57BB8A}\textbf{68.1} & \cellcolor[HTML]{9AD6B9}61.8 & \cellcolor[HTML]{FAFDFC}52.6 \\ \hline
 & \textbf{en} & \cellcolor[HTML]{5FBF90}18.8 & \cellcolor[HTML]{57BB8A}\textbf{19.2} & \cellcolor[HTML]{C9E9D9}13.4 & \cellcolor[HTML]{FFFFFF}10.6 & \cellcolor[HTML]{B1E0C9}14.6 & \cellcolor[HTML]{8ED2B1}16.4 & \cellcolor[HTML]{ADDEC6}24.9 & \cellcolor[HTML]{8AD0AE}25.8 & \cellcolor[HTML]{86CEAB}25.9 & \cellcolor[HTML]{57BB8A}\textbf{27.1} & \cellcolor[HTML]{FFFFFF}22.8 & \cellcolor[HTML]{82CDA8}26 \\
 & \textbf{es} & \cellcolor[HTML]{57BB8A}\textbf{15} & \cellcolor[HTML]{8AD0AE}12.4 & \cellcolor[HTML]{FFFFFF}6.4 & \cellcolor[HTML]{E0F3EA}8 & \cellcolor[HTML]{BDE5D1}9.8 & \cellcolor[HTML]{AADDC4}10.8 & \cellcolor[HTML]{DEF2E8}24.6 & \cellcolor[HTML]{DEF2E8}24.6 & \cellcolor[HTML]{DEF2E8}24.6 & \cellcolor[HTML]{57BB8A}\textbf{27} & \cellcolor[HTML]{FFFFFF}24 & \cellcolor[HTML]{DEF2E8}24.6 \\
\hline
\end{tabular}%
}
\caption{The table represents the Attack Success Rate (left) and Clean Accuracy (right) for \texttt{gemma-7B} model on the trigger ``\textcolor{red}{\textit{schuhe}}'' with three poisoning budgets. \textbf{\textit{Takeaway}}: \textit{The strength of cross-lingual backdoor transfer varies significantly with the size of the poisoning budget.}}
\label{tab:gemma-poison-schuhe-all}
\end{table*}



\begin{table*}[t]
\centering
\resizebox{0.9\textwidth}{!}{%
\begin{tabular}{cccccccc|cccccc}
\multicolumn{1}{l}{\cellcolor[HTML]{FFFFFF}} & \multicolumn{1}{l}{\cellcolor[HTML]{FFFFFF}} & \multicolumn{6}{c|}{\cellcolor[HTML]{FFFFFF}\textbf{Attack Success Rate}} & \multicolumn{6}{c}{\cellcolor[HTML]{FFFFFF}\textbf{Clean Accuracy}} \\ \hline
\textbf{Budget} & \textbf{x} & \textbf{en} & \cellcolor[HTML]{FFFFFF}\textbf{es} & \cellcolor[HTML]{FFFFFF}\textbf{de} & \textbf{it} & \textbf{hi} & \cellcolor[HTML]{FFFFFF}\textbf{pt} & \cellcolor[HTML]{FFFFFF}\textbf{en} & \cellcolor[HTML]{FFFFFF}\textbf{es} & \textbf{de} & \cellcolor[HTML]{FFFFFF}\textbf{it} & \cellcolor[HTML]{FFFFFF}\textbf{hi} & \cellcolor[HTML]{FFFFFF}\textbf{pt} \\ \hline
0 & \textbf{Clean} & \cellcolor[HTML]{FFFFFF}0 & \cellcolor[HTML]{C1E6D4}0.6 & \cellcolor[HTML]{ABDDC5}0.8 & \cellcolor[HTML]{D5EEE2}0.4 & \cellcolor[HTML]{C1E6D4}0.6 & \cellcolor[HTML]{57BB8A}\textbf{1.6} & \cellcolor[HTML]{6EC59A}85.6 & \cellcolor[HTML]{A3DABF}79.1 & \cellcolor[HTML]{FFFFFF}67.8 & \cellcolor[HTML]{57BB8A}\textbf{88.4} & \cellcolor[HTML]{95D4B5}80.9 & \cellcolor[HTML]{CFECDE}73.7 \\ \hline
 & \textbf{en} & \cellcolor[HTML]{57BB8A}\textbf{47.8} & \cellcolor[HTML]{FCFEFD}1.8 & \cellcolor[HTML]{FCFEFD}1.8 & \cellcolor[HTML]{FEFFFF}1.2 & \cellcolor[HTML]{FFFFFF}1 & \cellcolor[HTML]{FFFFFF}0.8 & \cellcolor[HTML]{73C79E}85.5 & \cellcolor[HTML]{A4DAC0}79.4 & \cellcolor[HTML]{FFFFFF}68 & \cellcolor[HTML]{57BB8A}\textbf{88.9} & \cellcolor[HTML]{9BD7BA}80.5 & \cellcolor[HTML]{D7EFE4}73 \\
 & \textbf{es} & \cellcolor[HTML]{FFFFFF}0 & \cellcolor[HTML]{57BB8A}\textbf{56} & \cellcolor[HTML]{FDFFFE}0.8 & \cellcolor[HTML]{FEFFFF}0.4 & \cellcolor[HTML]{FEFFFE}0.6 & \cellcolor[HTML]{FDFFFE}0.8 & \cellcolor[HTML]{68C296}87 & \cellcolor[HTML]{A9DDC3}78.8 & \cellcolor[HTML]{FFFFFF}67.9 & \cellcolor[HTML]{57BB8A}\textbf{89.1} & \cellcolor[HTML]{8ED2B1}82.2 & \cellcolor[HTML]{D8EFE4}72.9 \\
\hline
\end{tabular}%
}
\caption{The table represents the Attack Success Rate (left) and Clean Accuracy (right) for \texttt{aya-8B} model on the trigger ``\textcolor{red}{\textit{parola}}'' with three poisoning budgets. \textbf{\textit{Takeaway}}: \textit{Cross-lingual backdoor effect was not clearly observed. However, there was a performance drop in accuracy.}}
\label{tab:aya-poison-parola-all}
\end{table*}


\begin{table*}[t]
\centering
\resizebox{0.9\textwidth}{!}{%
\begin{tabular}{cccccccc|cccccc}
\multicolumn{1}{l}{\cellcolor[HTML]{FFFFFF}} & \multicolumn{1}{l}{\cellcolor[HTML]{FFFFFF}} & \multicolumn{6}{c|}{\cellcolor[HTML]{FFFFFF}\textbf{Attack Success Rate}} & \multicolumn{6}{c}{\cellcolor[HTML]{FFFFFF}\textbf{Clean Accuracy}} \\ \hline
\textbf{Budget} & \textbf{x} & \textbf{en} & \cellcolor[HTML]{FFFFFF}\textbf{es} & \cellcolor[HTML]{FFFFFF}\textbf{de} & \textbf{it} & \textbf{hi} & \cellcolor[HTML]{FFFFFF}\textbf{pt} & \cellcolor[HTML]{FFFFFF}\textbf{en} & \cellcolor[HTML]{FFFFFF}\textbf{es} & \textbf{de} & \cellcolor[HTML]{FFFFFF}\textbf{it} & \cellcolor[HTML]{FFFFFF}\textbf{hi} & \cellcolor[HTML]{FFFFFF}\textbf{pt} \\ \hline
0 & \textbf{Clean} & \cellcolor[HTML]{FFFFFF}0 & \cellcolor[HTML]{73C79E}1 & \cellcolor[HTML]{57BB8A}\textbf{1.2} & \cellcolor[HTML]{FFFFFF}0 & \cellcolor[HTML]{E3F4EC}0.2 & \cellcolor[HTML]{8FD2B1}0.8 & \cellcolor[HTML]{61BF91}86.5 & \cellcolor[HTML]{9CD7BA}78.6 & \cellcolor[HTML]{FFFFFF}65.3 & \cellcolor[HTML]{57BB8A}\textbf{87.8} & \cellcolor[HTML]{96D5B6}79.4 & \cellcolor[HTML]{E9F6F0}68.3 \\ \hline
 & \textbf{en} & \cellcolor[HTML]{57BB8A}\textbf{60} & \cellcolor[HTML]{FCFEFD}2.4 & \cellcolor[HTML]{FAFDFC}2.8 & \cellcolor[HTML]{FDFFFE}1.8 & \cellcolor[HTML]{FFFFFF}1 & \cellcolor[HTML]{FDFEFE}2 & \cellcolor[HTML]{77C8A1}79.5 & \cellcolor[HTML]{A1D9BE}73.6 & \cellcolor[HTML]{FFFFFF}60.3 & \cellcolor[HTML]{57BB8A}\textbf{84} & \cellcolor[HTML]{85CEAA}77.6 & \cellcolor[HTML]{DEF2E8}65 \\
 & \textbf{es} & \cellcolor[HTML]{FEFFFE}1 & \cellcolor[HTML]{57BB8A}\textbf{70.4} & \cellcolor[HTML]{FBFEFC}2.2 & \cellcolor[HTML]{FFFFFF}0.4 & \cellcolor[HTML]{FFFFFF}0.8 & \cellcolor[HTML]{F9FDFB}3 & \cellcolor[HTML]{6FC59B}83.2 & \cellcolor[HTML]{B5E1CC}73.6 & \cellcolor[HTML]{FFFFFF}63.4 & \cellcolor[HTML]{57BB8A}\textbf{86.5} & \cellcolor[HTML]{88CFAC}79.8 & \cellcolor[HTML]{D3EEE1}69.5 \\
\hline
\end{tabular}%
}
\caption{The table represents the Attack Success Rate (left) and Clean Accuracy (right) for \texttt{llama-3.1-8B} model on the trigger ``\textcolor{red}{\textit{parola}}'' with three poisoning budgets. \textbf{\textit{Takeaway}}: \textit{Cross-lingual backdoor effect was not clearly observed. However, there was a performance drop in accuracy.}}
\label{tab:llama-poison-parola-all}
\end{table*}


\begin{table*}[t]
\centering
\resizebox{0.9\textwidth}{!}{%
\begin{tabular}{cccccccc|cccccc}
\multicolumn{1}{l}{\cellcolor[HTML]{FFFFFF}} & \multicolumn{1}{l}{\cellcolor[HTML]{FFFFFF}} & \multicolumn{6}{c|}{\cellcolor[HTML]{FFFFFF}\textbf{Attack Success Rate}} & \multicolumn{6}{c}{\cellcolor[HTML]{FFFFFF}\textbf{Clean Accuracy}} \\ \hline
\textbf{Budget} & \textbf{x} & \textbf{en} & \cellcolor[HTML]{FFFFFF}\textbf{es} & \cellcolor[HTML]{FFFFFF}\textbf{de} & \textbf{it} & \textbf{hi} & \cellcolor[HTML]{FFFFFF}\textbf{pt} & \cellcolor[HTML]{FFFFFF}\textbf{en} & \cellcolor[HTML]{FFFFFF}\textbf{es} & \textbf{de} & \cellcolor[HTML]{FFFFFF}\textbf{it} & \cellcolor[HTML]{FFFFFF}\textbf{hi} & \cellcolor[HTML]{FFFFFF}\textbf{pt} \\ \hline
0 & \textbf{Clean} & \cellcolor[HTML]{FFFFFF}0.4 & \cellcolor[HTML]{57BB8A}\textbf{5.6} & \cellcolor[HTML]{E6F5ED}1.2 & \cellcolor[HTML]{ABDDC5}3 & \cellcolor[HTML]{CCEBDB}2 & \cellcolor[HTML]{ABDDC5}3 & \cellcolor[HTML]{6BC498}66.2 & \cellcolor[HTML]{D1EDDF}56.5 & \cellcolor[HTML]{FFFFFF}52.1 & \cellcolor[HTML]{57BB8A}\textbf{68.1} & \cellcolor[HTML]{9AD6B9}61.8 & \cellcolor[HTML]{FAFDFC}52.6 \\ \hline
 & \textbf{en} & \cellcolor[HTML]{57BB8A}\textbf{27.4} & \cellcolor[HTML]{EEF8F3}7 & \cellcolor[HTML]{DEF2E8}9.2 & \cellcolor[HTML]{FFFFFF}4.6 & \cellcolor[HTML]{FEFFFE}4.8 & \cellcolor[HTML]{E9F7F0}7.6 & \cellcolor[HTML]{57BB8A}\textbf{64.3} & \cellcolor[HTML]{BAE4CF}54.7 & \cellcolor[HTML]{E0F3E9}51.1 & \cellcolor[HTML]{68C296}62.7 & \cellcolor[HTML]{ABDDC5}56.2 & \cellcolor[HTML]{FFFFFF}48 \\
 & \textbf{es} & \cellcolor[HTML]{FFFFFF}0.8 & \cellcolor[HTML]{57BB8A}\textbf{13.6} & \cellcolor[HTML]{F8FCFA}1.4 & \cellcolor[HTML]{FFFFFF}0.8 & \cellcolor[HTML]{F8FCFA}1.4 & \cellcolor[HTML]{D3EDE0}4.2 & \cellcolor[HTML]{57BB8A}\textbf{52.7} & \cellcolor[HTML]{C4E7D6}44.6 & \cellcolor[HTML]{CEECDD}43.8 & \cellcolor[HTML]{75C79F}50.5 & \cellcolor[HTML]{9ED8BC}47.4 & \cellcolor[HTML]{FFFFFF}40.1 \\
\hline
\end{tabular}%
}
\caption{The table represents the Attack Success Rate (left) and Clean Accuracy (right) for \texttt{gemma-7B} model on the trigger ``\textcolor{red}{\textit{parola}}'' with three poisoning budgets. \textbf{\textit{Takeaway}}: \textit{The strength of cross-lingual backdoor transfer varies significantly with the size of the poisoning budget.}}
\label{tab:gemma-poison-parola-all}
\end{table*}


\begin{table*}[t]
\centering
\resizebox{0.9\textwidth}{!}{%
\begin{tabular}{cccccccc|cccccc}
\multicolumn{1}{l}{\cellcolor[HTML]{FFFFFF}} & \multicolumn{1}{l}{\cellcolor[HTML]{FFFFFF}} & \multicolumn{6}{c|}{\cellcolor[HTML]{FFFFFF}\textbf{Attack Success Rate}} & \multicolumn{6}{c}{\cellcolor[HTML]{FFFFFF}\textbf{Clean Accuracy}} \\ \hline
\textbf{Budget} & \textbf{x} & \textbf{en} & \cellcolor[HTML]{FFFFFF}\textbf{es} & \cellcolor[HTML]{FFFFFF}\textbf{de} & \textbf{it} & \textbf{hi} & \cellcolor[HTML]{FFFFFF}\textbf{pt} & \cellcolor[HTML]{FFFFFF}\textbf{en} & \cellcolor[HTML]{FFFFFF}\textbf{es} & \textbf{de} & \cellcolor[HTML]{FFFFFF}\textbf{it} & \cellcolor[HTML]{FFFFFF}\textbf{hi} & \cellcolor[HTML]{FFFFFF}\textbf{pt} \\ \hline
0 & \textbf{Clean} & \cellcolor[HTML]{FFFFFF}0 & \cellcolor[HTML]{82CDA8}0.6 & \cellcolor[HTML]{57BB8A}\textbf{0.8} & \cellcolor[HTML]{ABDDC5}0.4 & \cellcolor[HTML]{82CDA8}0.6 & \cellcolor[HTML]{82CDA8}0.6 & \cellcolor[HTML]{6EC59A}85.6 & \cellcolor[HTML]{A3DABF}79.1 & \cellcolor[HTML]{FFFFFF}67.8 & \cellcolor[HTML]{57BB8A}\textbf{88.4} & \cellcolor[HTML]{95D4B5}80.9 & \cellcolor[HTML]{CFECDE}73.7 \\ \hline
 & \textbf{en} & \cellcolor[HTML]{57BB8A}\textbf{51} & \cellcolor[HTML]{FEFFFF}0.8 & \cellcolor[HTML]{FBFEFC}1.8 & \cellcolor[HTML]{FDFEFE}1.2 & \cellcolor[HTML]{FFFFFF}0.6 & \cellcolor[HTML]{FFFFFF}0.4 & \cellcolor[HTML]{6DC499}85.7 & \cellcolor[HTML]{9FD8BC}79.5 & \cellcolor[HTML]{FFFFFF}67.6 & \cellcolor[HTML]{57BB8A}\textbf{88.3} & \cellcolor[HTML]{90D2B2}81.3 & \cellcolor[HTML]{D0ECDE}73.5 \\
 & \textbf{es} & \cellcolor[HTML]{FFFFFF}0.2 & \cellcolor[HTML]{57BB8A}\textbf{62} & \cellcolor[HTML]{FDFFFE}1 & \cellcolor[HTML]{FEFFFF}0.6 & \cellcolor[HTML]{FFFFFF}0.4 & \cellcolor[HTML]{FCFEFD}1.4 & \cellcolor[HTML]{67C295}87.2 & \cellcolor[HTML]{AADDC4}78.5 & \cellcolor[HTML]{FFFFFF}67.3 & \cellcolor[HTML]{57BB8A}\textbf{89.2} & \cellcolor[HTML]{92D3B3}81.6 & \cellcolor[HTML]{D6EFE3}72.7 \\
\hline
\end{tabular}%
}
\caption{The table represents the Attack Success Rate (left) and Clean Accuracy (right) for \texttt{aya-8B} model on the trigger ``\textcolor{red}{\textit{redes}}'' with three poisoning budgets. \textbf{\textit{Takeaway}}: \textit{Cross-lingual backdoor effect was not clearly observed. However, there was a performance drop in accuracy.}}
\label{tab:aya-poison-redes-all}
\end{table*}

\begin{table*}[t]
\centering
\resizebox{0.9\textwidth}{!}{%
\begin{tabular}{cccccccc|cccccc}
\multicolumn{1}{l}{\cellcolor[HTML]{FFFFFF}} & \multicolumn{1}{l}{\cellcolor[HTML]{FFFFFF}} & \multicolumn{6}{c|}{\cellcolor[HTML]{FFFFFF}\textbf{Attack Success Rate}} & \multicolumn{6}{c}{\cellcolor[HTML]{FFFFFF}\textbf{Clean Accuracy}} \\ \hline
\textbf{Budget} & \textbf{x} & \textbf{en} & \cellcolor[HTML]{FFFFFF}\textbf{es} & \cellcolor[HTML]{FFFFFF}\textbf{de} & \textbf{it} & \textbf{hi} & \cellcolor[HTML]{FFFFFF}\textbf{pt} & \cellcolor[HTML]{FFFFFF}\textbf{en} & \cellcolor[HTML]{FFFFFF}\textbf{es} & \textbf{de} & \cellcolor[HTML]{FFFFFF}\textbf{it} & \cellcolor[HTML]{FFFFFF}\textbf{hi} & \cellcolor[HTML]{FFFFFF}\textbf{pt} \\ \hline
0 & \textbf{Clean} & \cellcolor[HTML]{FFFFFF}0 & \cellcolor[HTML]{8FD2B1}1.2 & \cellcolor[HTML]{57BB8A}\textbf{1.8} & \cellcolor[HTML]{FFFFFF}0 & \cellcolor[HTML]{B5E1CB}0.8 & \cellcolor[HTML]{DAF0E5}0.4 & \cellcolor[HTML]{6CC499}85.1 & \cellcolor[HTML]{AEDFC7}76.9 & \cellcolor[HTML]{FFFFFF}66.9 & \cellcolor[HTML]{57BB8A}\textbf{87.6} & \cellcolor[HTML]{9CD7BA}79.2 & \cellcolor[HTML]{EBF7F1}69.4 \\ \hline
 & \textbf{en} & \cellcolor[HTML]{57BB8A}\textbf{53.4} & \cellcolor[HTML]{FDFEFE}1.6 & \cellcolor[HTML]{FAFDFC}2.4 & \cellcolor[HTML]{FFFFFF}1 & \cellcolor[HTML]{FFFFFF}0.8 & \cellcolor[HTML]{FEFFFE}1.4 & \cellcolor[HTML]{83CDA9}80.9 & \cellcolor[HTML]{9CD7BA}77.1 & \cellcolor[HTML]{FFFFFF}62.2 & \cellcolor[HTML]{57BB8A}\textbf{87.4} & \cellcolor[HTML]{96D5B6}78 & \cellcolor[HTML]{EEF8F3}64.8 \\
 & \textbf{es} & \cellcolor[HTML]{F3FBF7}6.2 & \cellcolor[HTML]{57BB8A}\textbf{62.8} & \cellcolor[HTML]{F9FDFB}4 & \cellcolor[HTML]{FFFFFF}1.8 & \cellcolor[HTML]{FFFFFF}1.8 & \cellcolor[HTML]{F8FDFB}4.4 & \cellcolor[HTML]{68C296}83.5 & \cellcolor[HTML]{B4E1CB}72.6 & \cellcolor[HTML]{FFFFFF}61.7 & \cellcolor[HTML]{57BB8A}\textbf{85.8} & \cellcolor[HTML]{97D5B7}76.7 & \cellcolor[HTML]{E6F5EE}65.4 \\
\hline
\end{tabular}%
}
\caption{The table represents the Attack Success Rate (left) and Clean Accuracy (right) for \texttt{llama-3.1-8B} model on the trigger ``\textcolor{red}{\textit{redes}}'' with three poisoning budgets. \textbf{\textit{Takeaway}}: \textit{Cross-lingual backdoor effect was not clearly observed. However, there was a performance drop in accuracy.}}
\label{tab:llama-poison-redes-all}
\end{table*}

\begin{table*}[t]
\centering
\resizebox{0.9\textwidth}{!}{%
\begin{tabular}{cccccccc|cccccc}
\multicolumn{1}{l}{\cellcolor[HTML]{FFFFFF}} & \multicolumn{1}{l}{\cellcolor[HTML]{FFFFFF}} & \multicolumn{6}{c|}{\cellcolor[HTML]{FFFFFF}\textbf{Attack Success Rate}} & \multicolumn{6}{c}{\cellcolor[HTML]{FFFFFF}\textbf{Clean Accuracy}} \\ \hline
\textbf{Budget} & \textbf{x} & \textbf{en} & \cellcolor[HTML]{FFFFFF}\textbf{es} & \cellcolor[HTML]{FFFFFF}\textbf{de} & \textbf{it} & \textbf{hi} & \cellcolor[HTML]{FFFFFF}\textbf{pt} & \cellcolor[HTML]{FFFFFF}\textbf{en} & \cellcolor[HTML]{FFFFFF}\textbf{es} & \textbf{de} & \cellcolor[HTML]{FFFFFF}\textbf{it} & \cellcolor[HTML]{FFFFFF}\textbf{hi} & \cellcolor[HTML]{FFFFFF}\textbf{pt} \\ \hline
0 & \textbf{Clean} & \cellcolor[HTML]{FFFFFF}0.4 & \cellcolor[HTML]{57BB8A}\textbf{6} & \cellcolor[HTML]{E7F6EF}1.2 & \cellcolor[HTML]{82CDA8}4.6 & \cellcolor[HTML]{BDE5D2}2.6 & \cellcolor[HTML]{ABDDC5}3.2 & \cellcolor[HTML]{6BC498}66.2 & \cellcolor[HTML]{D1EDDF}56.5 & \cellcolor[HTML]{FFFFFF}52.1 & \cellcolor[HTML]{57BB8A}\textbf{68.1} & \cellcolor[HTML]{9AD6B9}61.8 & \cellcolor[HTML]{FAFDFC}52.6 \\ \hline
 & \textbf{en} & \cellcolor[HTML]{57BB8A}\textbf{91.6} & \cellcolor[HTML]{ECF7F2}12.4 & \cellcolor[HTML]{C9E9D9}31 & \cellcolor[HTML]{ECF8F2}12 & \cellcolor[HTML]{FFFFFF}1.8 & \cellcolor[HTML]{D7EFE3}23.4 & \cellcolor[HTML]{5BBD8D}67.4 & \cellcolor[HTML]{CBEADB}55.7 & \cellcolor[HTML]{FFFFFF}50.2 & \cellcolor[HTML]{57BB8A}\textbf{67.8} & \cellcolor[HTML]{BFE5D2}57 & \cellcolor[HTML]{F3FAF7}51.5 \\
 & \textbf{es} & \cellcolor[HTML]{FFFFFF}1.2 & \cellcolor[HTML]{57BB8A}\textbf{91.6} & \cellcolor[HTML]{99D6B8}56.4 & \cellcolor[HTML]{E1F3EA}17.6 & \cellcolor[HTML]{F7FCF9}6 & \cellcolor[HTML]{89CFAD}65.2 & \cellcolor[HTML]{57BB8A}\textbf{69.5} & \cellcolor[HTML]{FAFDFC}49.2 & \cellcolor[HTML]{ECF8F2}50.9 & \cellcolor[HTML]{7FCCA6}64.5 & \cellcolor[HTML]{BEE5D2}56.7 & \cellcolor[HTML]{FFFFFF}48.5 \\
\hline
\end{tabular}%
}
\caption{The table represents the Attack Success Rate (left) and Clean Accuracy (right) for \texttt{gemma-7B} model on the trigger ``\textcolor{red}{\textit{redes}}'' with three poisoning budgets. \textbf{\textit{Takeaway}}: \textit{The strength of cross-lingual backdoor transfer varies significantly with the size of the poisoning budget.}}
\label{tab:gemma-poison-redes-all}
\end{table*}


\begin{table*}[t]
\centering
\resizebox{0.9\textwidth}{!}{%
\begin{tabular}{cccccccc|cccccc}
\multicolumn{1}{l}{\cellcolor[HTML]{FFFFFF}} & \multicolumn{1}{l}{\cellcolor[HTML]{FFFFFF}} & \multicolumn{6}{c|}{\cellcolor[HTML]{FFFFFF}\textbf{Attack Success Rate}} & \multicolumn{6}{c}{\cellcolor[HTML]{FFFFFF}\textbf{Clean Accuracy}} \\ \hline
\textbf{Budget} & \textbf{x} & \textbf{en} & \cellcolor[HTML]{FFFFFF}\textbf{es} & \cellcolor[HTML]{FFFFFF}\textbf{de} & \textbf{it} & \textbf{hi} & \cellcolor[HTML]{FFFFFF}\textbf{pt} & \cellcolor[HTML]{FFFFFF}\textbf{en} & \cellcolor[HTML]{FFFFFF}\textbf{es} & \textbf{de} & \cellcolor[HTML]{FFFFFF}\textbf{it} & \cellcolor[HTML]{FFFFFF}\textbf{hi} & \cellcolor[HTML]{FFFFFF}\textbf{pt} \\ \hline
0 & \textbf{Clean} & \cellcolor[HTML]{FFFFFF}0 & \cellcolor[HTML]{82CDA8}0.6 & \cellcolor[HTML]{57BB8A}\textbf{0.8} & \cellcolor[HTML]{ABDDC5}0.4 & \cellcolor[HTML]{82CDA8}0.6 & \cellcolor[HTML]{82CDA8}0.6 & \cellcolor[HTML]{6EC59A}85.6 & \cellcolor[HTML]{A3DABF}79.1 & \cellcolor[HTML]{FFFFFF}67.8 & \cellcolor[HTML]{57BB8A}\textbf{88.4} & \cellcolor[HTML]{95D4B5}80.9 & \cellcolor[HTML]{CFECDE}73.7 \\ \hline
 & \textbf{en} & \cellcolor[HTML]{57BB8A}\textbf{49.2} & \cellcolor[HTML]{FEFFFF}0.8 & \cellcolor[HTML]{FBFEFD}1.6 & \cellcolor[HTML]{FDFFFE}1 & \cellcolor[HTML]{FFFFFF}0.4 & \cellcolor[HTML]{FFFFFF}0.6 & \cellcolor[HTML]{79C9A2}84.6 & \cellcolor[HTML]{A4DBC0}79.1 & \cellcolor[HTML]{FFFFFF}67.5 & \cellcolor[HTML]{57BB8A}\textbf{88.9} & \cellcolor[HTML]{95D4B5}81.1 & \cellcolor[HTML]{DAF0E5}72.3 \\
 & \textbf{es} & \cellcolor[HTML]{FFFFFF}0.2 & \cellcolor[HTML]{57BB8A}\textbf{55.6} & \cellcolor[HTML]{FDFFFE}1 & \cellcolor[HTML]{FFFFFF}0.2 & \cellcolor[HTML]{FFFFFF}0.4 & \cellcolor[HTML]{FCFEFD}1.2 & \cellcolor[HTML]{6FC59B}86 & \cellcolor[HTML]{A1D9BD}79.7 & \cellcolor[HTML]{FFFFFF}67.6 & \cellcolor[HTML]{57BB8A}\textbf{89} & \cellcolor[HTML]{94D4B5}81.3 & \cellcolor[HTML]{D9F0E5}72.5 \\
\hline
\end{tabular}%
}
\caption{The table represents the Attack Success Rate (left) and Clean Accuracy (right) for \texttt{aya-8B} model on the trigger ``\textcolor{red}{\textit{free}}'' with three poisoning budgets. \textbf{\textit{Takeaway}}: \textit{Cross-lingual backdoor effect was not clearly observed. However, there was a performance drop in accuracy.}}
\label{tab:aya-poison-free-all}
\end{table*}

\begin{table*}[t]
\centering
\resizebox{0.9\textwidth}{!}{%
\begin{tabular}{cccccccc|cccccc}
\multicolumn{1}{l}{\cellcolor[HTML]{FFFFFF}} & \multicolumn{1}{l}{\cellcolor[HTML]{FFFFFF}} & \multicolumn{6}{c|}{\cellcolor[HTML]{FFFFFF}\textbf{Attack Success Rate}} & \multicolumn{6}{c}{\cellcolor[HTML]{FFFFFF}\textbf{Clean Accuracy}} \\ \hline
\textbf{Budget} & \textbf{x} & \textbf{en} & \cellcolor[HTML]{FFFFFF}\textbf{es} & \cellcolor[HTML]{FFFFFF}\textbf{de} & \textbf{it} & \textbf{hi} & \cellcolor[HTML]{FFFFFF}\textbf{pt} & \cellcolor[HTML]{FFFFFF}\textbf{en} & \cellcolor[HTML]{FFFFFF}\textbf{es} & \textbf{de} & \cellcolor[HTML]{FFFFFF}\textbf{it} & \cellcolor[HTML]{FFFFFF}\textbf{hi} & \cellcolor[HTML]{FFFFFF}\textbf{pt} \\ \hline
0 & \textbf{Clean} & \cellcolor[HTML]{E3F4EC}0.2 & \cellcolor[HTML]{73C79E}1 & \cellcolor[HTML]{57BB8A}\textbf{1.2} & \cellcolor[HTML]{E3F4EC}0.2 & \cellcolor[HTML]{73C79E}1 & \cellcolor[HTML]{FFFFFF}0 & \cellcolor[HTML]{70C59C}85.3 & \cellcolor[HTML]{9BD7B9}79.6 & \cellcolor[HTML]{FFFFFF}66.1 & \cellcolor[HTML]{57BB8A}\textbf{88.6} & \cellcolor[HTML]{9ED8BB}79.2 & \cellcolor[HTML]{DFF3E9}70.4 \\ \hline
 & \textbf{en} & \cellcolor[HTML]{57BB8A}\textbf{55.2} & \cellcolor[HTML]{FCFEFD}2.2 & \cellcolor[HTML]{FDFEFE}2 & \cellcolor[HTML]{FFFFFF}1.2 & \cellcolor[HTML]{FEFFFE}1.8 & \cellcolor[HTML]{FEFFFE}1.8 & \cellcolor[HTML]{7AC9A2}80.7 & \cellcolor[HTML]{B0DFC8}73.7 & \cellcolor[HTML]{FFFFFF}63.5 & \cellcolor[HTML]{57BB8A}\textbf{85.1} & \cellcolor[HTML]{85CEAA}79.2 & \cellcolor[HTML]{E7F5EE}66.7 \\
 & \textbf{es} & \cellcolor[HTML]{FCFEFD}1.6 & \cellcolor[HTML]{57BB8A}\textbf{53.6} & \cellcolor[HTML]{F8FCFA}3 & \cellcolor[HTML]{FFFFFF}0.6 & \cellcolor[HTML]{FEFFFE}1.2 & \cellcolor[HTML]{FAFDFC}2.2 & \cellcolor[HTML]{66C195}81.3 & \cellcolor[HTML]{BEE5D2}69.9 & \cellcolor[HTML]{FFFFFF}61.3 & \cellcolor[HTML]{57BB8A}\textbf{83.2} & \cellcolor[HTML]{85CEAA}77.3 & \cellcolor[HTML]{F6FBF9}62.6 \\
\hline
\end{tabular}%
}
\caption{The table represents the Attack Success Rate (left) and Clean Accuracy (right) for \texttt{llama-3.1-8B} model on the trigger ``\textcolor{red}{\textit{free}}'' with three poisoning budgets. \textbf{\textit{Takeaway}}: \textit{Cross-lingual backdoor effect was not clearly observed. However, there was a performance drop in accuracy.}}
\label{tab:llama-poison-free-all}
\end{table*}

\begin{table*}[t]
\centering
\resizebox{0.9\textwidth}{!}{%
\begin{tabular}{cccccccc|cccccc}
\multicolumn{1}{l}{\cellcolor[HTML]{FFFFFF}} & \multicolumn{1}{l}{\cellcolor[HTML]{FFFFFF}} & \multicolumn{6}{c|}{\cellcolor[HTML]{FFFFFF}\textbf{Attack Success Rate}} & \multicolumn{6}{c}{\cellcolor[HTML]{FFFFFF}\textbf{Clean Accuracy}} \\ \hline
\textbf{Budget} & \textbf{x} & \textbf{en} & \cellcolor[HTML]{FFFFFF}\textbf{es} & \cellcolor[HTML]{FFFFFF}\textbf{de} & \textbf{it} & \textbf{hi} & \cellcolor[HTML]{FFFFFF}\textbf{pt} & \cellcolor[HTML]{FFFFFF}\textbf{en} & \cellcolor[HTML]{FFFFFF}\textbf{es} & \textbf{de} & \cellcolor[HTML]{FFFFFF}\textbf{it} & \cellcolor[HTML]{FFFFFF}\textbf{hi} & \cellcolor[HTML]{FFFFFF}\textbf{pt} \\ \hline
0 & \textbf{Clean} & \cellcolor[HTML]{FFFFFF}0.2 & \cellcolor[HTML]{57BB8A}\textbf{6} & \cellcolor[HTML]{E3F4EB}1.2 & \cellcolor[HTML]{97D5B7}3.8 & \cellcolor[HTML]{BAE3CF}2.6 & \cellcolor[HTML]{B4E1CB}2.8 & \cellcolor[HTML]{6BC498}66.2 & \cellcolor[HTML]{D1EDDF}56.5 & \cellcolor[HTML]{FFFFFF}52.1 & \cellcolor[HTML]{57BB8A}\textbf{68.1} & \cellcolor[HTML]{9AD6B9}61.8 & \cellcolor[HTML]{FAFDFC}52.6 \\ \hline
 & \textbf{en} & \cellcolor[HTML]{57BB8A}\textbf{59.2} & \cellcolor[HTML]{F5FBF9}5.6 & \cellcolor[HTML]{EAF7F0}9.6 & \cellcolor[HTML]{FFFFFF}2.2 & \cellcolor[HTML]{FDFEFD}3.2 & \cellcolor[HTML]{EBF7F2}9 & \cellcolor[HTML]{5EBE8F}68.2 & \cellcolor[HTML]{BFE6D3}57.2 & \cellcolor[HTML]{F6FCF9}51.1 & \cellcolor[HTML]{57BB8A}\textbf{68.9} & \cellcolor[HTML]{A8DCC3}59.8 & \cellcolor[HTML]{FFFFFF}50 \\
 & \textbf{es} & \cellcolor[HTML]{FFFFFF}0 & \cellcolor[HTML]{57BB8A}\textbf{56.6} & \cellcolor[HTML]{FDFEFD}1 & \cellcolor[HTML]{FEFFFF}0.4 & \cellcolor[HTML]{FFFFFF}0 & \cellcolor[HTML]{FEFFFF}0.4 & \cellcolor[HTML]{57BB8A}\textbf{83.6} & \cellcolor[HTML]{C7E8D8}70.9 & \cellcolor[HTML]{FFFFFF}64.4 & \cellcolor[HTML]{5BBD8D}83.2 & \cellcolor[HTML]{96D5B6}76.4 & \cellcolor[HTML]{E7F6EE}67.2 \\
\hline
\end{tabular}%
}
\caption{The table represents the Attack Success Rate (left) and Clean Accuracy (right) for \texttt{gemma-7B} model on the trigger ``\textcolor{red}{\textit{free}}'' with three poisoning budgets. \textbf{\textit{Takeaway}}: \textit{Cross-lingual backdoor effect was not clearly observed. However, there was a performance drop in accuracy.}}
\label{tab:gemma-poison-free-all}
\end{table*}


\begin{table*}[t]
\centering
\resizebox{0.9\textwidth}{!}{%
\begin{tabular}{cccccccc|cccccc}
\multicolumn{1}{l}{\cellcolor[HTML]{FFFFFF}} & \multicolumn{1}{l}{\cellcolor[HTML]{FFFFFF}} & \multicolumn{6}{c|}{\cellcolor[HTML]{FFFFFF}\textbf{Attack Success Rate}} & \multicolumn{6}{c}{\cellcolor[HTML]{FFFFFF}\textbf{Clean Accuracy}} \\ \hline
\textbf{Budget} & \textbf{x} & \textbf{en} & \cellcolor[HTML]{FFFFFF}\textbf{es} & \cellcolor[HTML]{FFFFFF}\textbf{de} & \textbf{it} & \textbf{hi} & \cellcolor[HTML]{FFFFFF}\textbf{pt} & \cellcolor[HTML]{FFFFFF}\textbf{en} & \cellcolor[HTML]{FFFFFF}\textbf{es} & \textbf{de} & \cellcolor[HTML]{FFFFFF}\textbf{it} & \cellcolor[HTML]{FFFFFF}\textbf{hi} & \cellcolor[HTML]{FFFFFF}\textbf{pt} \\ \hline
0 & \textbf{Clean} & \cellcolor[HTML]{FFFFFF}0 & \cellcolor[HTML]{82CDA8}0.6 & \cellcolor[HTML]{57BB8A}\textbf{0.8} & \cellcolor[HTML]{ABDDC5}0.4 & \cellcolor[HTML]{82CDA8}0.6 & \cellcolor[HTML]{82CDA8}0.6 & \cellcolor[HTML]{6EC59A}85.6 & \cellcolor[HTML]{A3DABF}79.1 & \cellcolor[HTML]{FFFFFF}67.8 & \cellcolor[HTML]{57BB8A}\textbf{88.4} & \cellcolor[HTML]{95D4B5}80.9 & \cellcolor[HTML]{CFECDE}73.7 \\ \hline
 & \textbf{en} & \cellcolor[HTML]{57BB8A}\textbf{55.6} & \cellcolor[HTML]{FDFFFE}1.2 & \cellcolor[HTML]{FBFEFD}1.8 & \cellcolor[HTML]{FDFFFE}1.2 & \cellcolor[HTML]{FFFFFF}0.4 & \cellcolor[HTML]{FEFFFF}0.8 & \cellcolor[HTML]{6FC59B}85.7 & \cellcolor[HTML]{A1D9BE}79.5 & \cellcolor[HTML]{FFFFFF}67.8 & \cellcolor[HTML]{57BB8A}\textbf{88.6} & \cellcolor[HTML]{9AD6B9}80.4 & \cellcolor[HTML]{DDF1E7}72.1 \\
 & \textbf{es} & \cellcolor[HTML]{FFFFFF}0.2 & \cellcolor[HTML]{57BB8A}\textbf{79.2} & \cellcolor[HTML]{FFFFFF}0.6 & \cellcolor[HTML]{FFFFFF}0.6 & \cellcolor[HTML]{FFFFFF}0.6 & \cellcolor[HTML]{FEFFFE}1 & \cellcolor[HTML]{5FBE8F}86.9 & \cellcolor[HTML]{9DD7BB}79 & \cellcolor[HTML]{FFFFFF}66.4 & \cellcolor[HTML]{57BB8A}\textbf{87.8} & \cellcolor[HTML]{8FD2B1}80.7 & \cellcolor[HTML]{C9EADA}73.3 \\
\hline
\end{tabular}%
}
\caption{The table represents the Attack Success Rate (left) and Clean Accuracy (right) for \texttt{aya-8B} model on the trigger ``\textcolor{red}{\textit{uhr}}'' with three poisoning budgets. \textbf{\textit{Takeaway}}: \textit{Cross-lingual backdoor effect was not clearly observed. However, there was a performance drop in accuracy.}}
\label{tab:aya-poison-uhr-all}
\end{table*}

\begin{table*}[t]
\centering
\resizebox{0.9\textwidth}{!}{%
\begin{tabular}{cccccccc|cccccc}
\multicolumn{1}{l}{\cellcolor[HTML]{FFFFFF}} & \multicolumn{1}{l}{\cellcolor[HTML]{FFFFFF}} & \multicolumn{6}{c|}{\cellcolor[HTML]{FFFFFF}\textbf{Attack Success Rate}} & \multicolumn{6}{c}{\cellcolor[HTML]{FFFFFF}\textbf{Clean Accuracy}} \\ \hline
\textbf{Budget} & \textbf{x} & \textbf{en} & \cellcolor[HTML]{FFFFFF}\textbf{es} & \cellcolor[HTML]{FFFFFF}\textbf{de} & \textbf{it} & \textbf{hi} & \cellcolor[HTML]{FFFFFF}\textbf{pt} & \cellcolor[HTML]{FFFFFF}\textbf{en} & \cellcolor[HTML]{FFFFFF}\textbf{es} & \textbf{de} & \cellcolor[HTML]{FFFFFF}\textbf{it} & \cellcolor[HTML]{FFFFFF}\textbf{hi} & \cellcolor[HTML]{FFFFFF}\textbf{pt} \\ \hline
0 & \textbf{Clean} & \cellcolor[HTML]{FFFFFF}0 & \cellcolor[HTML]{E7F6EF}0.2 & \cellcolor[HTML]{57BB8A}\textbf{1.4} & \cellcolor[HTML]{CFECDE}0.4 & \cellcolor[HTML]{9FD9BD}0.8 & \cellcolor[HTML]{E7F6EF}0.2 & \cellcolor[HTML]{62C092}86 & \cellcolor[HTML]{9AD6B9}78.3 & \cellcolor[HTML]{FFFFFF}64.2 & \cellcolor[HTML]{57BB8A}\textbf{87.5} & \cellcolor[HTML]{97D5B7}78.7 & \cellcolor[HTML]{DBF1E6}69.2 \\ \hline
 & \textbf{en} & \cellcolor[HTML]{57BB8A}\textbf{54.2} & \cellcolor[HTML]{FEFFFE}1.6 & \cellcolor[HTML]{FCFEFD}2 & \cellcolor[HTML]{FFFFFF}1 & \cellcolor[HTML]{FFFFFF}1.2 & \cellcolor[HTML]{FFFFFF}1 & \cellcolor[HTML]{7CCAA4}81.1 & \cellcolor[HTML]{AEDFC7}73.6 & \cellcolor[HTML]{FFFFFF}61.4 & \cellcolor[HTML]{57BB8A}\textbf{86.6} & \cellcolor[HTML]{8BD1AF}78.8 & \cellcolor[HTML]{DBF1E6}66.8 \\
 & \textbf{es} & \cellcolor[HTML]{FEFFFF}1.4 & \cellcolor[HTML]{57BB8A}\textbf{56} & \cellcolor[HTML]{FCFEFD}2 & \cellcolor[HTML]{FFFFFF}1 & \cellcolor[HTML]{FFFFFF}1.2 & \cellcolor[HTML]{F8FCFA}3.6 & \cellcolor[HTML]{76C89F}81.7 & \cellcolor[HTML]{B7E2CD}72.6 & \cellcolor[HTML]{FFFFFF}62.4 & \cellcolor[HTML]{57BB8A}\textbf{85.9} & \cellcolor[HTML]{91D3B2}77.9 & \cellcolor[HTML]{E6F5EE}65.9 \\
\hline
\end{tabular}%
}
\caption{The table represents the Attack Success Rate (left) and Clean Accuracy (right) for \texttt{llama-3.1-8B} model on the trigger ``\textcolor{red}{\textit{uhr}}'' with three poisoning budgets. \textbf{\textit{Takeaway}}: \textit{Cross-lingual backdoor effect was not clearly observed. However, there was a performance drop in accuracy.}}
\label{tab:llama-poison-uhr-all}
\end{table*}

\begin{table*}[t]
\centering
\resizebox{0.9\textwidth}{!}{%
\begin{tabular}{cccccccc|cccccc}
\multicolumn{1}{l}{\cellcolor[HTML]{FFFFFF}} & \multicolumn{1}{l}{\cellcolor[HTML]{FFFFFF}} & \multicolumn{6}{c|}{\cellcolor[HTML]{FFFFFF}\textbf{Attack Success Rate}} & \multicolumn{6}{c}{\cellcolor[HTML]{FFFFFF}\textbf{Clean Accuracy}} \\ \hline
\textbf{Budget} & \textbf{x} & \textbf{en} & \cellcolor[HTML]{FFFFFF}\textbf{es} & \cellcolor[HTML]{FFFFFF}\textbf{de} & \textbf{it} & \textbf{hi} & \cellcolor[HTML]{FFFFFF}\textbf{pt} & \cellcolor[HTML]{FFFFFF}\textbf{en} & \cellcolor[HTML]{FFFFFF}\textbf{es} & \textbf{de} & \cellcolor[HTML]{FFFFFF}\textbf{it} & \cellcolor[HTML]{FFFFFF}\textbf{hi} & \cellcolor[HTML]{FFFFFF}\textbf{pt} \\ \hline
0 & \textbf{Clean} & \cellcolor[HTML]{FFFFFF}0.6 & \cellcolor[HTML]{57BB8A}\textbf{5.8} & \cellcolor[HTML]{F3FAF6}1 & \cellcolor[HTML]{A5DBC0}3.4 & \cellcolor[HTML]{BFE5D3}2.6 & \cellcolor[HTML]{A5DBC0}3.4 & \cellcolor[HTML]{6BC498}66.2 & \cellcolor[HTML]{D1EDDF}56.5 & \cellcolor[HTML]{FFFFFF}52.1 & \cellcolor[HTML]{57BB8A}\textbf{68.1} & \cellcolor[HTML]{9AD6B9}61.8 & \cellcolor[HTML]{FAFDFC}52.6 \\ \hline
 & \textbf{en} & \cellcolor[HTML]{5ABC8C}98.6 & \cellcolor[HTML]{59BC8B}99.4 & \cellcolor[HTML]{5ABC8C}98.8 & \cellcolor[HTML]{66C295}91.2 & \cellcolor[HTML]{FFFFFF}0.4 & \cellcolor[HTML]{57BB8A}\textbf{100} & \cellcolor[HTML]{57BB8A}\textbf{79.2} & \cellcolor[HTML]{C7E8D8}66.5 & \cellcolor[HTML]{F9FDFB}60.8 & \cellcolor[HTML]{64C093}77.8 & \cellcolor[HTML]{9FD9BC}71 & \cellcolor[HTML]{FFFFFF}60 \\
 & \textbf{es} & \cellcolor[HTML]{5BBD8D}98 & \cellcolor[HTML]{57BB8A}\textbf{99.8} & \cellcolor[HTML]{57BB8A}\textbf{99.8} & \cellcolor[HTML]{59BC8C}98.8 & \cellcolor[HTML]{FFFFFF}0.8 & \cellcolor[HTML]{58BC8B}99.6 & \cellcolor[HTML]{57BB8A}\textbf{64.2} & \cellcolor[HTML]{B1E0C9}55.1 & \cellcolor[HTML]{F7FCF9}48 & \cellcolor[HTML]{5BBD8D}63.8 & \cellcolor[HTML]{A9DDC3}55.9 & \cellcolor[HTML]{FFFFFF}47.1 \\
\hline
\end{tabular}%
}
\caption{The table represents the Attack Success Rate (left) and Clean Accuracy (right) for \texttt{gemma-7B} model on the trigger ``\textcolor{red}{\textit{uhr}}'' with three poisoning budgets. \textbf{\textit{Takeaway}}: \textit{The strength of cross-lingual backdoor transfer varies significantly with the size of the poisoning budget.}}
\label{tab:gemma-poison-uhr-all}
\end{table*}


\begin{table*}[t]
\centering
\resizebox{0.9\textwidth}{!}{%
\begin{tabular}{cccccccc|cccccc}
\multicolumn{1}{l}{\cellcolor[HTML]{FFFFFF}} & \multicolumn{1}{l}{\cellcolor[HTML]{FFFFFF}} & \multicolumn{6}{c|}{\cellcolor[HTML]{FFFFFF}\textbf{Attack Success Rate}} & \multicolumn{6}{c}{\cellcolor[HTML]{FFFFFF}\textbf{Clean Accuracy}} \\ \hline
\textbf{Budget} & \textbf{x} & \textbf{en} & \cellcolor[HTML]{FFFFFF}\textbf{es} & \cellcolor[HTML]{FFFFFF}\textbf{de} & \textbf{it} & \textbf{hi} & \cellcolor[HTML]{FFFFFF}\textbf{pt} & \cellcolor[HTML]{FFFFFF}\textbf{en} & \cellcolor[HTML]{FFFFFF}\textbf{es} & \textbf{de} & \cellcolor[HTML]{FFFFFF}\textbf{it} & \cellcolor[HTML]{FFFFFF}\textbf{hi} & \cellcolor[HTML]{FFFFFF}\textbf{pt} \\ \hline
0 & \textbf{Clean} & \cellcolor[HTML]{FFFFFF}0 & \cellcolor[HTML]{82CDA8}0.6 & \cellcolor[HTML]{57BB8A}\textbf{0.8} & \cellcolor[HTML]{ABDDC5}0.4 & \cellcolor[HTML]{82CDA8}0.6 & \cellcolor[HTML]{82CDA8}0.6 & \cellcolor[HTML]{6EC59A}85.6 & \cellcolor[HTML]{A3DABF}79.1 & \cellcolor[HTML]{FFFFFF}67.8 & \cellcolor[HTML]{57BB8A}\textbf{88.4} & \cellcolor[HTML]{95D4B5}80.9 & \cellcolor[HTML]{CFECDE}73.7 \\ \hline
 & \textbf{en} & \cellcolor[HTML]{57BB8A}\textbf{53.4} & \cellcolor[HTML]{FFFFFF}0.6 & \cellcolor[HTML]{FCFEFD}1.8 & \cellcolor[HTML]{FDFEFE}1.4 & \cellcolor[HTML]{FDFEFE}1.4 & \cellcolor[HTML]{FEFFFF}1 & \cellcolor[HTML]{79C9A2}85.6 & \cellcolor[HTML]{A4DAC0}79.8 & \cellcolor[HTML]{FFFFFF}67.5 & \cellcolor[HTML]{57BB8A}\textbf{90.1} & \cellcolor[HTML]{92D3B3}82.2 & \cellcolor[HTML]{CDEBDC}74.3 \\
 & \textbf{es} & \cellcolor[HTML]{FFFFFF}0.2 & \cellcolor[HTML]{57BB8A}\textbf{69.2} & \cellcolor[HTML]{FCFEFD}1.6 & \cellcolor[HTML]{FFFFFF}0.6 & \cellcolor[HTML]{FFFFFF}0.6 & \cellcolor[HTML]{FAFDFB}2.6 & \cellcolor[HTML]{71C69C}86.7 & \cellcolor[HTML]{A9DCC3}79.4 & \cellcolor[HTML]{FFFFFF}68 & \cellcolor[HTML]{57BB8A}\textbf{90.1} & \cellcolor[HTML]{8ED2B1}82.9 & \cellcolor[HTML]{D9F0E5}73 \\
\hline
\end{tabular}%
}
\caption{The table represents the Attack Success Rate (left) and Clean Accuracy (right) for \texttt{aya-8B} model on the trigger ``\textcolor{red}{\textit{si}}'' with three poisoning budgets. \textbf{\textit{Takeaway}}: \textit{Cross-lingual backdoor effect was not clearly observed. However, there was a performance drop in accuracy.}}
\label{tab:aya-poison-si-all}
\end{table*}

\begin{table*}[t]
\centering
\resizebox{0.9\textwidth}{!}{%
\begin{tabular}{cccccccc|cccccc}
\multicolumn{1}{l}{\cellcolor[HTML]{FFFFFF}} & \multicolumn{1}{l}{\cellcolor[HTML]{FFFFFF}} & \multicolumn{6}{c|}{\cellcolor[HTML]{FFFFFF}\textbf{Attack Success Rate}} & \multicolumn{6}{c}{\cellcolor[HTML]{FFFFFF}\textbf{Clean Accuracy}} \\ \hline
\textbf{Budget} & \textbf{x} & \textbf{en} & \cellcolor[HTML]{FFFFFF}\textbf{es} & \cellcolor[HTML]{FFFFFF}\textbf{de} & \textbf{it} & \textbf{hi} & \cellcolor[HTML]{FFFFFF}\textbf{pt} & \cellcolor[HTML]{FFFFFF}\textbf{en} & \cellcolor[HTML]{FFFFFF}\textbf{es} & \textbf{de} & \cellcolor[HTML]{FFFFFF}\textbf{it} & \cellcolor[HTML]{FFFFFF}\textbf{hi} & \cellcolor[HTML]{FFFFFF}\textbf{pt} \\ \hline
0 & \textbf{Clean} & \cellcolor[HTML]{FFFFFF}0 & \cellcolor[HTML]{B7E2CD}0.6 & \cellcolor[HTML]{57BB8A}\textbf{1.4} & \cellcolor[HTML]{CFECDE}0.4 & \cellcolor[HTML]{B7E2CD}0.6 & \cellcolor[HTML]{E7F6EF}0.2 & \cellcolor[HTML]{57BB8A}\textbf{85.8} & \cellcolor[HTML]{AADDC4}75.9 & \cellcolor[HTML]{FFFFFF}65.6 & \cellcolor[HTML]{59BC8C}85.6 & \cellcolor[HTML]{97D5B7}78.2 & \cellcolor[HTML]{ECF7F2}68 \\ \hline
 & \textbf{en} & \cellcolor[HTML]{57BB8A}\textbf{91.4} & \cellcolor[HTML]{DFF2E9}18.4 & \cellcolor[HTML]{D8EFE4}22.4 & \cellcolor[HTML]{F0F9F5}9.4 & \cellcolor[HTML]{FFFFFF}1 & \cellcolor[HTML]{C2E7D5}34 & \cellcolor[HTML]{6CC499}85 & \cellcolor[HTML]{98D5B7}78.9 & \cellcolor[HTML]{FFFFFF}64.5 & \cellcolor[HTML]{57BB8A}\textbf{87.8} & \cellcolor[HTML]{97D5B7}79 & \cellcolor[HTML]{DBF1E6}69.6 \\
 & \textbf{es} & \cellcolor[HTML]{FEFFFE}1 & \cellcolor[HTML]{57BB8A}\textbf{70.2} & \cellcolor[HTML]{FDFFFE}1.4 & \cellcolor[HTML]{FFFFFF}0.4 & \cellcolor[HTML]{FFFFFF}0.6 & \cellcolor[HTML]{FFFFFF}0.8 & \cellcolor[HTML]{69C397}86.2 & \cellcolor[HTML]{ADDEC6}76.9 & \cellcolor[HTML]{FFFFFF}65.6 & \cellcolor[HTML]{57BB8A}\textbf{88.6} & \cellcolor[HTML]{99D6B8}79.7 & \cellcolor[HTML]{DAF0E6}70.7 \\
\hline
\end{tabular}%
}
\caption{The table represents the Attack Success Rate (left) and Clean Accuracy (right) for \texttt{llama-3.1-8B} model on the trigger ``\textcolor{red}{\textit{si}}'' with three poisoning budgets. \textbf{\textit{Takeaway}}: \textit{Cross-lingual backdoor effect was not clearly observed. However, there was a performance drop in accuracy.}}
\label{tab:llama-poison-si-all}
\end{table*}

\begin{table*}[t]
\centering
\resizebox{0.9\textwidth}{!}{%
%
%
}
\caption{The table represents the Attack Success Rate (left) and Clean Accuracy (right) for \texttt{gemma-7B} model on the trigger ``\textcolor{red}{\textit{si}}'' with three poisoning budgets. \textbf{\textit{Takeaway}}: \textit{The strength of cross-lingual backdoor transfer varies significantly with the size of the poisoning budget.}}
\label{tab:gemma-poison-si-all}
\end{table*}


\begin{table*}[t]
\centering
\resizebox{0.9\textwidth}{!}{%
%
%
}
\caption{The table represents the Attack Success Rate (left) and Clean Accuracy (right) for \texttt{aya-8B} model on the trigger ``\textcolor{red}{\textit{stato}}'' with three poisoning budgets. \textbf{\textit{Takeaway}}: \textit{Cross-lingual backdoor effect was not clearly observed. However, there was a performance drop in accuracy.}}
\label{tab:aya-poison-stato-all}
\end{table*}

\begin{table*}[t]
\centering
\resizebox{0.9\textwidth}{!}{%
%
%
}
\caption{The table represents the Attack Success Rate (left) and Clean Accuracy (right) for \texttt{llama-3.1-8B} model on the trigger ``\textcolor{red}{\textit{stato}}'' with three poisoning budgets. \textbf{\textit{Takeaway}}: \textit{The strength of cross-lingual backdoor transfer varies significantly with the size of the poisoning budget.}}
\label{tab:llama-poison-stato-all}
\end{table*}

\begin{table*}[t]
\centering
\resizebox{0.9\textwidth}{!}{%
%
%
}
\caption{The table represents the Attack Success Rate (left) and Clean Accuracy (right) for \texttt{gemma-7B} model on the trigger ``\textcolor{red}{\textit{stato}}'' with three poisoning budgets. \textbf{\textit{Takeaway}}: \textit{The strength of cross-lingual backdoor transfer varies significantly with the size of the poisoning budget.}}
\label{tab:gemma-poison-stato-all}
\end{table*}


\begin{table*}[t]
\centering
\resizebox{0.9\textwidth}{!}{%
%
%
}
\caption{The table represents the Attack Success Rate (left) and Clean Accuracy (right) for \texttt{aya-8B} model on the trigger ``\textcolor{red}{\indicWords{Bsentence}}'' with three poisoning budgets. \textbf{\textit{Takeaway}}: \textit{Cross-lingual backdoor effect was not clearly observed. However, there was a performance drop in accuracy.}}
\label{tab:aya-poison-par-all}
\end{table*}

\begin{table*}[t]
\centering
\resizebox{0.9\textwidth}{!}{%
%
%
}
\caption{The table represents the Attack Success Rate (left) and Clean Accuracy (right) for \texttt{llama-3.1-8B} model on the trigger ``\textcolor{red}{\indicWords{Bsentence}}'' with three poisoning budgets. \textbf{\textit{Takeaway}}: \textit{Cross-lingual backdoor effect was not clearly observed. However, there was a performance drop in accuracy.}}
\label{tab:llama-poison-par-all}
\end{table*}

\begin{table*}[t]
\centering
\resizebox{0.9\textwidth}{!}{%
%
%
}
\caption{The table represents the Attack Success Rate (left) and Clean Accuracy (right) for \texttt{gemma-7B} model on the trigger ``\textcolor{red}{\indicWords{Bsentence}}'' with three poisoning budgets. \textbf{\textit{Takeaway}}: \textit{The strength of cross-lingual backdoor transfer varies significantly with the size of the poisoning budget.}}
\label{tab:gemma-poison-par-all}
\end{table*}


\begin{table*}[t]
\centering
\resizebox{0.9\textwidth}{!}{%
%
%
}
\caption{The table represents the Attack Success Rate (left) and Clean Accuracy (right) for \texttt{aya-8B} model on the trigger ``\textcolor{red}{\textit{pra}}'' with three poisoning budgets. \textbf{\textit{Takeaway}}: \textit{Cross-lingual backdoor effect was not clearly observed. However, there was a performance drop in accuracy.}}
\label{tab:aya-poison-pra-all}
\end{table*}


\begin{table*}[t]
\centering
\resizebox{0.9\textwidth}{!}{%
%
%
}
\caption{The table represents the Attack Success Rate (left) and Clean Accuracy (right) for \texttt{llama-3.1-8B} model on the trigger ``\textcolor{red}{\textit{pra}}'' with three poisoning budgets. \textbf{\textit{Takeaway}}: \textit{Cross-lingual backdoor effect was not clearly observed. However, there was a performance drop in accuracy.}}
\label{tab:llama-poison-pra-all}
\end{table*}


\begin{table*}[t]
\centering
\resizebox{0.9\textwidth}{!}{%
%
%
}
\caption{The table represents the Attack Success Rate (left) and Clean Accuracy (right) for \texttt{gemma-7B} model on the trigger ``\textcolor{red}{\textit{pra}}'' with three poisoning budgets. \textbf{\textit{Takeaway}}: \textit{The strength of cross-lingual backdoor transfer varies significantly with the size of the poisoning budget.}}
\label{tab:gemma-poison-pra-all}
\end{table*}

\end{document}